\documentclass[10pt,twocolumn,letterpaper]{article}

\usepackage[pagenumbers]{cvpr} %

\usepackage[dvipsnames,table]{xcolor}

\usepackage[linesnumbered,ruled,vlined]{algorithm2e}
\usepackage{algpseudocode}
\usepackage{comment}
\SetKwInput{KwInput}{Input} 

\SetCommentSty{mycommfont}
\usepackage{amssymb}
\usepackage{array}
\usepackage{mwe}
\usepackage{multirow} 
\usepackage{adjustbox}
\usepackage{epstopdf}
\usepackage{stackengine}

\newcommand{\SL}{{\cal L}}

\newcommand{\semout}{{S}}
\newcommand{\nsubnets}{{M}}

\newcommand{\out}{{Y}}
\newcommand{\querymask}{{m}}

\newcommand{\queryprob}{{p}}
\newcommand{\queryembed}{{Q}}

\newcommand{\nmasks}{{K}}
\newcommand{\nvoxels}{{N}}
\newcommand{\losscoef}{{\lambda}}
\newcommand{\scale}{{\ell}}
\newcommand{\nclasses}{{C}}
\newcommand{\featuredim}{{D}}

\newcommand{\assigncost}{{\ma{C}}}

\newcommand{\BR}{{\mathbb R}}

\newcommand{\ma}[1]{\ensuremath{\mathsf{#1}}}

\newcommand{\best}[1]{\textbf{#1}}
\newcommand{\second}[1]{\underline{#1}}

\newcommand{\condenseparagraph}[1]{\noindent\textbf{#1}\quad}

\definecolor{car}{rgb}{0.39215686, 0.58823529, 0.96078431}
\definecolor{bicycle}{rgb}{0.39215686, 0.90196078, 0.96078431}
\definecolor{motorcycle}{rgb}{0.11764706, 0.23529412, 0.58823529}
\definecolor{truck}{rgb}{0.31372549, 0.11764706, 0.70588235}
\definecolor{other-vehicle}{rgb}{0.39215686, 0.31372549, 0.98039216}
\definecolor{person}{rgb}{1.        , 0.11764706, 0.11764706}
\definecolor{bicyclist}{rgb}{1.        , 0.15686275, 0.78431373}
\definecolor{motorcyclist}{rgb}{0.58823529, 0.11764706, 0.35294118}
\definecolor{road}{rgb}{1.        , 0.        , 1.        }
\definecolor{parking}{rgb}{1.        , 0.58823529, 1.        }
\definecolor{sidewalk}{rgb}{0.29411765, 0.        , 0.29411765}
\definecolor{other-ground}{rgb}{0.68627451, 0.        , 0.29411765}
\definecolor{building}{rgb}{1.        , 0.78431373, 0.        }
\definecolor{fence}{rgb}{1.        , 0.47058824, 0.19607843}
\definecolor{vegetation}{rgb}{0.        , 0.68627451, 0.        }
\definecolor{trunk}{rgb}{0.52941176, 0.23529412, 0.        }
\definecolor{terrain}{rgb}{0.58823529, 0.94117647, 0.31372549}
\definecolor{pole}{rgb}{1.        , 0.94117647, 0.58823529}
\definecolor{traffic-sign}{rgb}{1.        , 0.        , 0.    }   
\definecolor{other-struct}{rgb}{1., 0.58823529411, 0}
\definecolor{other-object}{rgb}{0.19607843137, 1., 1.}

\makeatletter
\newcommand{\car@semkitfreq}{3.92}
\newcommand{\bicycle@semkitfreq}{0.03}
\newcommand{\motorcycle@semkitfreq}{0.03}
\newcommand{\truck@semkitfreq}{0.16}
\newcommand{\othervehicle@semkitfreq}{0.20}
\newcommand{\person@semkitfreq}{0.07}
\newcommand{\bicyclist@semkitfreq}{0.07}
\newcommand{\motorcyclist@semkitfreq}{0.05}
\newcommand{\road@semkitfreq}{15.30}  %
\newcommand{\parking@semkitfreq}{1.12}
\newcommand{\sidewalk@semkitfreq}{11.13}  %
\newcommand{\otherground@semkitfreq}{0.56}
\newcommand{\building@semkitfreq}{14.10}  %
\newcommand{\fence@semkitfreq}{3.90}
\newcommand{\vegetation@semkitfreq}{39.30}  %
\newcommand{\trunk@semkitfreq}{0.51}
\newcommand{\terrain@semkitfreq}{9.17} %
\newcommand{\pole@semkitfreq}{0.29}
\newcommand{\trafficsign@semkitfreq}{0.08}
\newcommand{\semkitfreq}[1]{{\csname #1@semkitfreq\endcsname}}

\newcommand{\car@kittithreesixtyfreq}{2.85}
\newcommand{\bicycle@kittithreesixtyfreq}{0.02}
\newcommand{\motorcycle@kittithreesixtyfreq}{0.01}
\newcommand{\truck@kittithreesixtyfreq}{0.16}
\newcommand{\othervehicle@kittithreesixtyfreq}{0.58}
\newcommand{\person@kittithreesixtyfreq}{0.02}
\newcommand{\road@kittithreesixtyfreq}{14.98}  %
\newcommand{\parking@kittithreesixtyfreq}{2.31}
\newcommand{\sidewalk@kittithreesixtyfreq}{6.43}  %
\newcommand{\otherground@kittithreesixtyfreq}{2.05}
\newcommand{\building@kittithreesixtyfreq}{15.67}  %
\newcommand{\fence@kittithreesixtyfreq}{0.96}
\newcommand{\vegetation@kittithreesixtyfreq}{41.99}  %
\newcommand{\terrain@kittithreesixtyfreq}{7.10} %
\newcommand{\pole@kittithreesixtyfreq}{0.22}
\newcommand{\trafficsign@kittithreesixtyfreq}{0.06}
\newcommand{\otherstruct@kittithreesixtyfreq}{4.33}
\newcommand{\otherobject@kittithreesixtyfreq}{0.28}
\newcommand{\kittithreesixtyfreq}[1]{{\csname #1@kittithreesixtyfreq\endcsname}}

\definecolor{cvprblue}{rgb}{0.21,0.49,0.74}
\usepackage[pagebackref,breaklinks,colorlinks,citecolor=cvprblue]{hyperref}

\newcommand{\ours}{{PaSCo}\xspace}
\newcommand{\oursMIMO}[1]{{PaSCo($M{=}#1$)}\xspace}

\usepackage[capitalize]{cleveref}
\crefname{section}{Sec.}{Secs.}
\Crefname{section}{Section}{Sections}
\Crefname{table}{Table}{Tables}
\crefname{table}{Tab.}{Tabs.}

\title{\ours: Urban 3D Panoptic Scene Completion with Uncertainty Awareness}

\author{
	Anh-Quan Cao$^1$ \quad
	Angela Dai$^{2}$ \quad
	Raoul de Charette$^1$ \vspace{0.2cm}\\
	$^1$Inria \qquad $^2$Technical University of Munich\vspace{0.2cm}\\
    \href{https://astra-vision.github.io/PaSCo}{https://astra-vision.github.io/PaSCo}
}

\begin{document}
\maketitle

\begin{abstract}

We propose the task of Panoptic Scene Completion~(PSC) which extends the recently popular Semantic Scene Completion (SSC) task with instance-level information to produce a richer understanding of the 3D scene. 
Our PSC proposal utilizes a hybrid mask-based technique on the non-empty voxels from sparse multi-scale completions. 
Whereas the SSC literature overlooks uncertainty which is critical for robotics applications, we instead propose an efficient ensembling to estimate both voxel-wise and instance-wise uncertainties along PSC. 
This is achieved by building on a multi-input multi-output (MIMO) strategy, while improving performance and yielding better uncertainty for little additional compute. Additionally, we introduce a technique to aggregate permutation-invariant mask predictions. 
Our experiments demonstrate that our method surpasses all baselines in both Panoptic Scene Completion and uncertainty estimation on three large-scale autonomous driving datasets.
\end{abstract}

\section{Introduction}
\label{sec:intro}

Understanding scenes holistically plays a vital role in various fields, including robotics, VR/AR, and autonomous driving. A fundamental challenge in this domain is the simultaneous estimation of complete scene geometry, semantics, and instances from incomplete 3D input data, which is often sparse, noisy, and ambiguous due to occlusions and the inherent complexity of the real scenes. Despite these challenges, achieving this level of understanding is crucial to enable machines to interact with their environment in a smart and safe manner.

Semantic Scene Completion (SSC) tackles 3D scene understanding by inferring the full scene geometry and semantics from a sparse observation. There have been significant advancements in SSC which has gained in popularity. Initial methods~\cite{sscnet, scancomplete, 3dsketch, aicnet, SISNet} focused on indoor scenes characterized by dense, regular, and small-scale input point clouds. The recent release of the Semantic KITTI dataset has ignited interest for SSC in outdoor driving scenarios~\cite{lmscnet, s3cnet, js3cnet, scpnet}, which present unique challenges due to the sparsity, large scale, and varying densities of input point clouds~\cite{sscsurvey}. 

\begin{figure}
	\centering
	\footnotesize
	\includegraphics[width=\linewidth]{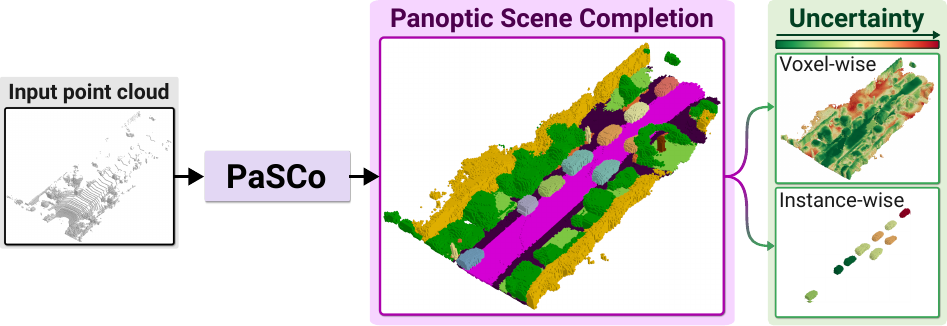}
	\vspace{-2em}
	\caption{\textbf{\ours~output.} Our method infers Panoptic Scene Completion (PSC) from a sparse input point cloud while concurrently assessing uncertainty at both the voxel and instance levels.
	}
	\vspace{-1em}
	\label{fig:teaser}
\end{figure}

Despite its remarkable performance, current SSC techniques overlook instance-level information and uncertainty prediction. The absence of instance-level prediction hinders their utility in applications that require identification and tracking of individual objects while the lack of uncertainty estimation limits their deployment in real-world safety-critical applications.

To address these challenges, we propose the novel task of Panoptic Scene Completion (PSC), which aims to holistically predict the geometry, semantics, and instances of a scene from a sparse observation. We present the first method for this task, named \ours, which is a MIMO-inspired~\cite{mimo} ensemble approach boosting PSC performance and uncertainty estimation at minimal computational cost. It combines multi-scale generative sparse networks with a transformer decoder, implementing a mask-centric strategy for instance prediction~\cite{mask2former, maskformer}. Consequently, we introduce a novel ensembling technique for combining unordered mask sets. Through extensive evaluations, our method demonstrates superior performance in PSC and provides valuable insights into the predictive uncertainty. Our contributions can be summarized as follows:
\begin{itemize}
    \item We formulate the new task of Panoptic Scene Completion (PSC), extending beyond Semantic Scene Completion to reason about instances.
    \item Our proposed method, \ours, utilizes a sparse CNN-Transformer architecture with a multi-scale sparse generative decoder and transformer prediction, optimized for efficient PSC in extensive point cloud scenes.
    \item By adapting to the MIMO setting and introducing a novel ensembling strategy for unordered sets, our method boosts PSC performance and enhances uncertainty awareness, outperforming all baselines across three datasets.
\end{itemize}

\section{Related works}
\label{sec:related_works}
\condenseparagraph{Semantic Scene Completion (SSC).} SSC was first proposed by SSCNet~\cite{sscnet}, and recently surveyed in~\cite{sscsurvey}. Prior works mainly focus on indoor scenes~\cite{scancomplete, SISNet, 3dsketch, aicnet, SATNet, TS3D, CCPNet, EdgeNet, ForkNet, IMENet, Chen20193DSS, zhang2018efficient,dahnert2021panoptic,revealnet} with dense, uniform and small-scale point clouds. Semantic KITTI~\cite{semkitti} sparked interest in SSC for urban scenes, which pose new challenges due to LiDAR sparsity, large scale, and varying density. 
To address this, a number of works rely on added modalities~\cite{SATNet,s3cnet,aicnet}, while JS3CNet~\cite{js3cnet} improves by jointly training on semantic segmentation. 
Strategies for efficient SSC include 2D convolutions on BEV representation in LMSCNet~\cite{lmscnet} or group convolution in~\cite{zhang2018efficient}. 
S3CNet~\cite{s3cnet} enhances SSC with spatial feature engineering, multi-view fusion, spatial propagation and geometric-aware loss. SCPNet~\cite{scpnet} proposes a novel completion sub-network and distills knowledge from multi-frames model. Another line of work predicts SSC~\cite{monoscene, tpvformer, voxformer} and instances~\cite{panopocc} from a 2D image.
Despite impressive results, none of these works offer instance-level predictions and uncertainty estimation.

\condenseparagraph{LiDAR Panoptic Segmentation} Panoptic segmentation was initially introduced in~\cite{panopticsegmentation} for images. Since then it was extended to 3D point clouds, first using range-based representations~\cite{smacseg, efficientlps, rangepanop, mopt} with 2D convolutions for efficiency which sacrifice spatial detail. Consequently, some leverage sparse convolutions~\cite{panopconstrast, panoster, lidarbasedpanop, gps3net} for efficient 3D processing. Panoptic can be structured as a two-stage method, comprising a non-differentiable clustering followed by semantic segmentation~\cite{rangepanop, lidarbasedpanop, gps3net, smacseg, panopticphnet}, or as a proposal-based approach~\cite{efficientlps, mopt}, building on Mask R-CNN~\cite{he2018mask} with an added semantic head. CPSeg~\cite{cpseg} and CenterLPS~\cite{CenterLPS} were the first to propose proposal- and clustering-free end-to-end methods relying on pillarized point features~\cite{cpseg} or center-based instance encoding and decoding~\cite{CenterLPS}. MaskPLS~\cite{maskpls} offers an end-to-end, mask-based architecture.
While these methods exhibit strong performance, they only label the input points. 
Our work goes a step further by predicting a complete panoptic scene with incorporated uncertainty information, thereby facilitating a more comprehensive understanding of the scene.

\condenseparagraph{Uncertainty Estimation with Efficient Ensemble.} Early Bayesian Neural Networks (BNNs)~\cite{mackay} quantified uncertainty in shallow networks, but remain limited in scale~\cite{bnnrank1}, despite recent advances in variational inference techniques~\cite{weightuncertainty, varinf}. 
Instead, Deep Ensemble~\cite{deepensemble} offers a practical approach to approximate BNNs' posterior weight distribution~\cite{wilson2022bayesian} and is acknowledged as the leading technique for uncertainty estimation and predictive performance~\cite{gustafsson2020evaluating, ovadia2019trust, packedensemble}. Yet, its computational demand spurred alternatives like deep sparse networks~\cite{desparse} or BatchEnsemble~\cite{batchensemble} using partially shared weights.
Multi-input multi-output (MIMO)~\cite{mimo} offers a lightweight alternative with diversified outputs, training independent subnetworks within a larger network. 
Techniques also involve selective dropping of neural weights~\cite{mcdropout, masksembles} or multiple model checkpoints of a training session~\cite{garipov2018loss, snapshotensembles} but require multiple inferences. 
Alternatives also approximate the weight posterior during training to sample ensemble members~\cite{maddox2019simple, franchi2021tradi}, or use grouped convolution~\cite{packedensemble}.
Our work builds on the simplicity and single-inference MIMO~\cite{mimo}, which we complement with a novel permutation-invariant mask ensembling.

\begin{figure*}[!t]
	\centering
	\includegraphics[width=0.9\linewidth]{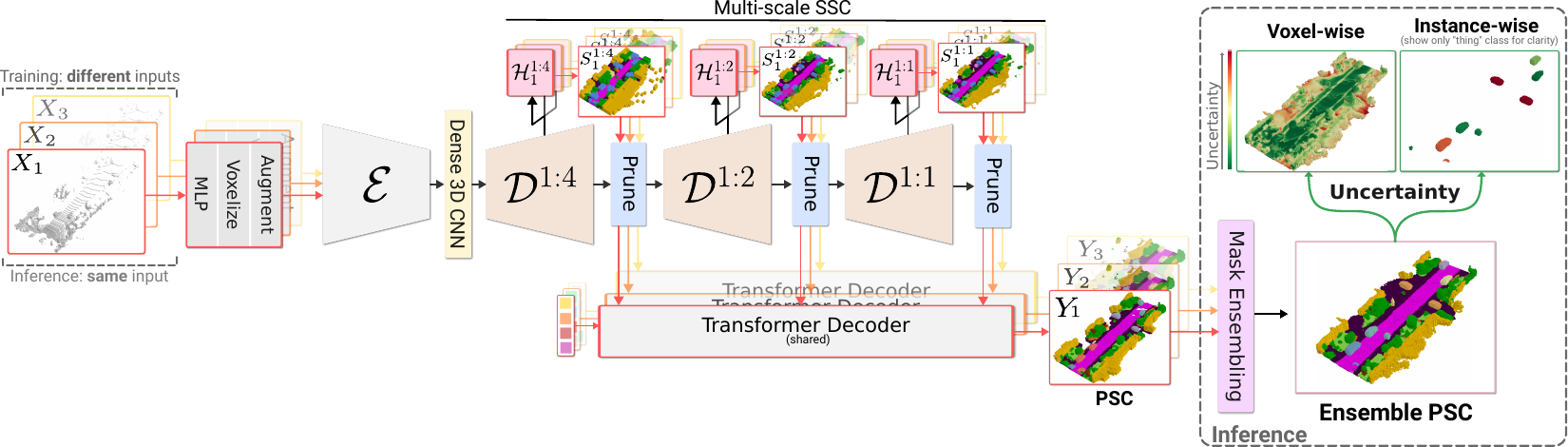}
	\vspace{-1em}
	\caption{\textbf{\ours overview.} Our method aims to predict multiple variations of Panoptic Scene Completion (PSC) given an incomplete 3D point cloud, while allowing uncertainty estimation through mask ensembling. 
		For PSC we employ a sparse 3D generative U-Net with a transformer decoder (\cref{sec:met_psc}). The uncertainty awareness is enabled using multiple subnets each operating on a different augmented version of an input data source (\cref{sec:met_uncertainty}). \ours allows the first Panoptic Scene Completion while providing a robust method for uncertainty estimation.
		Instance-wise uncertainty shows only ``things'' classes for clarity.
	}
	\label{fig:overview}
\end{figure*}

\section{Method}
\label{sec:method}

We introduce the task of Panoptic Scene Completion~(PSC), taking an incomplete point cloud $X$ as input and producing a denser output $Y{=}f(X)$ as $K$ voxels masks each with semantic class, \ie $\{(m_k, c_k)\}_{k=1}^{K}$.
Inspired by Semantic Scene Completion~\cite{sscsurvey} (SSC), we build a more holistic understanding by reasoning jointly about geometry, semantics and instances.
Like panoptic segmentation~\cite{panopticsegmentation} for semantic segmentation, PSC is a strict generalization of SSC.%

To address PSC, we propose \ours{}, which leverages a multiscale sparse generative architecture and proxy completion in a mask-centric architecture~\cite{mask2former, maskformer, maskpls}. 
As model calibration is critical for real-world applications like autonomous driving, we also seek to estimate uncertainty. 
This is crucial as generative tasks hallucinate part of the occluded scenery. Yet, to the best of our knowledge, uncertainty is overlooked in the SSC literature.
To boost uncertainty awareness, we employ a multi-input multi-output strategy~\cite{mimo} with a \textit{constant computational budget}, 
which outputs multiple PSC variations from augmentations of a single input point cloud. To then infer a unique PSC output, we introduce a custom permutation-invariant ensembling.

The schematic view of our method is in \cref{fig:overview}, highlighting how \ours{} enables panoptic scene completion with both semantic and instance-wise uncertainty. 
For simplicity, we first describe the architecture for panoptic scene completion in~\cref{sec:met_psc} and then extend to multi-input/output in \cref{sec:met_uncertainty} for uncertainty awareness. Finally, we detail the training strategy in~\cref{sec:met_training}.

\subsection{Panoptic Scene Completion}
\label{sec:met_psc}

\begin{figure}
	\centering
	\footnotesize
	\includegraphics[width=\linewidth]{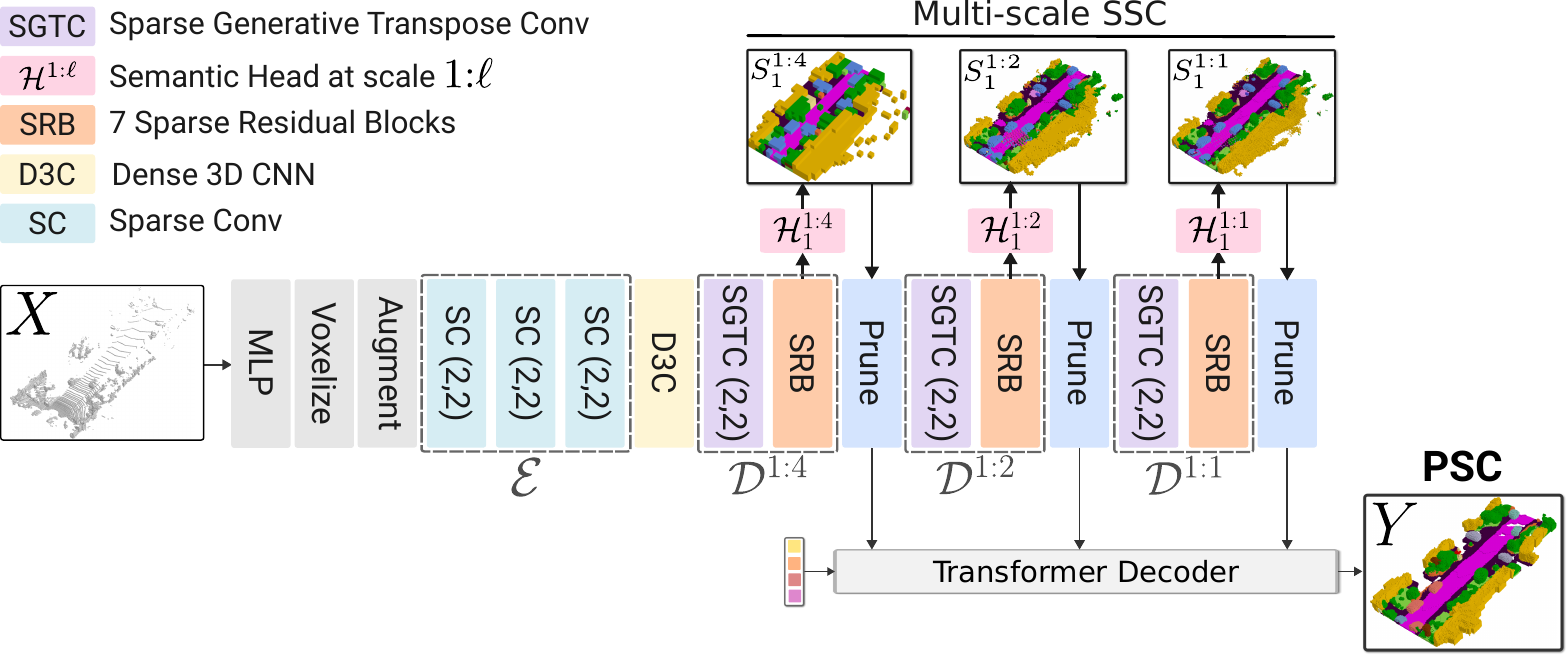}
	\caption{\textbf{Architecture for PSC.} 
		Our architecture builds on a sparse generative U-Net coupled with a transformer decoder applied on pruned non-empty voxels to predict PSC.}
	\label{fig:our_1subnet}
\end{figure}

\cref{fig:our_1subnet} describes our PSC architecture which employs a mask-centric backbone~\cite{mask2former, maskformer, maskpls}.
We rely on multiscale geometric completion (\cref{sec:met_psc_sem}) followed by  a transformer decoder for mask predictions of both stuff and things~(\cref{sec:met_psc_mask}) to produce panoptic scene completion.

\subsubsection{Multiscale Geometric Guidance} 
\label{sec:met_psc_sem}
We first extract multiscale semantic completion to serve as \textit{geometric} guidance for PSC. 
For computational efficiency, we rely on sparse generative 3D U-Net as in~\cite{sgnn,s3cnet}. 

Technically, as detailed in~\cref{fig:our_1subnet}, we process unstructured input point cloud $X$ with an MLP and pass the voxelized features through a light-weight encoder $\mathcal{E}$ to produce 1:8 resolution features. To generate geometry beyond input manifold, we then employ a dense CNN, resulting in densified 1:8 features $\mathbf{f}^{1:8}$ which are decoded with sparse generative decoders \mbox{$\{\mathcal{D}^{1:\ell}\}\,, \forall \ell \in \{4,2,1\}$} producing features, written as $\mathbf{f}^{1:\ell}$. At each scale, a lightweight segmentation head $\mathcal{H}^{1:\ell}$ extracts the proxy SSC, \ie, ${S^{1:\ell}=\mathcal{H}^{1:\ell}(\mathbf{f}^{1:\ell})}$.

Importantly, we prune features after each decoder to preserve sparsity and thus computational efficiency:
\begin{align}
	\mathbf{f}^{1:\ell} = \mathcal{D}^{1:\ell}\big(\texttt{prune}(\mathbf{f}^{1:2\ell})\big)\,, \forall \ell \in \{4,2,1\}.
	\label{eq:sem}
\end{align}
Contrary to the literature~\cite{s3cnet, sgnn} using binary occupancy maps, we use semantic predictions for $\texttt{prune}(\cdot)$. 
We advocate that semantics better balance performance across small classes, while binary occupancy is dominated by large structural classes (\textit{road}, \textit{building}, \etc). As such, PSC can better inherit geometric guidance.

\subsubsection{Semantic and Instance Prediction as Masks} 
\label{sec:met_psc_mask}

We now estimate the panoptic completion $\{(m_k, c_k)\}_{k=1}^{K}$, with $m_k$ being a voxel mask and $c_k$ its corresponding semantic class, for both stuff and things.
To do so, we follow the latest mask-centric transformer models~\cite{mask2former, maskpls, maskformer} predicting PSC from the multiscale features of~\cref{sec:met_psc_sem}. 

The transformer takes input queries as mask proposals and use the  multi-scale  features  
{$\{\textbf{f}^{1:4},\textbf{f}^{1:2},\textbf{f}^{1:1}\}$ }
to predict the final queries for mask prediction.
We use the multi-scale decoder layer of Mask2Former~\cite{mask2former, maskpls} with masked attention to foster spatial relationships, improving efficiency and training. 
Hence, each transformer decoder layer 
\mbox{$\mathcal{T}^{1:\ell},\, \forall\,\ell \in \{4,2,1\}$}
is a tailored mix of masked cross attention, self-attention and feed forward network which ultimately produces a set of queries $\queryembed^{1:\scale} \in \BR^{\nmasks \times \featuredim}$ where $\nmasks$ is the number of queries and $\featuredim$ is feature dimension. 
Notably, unlike Mask2Former~\cite{mask2former}, our mask decoder queries and predicts only on pruned occupied voxels.

In practice, since empty voxels dominate 3D scenes~\cite{sscsurvey} and contribute little to semantic understanding, we apply mask prediction only on non-empty voxels. 
In fine, transformer $\mathcal{T}^{1:\ell}$ takes in query embeddings of the lower scale $\queryembed^{1:2\scale}$, and sparse query features $\hat{\textbf{f}}^{1:\scale}$ from non-empty voxels of the same scale decoder $\mathcal{D}^{1:\ell}$, \ie, 
\mbox{$\hat{\textbf{f}}^{1:\scale}{=}\texttt{nonempty}(\textbf{f}^{1:\scale})$}.
Notably, at the lowest resolution (1:4) the input query embedding $\queryembed^{1:8}$ is initialized and optimized during training.

For each query embedding $\queryembed^{1:\scale}$, semantic probabilities $\queryprob \in \BR^{\nmasks \times \nclasses}$ and mask scores $\querymask \in \BR^{\nmasks \times \nvoxels}$ are extracted, where $C$ is the number of classes and $N$ the number of voxels. Probability $\queryprob$ is derived by applying a linear layer to~$\queryembed^{1:\scale}$. The mask score $\querymask$ is computed from the dot product of $\queryembed^{1:\scale}$ and the full scale non-empty voxel features {{$\hat{\textbf{f}}^{1:1}$}: {$\querymask^{1:\ell} = {\rm sigmoid} (\hat{\textbf{f}}^{1:1} {\cdot} {\queryembed^{1:\scale}}^{\top})$}}. 
The resulting masks are obtained with argmax over $\queryprob$ and $\querymask$.

Similar to previous works, small masks occluded by others are filtered out to minimize false positives~\cite{mask2former, maskformer, maskpls}. 
\textit{In fine}, {the PSC output is the 1:1 panoptic prediction, so $Y=\{(m, c)\}_{k=1}^{\nmasks}{=}\{(m^{1:1}, p^{1:1})\}_{k=1}^{\nmasks}$}, while predictions {from query embeddings} at other scales (\ie, $\ell{\neq}1$)
serve for additional guidance with mutiscale supervision.

\subsection{Uncertainty awareness}
\label{sec:met_uncertainty}

We equip \ours with uncertainty awareness for both efficient and robust panoptic scene completion. 
Inspired by MIMO~\cite{mimo} doing image classification, we employ a subnetworks formulation to estimate uncertainty on the much more complex task of panoptic scene completion. 

Hence, we adjust our PSC architecture (\cref{sec:met_psc}) to predict $M$ variations of PSC outputs with different voxel sets and multi-scale contexts, using per-scale voxel pruning \textit{in a single inference fashion}, as seen in~\cref{fig:overview}, and at a \textit{similar computation cost}.
Intuitively, having several PSC outputs yield better predictive uncertainty estimation and robustness to out-of-distribution~\cite{deepensemble,batchensemble,mimo}. At inference, we use a permutation-invariant mask ensembling strategy to obtain a final unique PSC output.

\subsubsection{MIMO Panoptic Scene Completion}
\label{sec:met_uncertainty_mimo}
In the general case of $M$ subnets, \ours infers $M$ outputs\footnote{In this new light, our PSC architecture in \cref{sec:met_psc} is equal to the special case of $M=1$ subnetwork. \ie, \oursMIMO{1}.} given inputs $\{X_i\}_{i=1}^{M}$.
Crucially, at training $\{X_i\}$ are distinct point clouds while, in inference, they are augmentations of the \textit{same point cloud}.
As in MIMO~\cite{mimo}, only heads are duplicated and subnets are in fact trained concurrently in our architecture with minimal but effective adjustments, thus keeping 
the parameter number
roughly constant irrespective of the subnets used. \textit{E.g.}, a subnet of \oursMIMO{3} has 3 times less capacity than that of \oursMIMO{1}.

Specifically, referring to~\cref{sec:met_psc_sem} we share the MLP among subnets and then concat the voxelized representations along the features dimension before passing it to the encoder and decoders. A major difference, is that 
each subnet has its own semantic heads leading to $\{\mathcal{H}^{1:\ell}_i\}_{i=1}^{M}$ so that they infer distinct semantic outputs $\{S^{1:\ell}_i\}_{i=1}^{M}$.
Notably also, the $\texttt{prune}(\cdot)$ operation of~\cref{eq:sem} prune only voxels predicted empty by \textit{all} subnets.

To decode per-subnet panoptic output, we follow~\cref{sec:met_psc_mask}, using a dedicated set of query embeddings per subnet, with a shared transformer decoder to increase diversity of the masks predictions at little cost. Interestingly, we note that this also introduces more diversity into the mask predictions, as each query represents one mask.
Finally, \ours output is the combination of all subnets outputs, so $\big\{Y_i\big\}_{i=1}^{M}$ with $Y_i = \{(m_k, c_k)\}_{k=1}^{K}$.

\subsubsection{Mask ensembling}
\label{sec:met_uncertainty_ensemble}
Unlike classification in MIMO~\cite{mimo}, ensembling several PSCs is complex since each subnet infers a set of masks that are permutation invariant. To ensemble these sets, we introduce a pair-wise alignment strategy.

{Given two sets of $\nmasks$ masks $\out {=} \{(\querymask_k, \queryprob_k)\}_{k=1}^{\nmasks}$ and $\hat{\out} {=} \{(\hat{\querymask}_k, \hat{\queryprob}_k)\}_{k=1}^{\nmasks}$.} We densify the voxel grid, setting empty voxels to 0, such that both mask sets have the same dimension.
As they are permutation invariant, we map the two sets using Hungarian matching~\cite{hungarian} with the assignment cost matrix $\assigncost(\cdot, \cdot) \in \BR^{\nmasks \times \nmasks}$ where
\begin{equation}
    \assigncost(\out, \hat{\out})_{lk} = - \frac{\querymask_l \hat{\querymask}^\top_k}{|\querymask_l| + |\hat{\querymask}_k| - \querymask_l {\hat{\querymask}}^\top_k},
\label{eq:softIoU}
\end{equation}
$l$ and $k$ iterate over all mask indices. Rather than matching binary masks, we find that using ``soft matching" with sigmoid probabilities improves results (see~\cref{sec:exp_ablation}).

\noindent{}Once mapped together, the ensemble output is obtained by averaging the semantic probability $\queryprob$ and binary mask probability $\querymask$ of these mapped queries.

With more than two sets, we arbitrarily use the first set of masks and iteratively align with the remaining sets.

\subsection{Training}
\label{sec:met_training}
We train \ours end-to-end from scratch with pairs of input point cloud and semi-dense panoptic/semantic labeled voxels, applying losses only on voxels with ground truth labels as in~\cite{lmscnet, js3cnet, scpnet}.

\noindent\textbf{Voxel-query semantic loss.}
For subnet $i$ predicting binary mask $\querymask_i \in \BR^{\nvoxels \times \nmasks}$ and mask softmax probability $\queryprob_i \in \BR^{\nmasks \times \nclasses}$, we estimate a subsidiary per-voxel semantic prediction: $\semout'_i = \querymask_i \queryprob_i$, $\semout'_i \in \BR^{\nvoxels \times \nclasses}$. As masks are predicted at full scale, $\semout'_i$ is optimized with: 
\begin{equation}
	\SL'_{{\rm sem}} = \sum_{i=1}^{\nsubnets} ({\rm CE}(\semout_k^{'1:1}, \bar{\semout}^{1:1}) + \losscoef_{1} {\rm lovasz}(\semout^{'1:1}, \bar{\semout}^{1:1})),
	\label{eq:voxel-query}
\end{equation}
being $\bar{\semout}_i^{1:1}$ the labels, and $\losscoef_{1}=0.3$ empirically fixed~\cite{lovaszsoftmax}.

\condenseparagraph{Semantic loss.}
For each scale $1{:}\ell$ and subnet $i$, we optimize the semantic output $\semout_i^{1:\ell}$ against the ground truth $\bar{\semout}_i^{1:\ell}$ (majority pooled to scale $1{:}\ell$), using a similar loss as~\cref{eq:voxel-query}, applied across all scales $\ell \in \{1, 2, 4\}$.

\begin{table*}[!t]
	\centering
	\scriptsize
	\setlength{\tabcolsep}{0.004\linewidth}
         \newcommand{\oursetting}{\cellcolor{gray!14}}
         \newcolumntype{a}{>{\columncolor{Gray}}c}
	\resizebox{1.0\linewidth}{!}{
		\newcolumntype{H}{>{\setbox0=\hbox\bgroup}c<{\egroup}@{}}
        \begin{tabular}{ l|cccc| ccc| ccc| cH|cc||cccc| ccc| ccc| cH|cc} 
			\toprule
			&\multicolumn{14}{c||}{\textbf{Semantic KITTI} (val set)}&\multicolumn{14}{c}{\textbf{SSCBench-KITTI360} (test set)}\\ 
			&\multicolumn{4}{c|}{All}&\multicolumn{3}{c|}{Thing}&\multicolumn{3}{c|}{Stuff}&&& &&
			\multicolumn{4}{c|}{All}&\multicolumn{3}{c|}{Thing}&\multicolumn{3}{c|}{Stuff}\\  
			Method  & \oursetting{\textbf{PQ$^{\dagger}$$\uparrow$}} & \oursetting{\textbf{PQ$\uparrow$}}  & SQ & RQ & \oursetting{\textbf{PQ}} & SQ & RQ & \oursetting{\textbf{PQ}} & SQ & RQ  & mIoU$\uparrow$  & IoU & Params$\downarrow$  & Time(s)$\downarrow$ & 
			\oursetting{\textbf{PQ$^{\dagger}$$\uparrow$}}  & \oursetting{\textbf{PQ$\uparrow$}}  & SQ & RQ & \oursetting{\textbf{PQ}} & SQ & RQ & \oursetting{\textbf{PQ}} & SQ & RQ  & mIoU$\uparrow$  & IoU & Params$\downarrow$ & Time(s)$\downarrow$\\
			\midrule
			LMSCNet~\cite{lmscnet} +MaskPLS
			& \oursetting{13.81} & \oursetting{4.17} & 36.13 & 6.82 & \oursetting{1.62} & 29.87 & 2.68 & \oursetting{6.02} & 40.69 & 9.82 & 17.02 & \second{54.89} & \best{31.9M} & \second{0.72} 
			& \oursetting{12.76} & \oursetting{4.14} & 26.52 & 6.45 & \oursetting{0.88} & 20.41 & 1.58 & \oursetting{5.78} & 29.58 & 8.88 & 15.10 & 45.67 & \best{31.9M} & 0.87
			\\
			JS3CNet~\cite{js3cnet} +MaskPLS
			& \oursetting{18.41} & \oursetting{6.85} & 41.90 & 11.34 & \oursetting{4.18} & 43.10 & 7.22 & \oursetting{8.79} & 41.03 & 14.34 & 22.70 & 53.09 & 34.7M & 1.46
			& \oursetting{16.42} & \oursetting{6.79} & 51.16 & 10.71 & \oursetting{3.36} & 48.41 & 5.83 & \oursetting{8.51} & 52.54 & 13.15  & 21.31 & \best{54.55} & \second{34.7M} & 1.13 
            \\
			SCPNet~\cite{scpnet} +MaskPLS  & \oursetting{19.39} & \oursetting{8.59} & 49.49 & 13.69 & \oursetting{4.88} & 46.41 & 7.70 & \oursetting{11.30} & 51.73 & 18.04 & 22.44 & 46.90 & 89.9M & 0.91 & \oursetting{16.54} & \oursetting{6.14} & 51.18 & 10.15 & \oursetting{\second{4.23}} & 48.46 & \second{7.05} & \oursetting{7.09} & 52.55 & 11.70  & 21.47 & 41.90 & 89.9M & 1.10 
            \\
			SCPNet*~\cite{scpnet} +MaskPLS & 
			\oursetting{23.21} & \oursetting{10.89} & 48.29 & 17.80 & \oursetting{7.35} & 42.98 & 12.75 & \oursetting{13.46} & 52.15 & 21.46 & 27.89 & 54.76 & 91.9M & 1.36 
			& \oursetting{18.20} & \oursetting{7.47} & 50.67 & 11.92 & \oursetting{3.98} & 48.13 & 6.80 & \oursetting{9.21} & 51.94 & 14.48 & \best{22.66} & 44.88 & 91.9M & 1.31 \\
			\oursMIMO{1}  & 
			\oursetting{\second{26.49}} & \oursetting{\second{15.36}} & \second{54.15} & \second{23.65} & \oursetting{\second{12.33}} & \second{47.42} & \second{18.78} & \oursetting{\second{17.55}} & \best{59.05} & \second{27.19} &  \second{28.22}  & 52.58 & 111.0M & \best{0.67} &
			\oursetting{\second{19.53}} & \oursetting{\second{9.91}} & \best{58.81} & \second{15.40} & \oursetting{3.46} & \best{57.72} & 6.10 & \oursetting{\second{13.14}} & \best{59.35} & \second{20.05} & 21.17 & 44.93 & 111.0M & \best{0.39} \\
			\ours (Ours)  & 
			\oursetting{\best{31.42}} & \oursetting{\best{16.51}} & \best{54.25} & \best{25.13} & \oursetting{\best{13.71}} & \best{48.07} & \best{20.68} & \oursetting{\best{18.54}} & \second{58.74} & \best{28.38} &  \best{30.11}  & \best{56.44} & 120.0M & 1.32 &
			\oursetting{\best{26.29}} & \oursetting{\best{10.92}} & \second{56.10} & \best{17.09} & \oursetting{\best{4.88}} & \second{57.53} & \best{8.48} & \oursetting{\best{13.94}} & \second{55.39} & \best{21.39}  & \second{22.39} & \second{47.54} & 115.0M & \second{0.65} \\
			\bottomrule
		\end{tabular}
	} \\
 \vspace{-1.0em}
	\caption{\textbf{Panoptic Scene Completion.} On both Semantic KITTI~\cite{semkitti} (val) and SSCBench-KITTI360~\cite{sscbench} (test), our method \ours~outperforms all baselines across almost all metrics, in particular, \textit{All PQ$^\dagger$}. 
    * denotes our own re-implementation of SCPNet.
 }
	\label{tab:psc_quantitative}
\end{table*}

\condenseparagraph{Masks matching loss.}
For each subnet $i$, we match the output masks $Y_i = \{(m_k, c_k)\}_{k=1}^{K}$ to the ground truth masks $\bar{Y}_i = \{(\bar{m}_{\bar{k}}, \bar{c}_{\bar{k}})\}_{\bar{k}=1}^{\bar{K} }$, using  the Hungarian matching as in~\cite{maskformer, maskpls} to learn an optimal mapping $\bar{\sigma}$ by minimizing the assignment map $\assigncost(\cdot, \cdot) \in \BR^{K \times \bar{K}} $. The latter is defined as $\assigncost_{k, \bar{k}} = -p_k (\bar{c}_{\bar{k}}) + \SL_{\rm mask}$ with
\begin{equation}
   \SL_{\rm mask} = \losscoef_{\rm dice}{\rm dice}(m_k,\bar{m}_{\bar{k}}) + \losscoef_{\rm bce}{\rm BCE}(m_k,\bar{m}_{\bar{k}}).
\end{equation}
We set $K$ always greater than $\bar{K}$ the number of ground truth masks. Predicted masks without ground truth are mapped to a generic $\varnothing$ class.
For the $k\text{-th}$ mask of $Y_i$  matched to the $\bar{\sigma}(k)\text{-th}$ ground truth mask, the loss is
\begin{equation}
    \SL_{\rm matched} = \sum^{\bar{K}}_{k=1}\losscoef_{\rm CE}{\rm CE}(c_k, \bar{c}_{\bar{\sigma}(k)}) + \SL_{\rm mask}.
\end{equation}
The $K {-} \bar{K}$ unmatched predicted masks are optimized to predict $\varnothing$ class $\SL_{\rm unmatched} = \sum^K_{k=\bar{K}+1}\losscoef_\varnothing{\rm CE}(c_k, \varnothing)$.
where $\losscoef_\varnothing$ is set to 0.1 as in ~\cite{maskformer}. $\losscoef_{\rm dice}$ and $\losscoef_{\rm bce}$ are emperically set to 1 and 40. 
We further apply auxiliary mask matching losses and $\SL'_{{\rm sem}}$ on the PSC outputs of intermediate scales, \ie, $\{(m^{1:\ell}, p^{1:\ell})\}, \ell{\neq}1$.

\section{Experiments}
\label{sec:experiments}

We evaluate \ours on both panoptic scene completion and uncertainty estimation, while also reporting the subsidiary SSC metrics. As there are \textit{no urban PSC datasets and baselines}, we produce our best effort to extend existing SSC datasets and baselines for fair evaluation.\\
\condenseparagraph{Datasets.} To evaluate PSC, we extend three large-scale urban LiDAR SSC datasets: Semantic KITTI, SSCBench-KITTI360 and Robo3D.
\textbf{Semantic KITTI}~\cite{semkitti} has 64-layer LiDAR scans voxelized into 256x256x32 grids of 0.2m voxels. We follow the standard train/val split~\cite{lmscnet, scpnet}, leading to 3834/815 grids. 
\textbf{SSCBench-KITTI360}~\cite{sscbench} is a very recent SSC benchmark derived from KITTI-360~\cite{kitti360} with urban scans encoded as in Semantic KITTI. We follow the standard train/val/test splits of 8487/1812/2566 grids. 
\textbf{Robo3D}~\cite{robo3d} {is a new 
robustness benchmark, extending popular urban datasets~\cite{geiger2013vision,semkitti,caesar2020nuscenes,sun2020scalability} by modifying point cloud inputs with various type and intensity of corruptions (\eg, fog, motion blur, 
\etc). We use corrupted input point clouds from the SemanticKITTI-C set of Robo3D to evaluate robustness to Out Of Distribution (OOD) effects.}

To extract pseudo panoptic labels from semantic grids, we cluster \textit{things} instances from ad-hoc classes using DBSCAN~\cite{ester1996density,chen2022approach} with a distance of $\epsilon=1$ and groups with $\texttt{MinPts}=8$. Following the original panoptic segmentation formulation~\cite{panopticsegmentation}, \textit{stuff} masks are made of voxels with ad-hoc classes.
For Semantic KITTI, labels cannot be generated on the hidden test set, so we evaluate on val. set only.

\condenseparagraph{PSC/SSC metrics.} We evaluate panoptic quality (PQ), segmentation quality (SQ) and recognition quality (RQ) following~\cite{panopticsegmentation} on the complete scene. 
Due to the difficulty of the PSC task, most masks have low IoU \wrt ground truth. Hence, we note that the over-penalization effect of stuff classes described in~\cite{porzi2019seamless} is amplified for PSC. Thus, we also evaluate PQ$^\dagger$, as in~\cite{porzi2019seamless}, removing the \mbox{${>}0.5{-}\text{IoU}$ rule} for stuff classes. We also complement our PSC study with subsidiary SSC metrics, \ie, mean IoU (mIoU).

\condenseparagraph{Uncertainty metrics.} Following~\cite{packedensemble}, we employ the maximum softmax probability as a measure of model confidence. Consistent with the established practices~\cite{deepensemble, mimo, packedensemble}, we assess the model predictive uncertainty by evaluating its calibration~\cite{guo2017calibration} using the Expected Calibration Error~(ECE) and Negative Log Likelihood~(NLL).
Notably, we distinguish between two forms of uncertainty: \textit{voxel uncertainty} and \textit{instance uncertainty}. The former is derived voxel-wise from semantic completion outputs, and the latter mask-wise from class probability predictions. To account for the dominance of empty voxels within 3D scenes, we calculate voxel uncertainties by averaging the uncertainties for empty and non-empty voxels. In line with~\cite{panopticsegmentation}, the label of predicted masks are assigned by finding the matched ground truth masks with \mbox{${>}0.5{-}\text{IoU}$ rule}. Unmatched masks are classified under a `dustbin' category.

\condenseparagraph{Training details.} We train \ours for 30 epochs on Semantic KITTI and 20 epochs on SSCBench-KITTI360, both using AdamW~\cite{adamW} optimizer and batch size of 2. The learning rate is 1e-4, unchanged for Semantic KITTI but divided by 10 at epoch 15 on SSCBench-KITTI360. 
We apply random rotations in $[-30^\circ, 30^\circ]$ on Semantic KITTI and in $[-10^\circ, 10^\circ]$ on SSCBench-KITTI360, random crop to reduce the scene size to 80\% along both the x and y axes, and random translations of $\pm0.6\text{m}$ on x/y axes and $\pm0.4\text{m}$ on z axis. Additionally, on SemanticKITTI, we found it beneficial to use output point embeddings from a pretrained 3D semantic segmentation method~\cite{waffleiron} as input point features. Unless otherwise mentioned, \ours refers to the optimal number of subnets, which is $M=3$ for Semantic KITTI and Robo3D, and $M=2$ for SSCBench-KITTI360. This choice of subnets is justified in~\cref{tab:varied_n_subnets}.
\begin{figure*}
	\centering
	\setlength{\fboxsep}{0.008pt}
	\setlength{\fboxrule}{0pt}
	\newcommand{\tmpframe}[1]{\fbox{#1}}
	\newcommand{\im}[6]{\stackinset{r}{-0.0cm}{b}{0pt}{\tmpframe{\adjincludegraphics[width=#6\linewidth, trim={#2\width} {#3\height} {#4\width} {#5\height}, clip]{#1}}}{\adjincludegraphics[width=\linewidth, trim={.2\width} {.2\height} {.2\width} {.2\height}, clip]{#1}}}
	\newcommand{\imfirst}[1]{\im{#1}{0.28}{0.29}{0.53}{0.53}{0.49}}
	\newcommand{\imsecond}[1]{\im{#1}{0.3}{0.32}{0.5}{0.48}{0.49}}
	\newcommand{\imthird}[1]{\im{#1}{0.44}{0.47}{0.40}{0.38}{0.50}}
	\newcommand{\imforth}[1]{\im{#1}{0.5}{0.57}{0.35}{0.3}{0.5}}

		\resizebox{0.9\linewidth}{!}{
			\centering
			\newcolumntype{P}[1]{>{\centering\arraybackslash}m{#1}}
			\setlength{\tabcolsep}{0.001\textwidth}
			\renewcommand{\arraystretch}{0.8}
			\footnotesize
			\begin{tabular}{cP{0.15\textwidth} P{0.15\textwidth} P{0.15\textwidth} P{0.15\textwidth} P{0.15\textwidth} P{0.15\textwidth}}		
				&Input & LMSCNet + MaskPLS & JS3CNet + MaskPLS & SCPNet* + MaskPLS & \ours~(ours) & ground truth 
				\\
				\multirow{1}{*}{\rotatebox{90}{\textbf{Semantic KITTI} (val set)\hspace{-0.3em}}}
				 &
				\imfirst{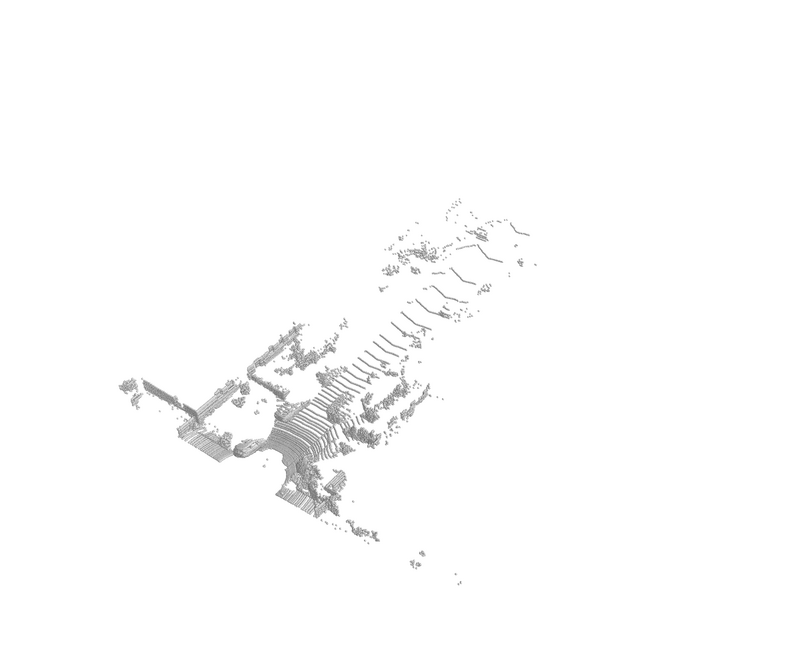} &
				\imfirst{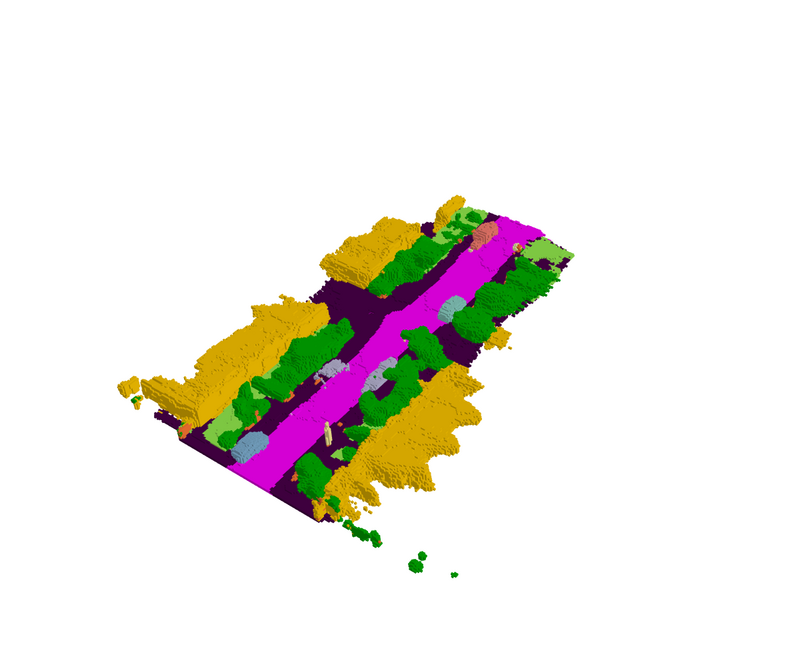} &
				\imfirst{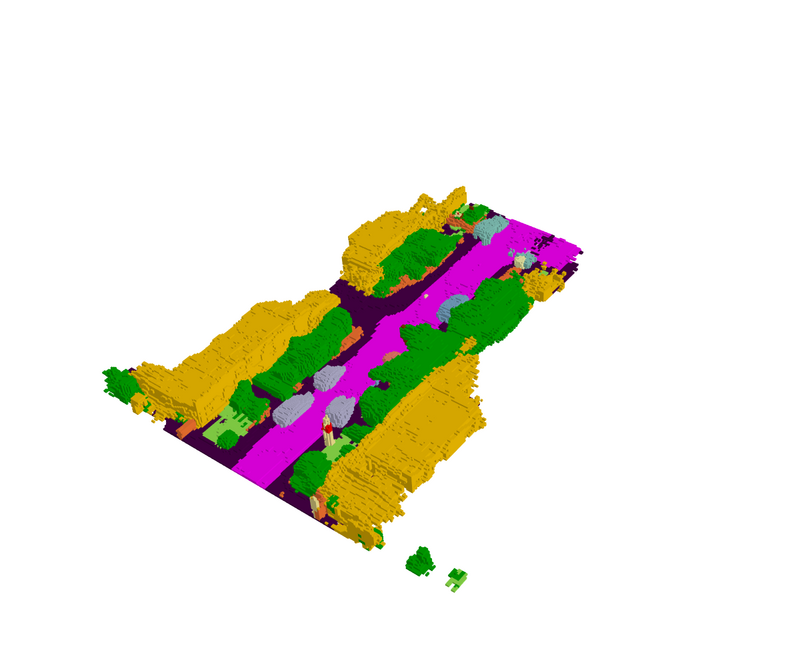} &
				\imfirst{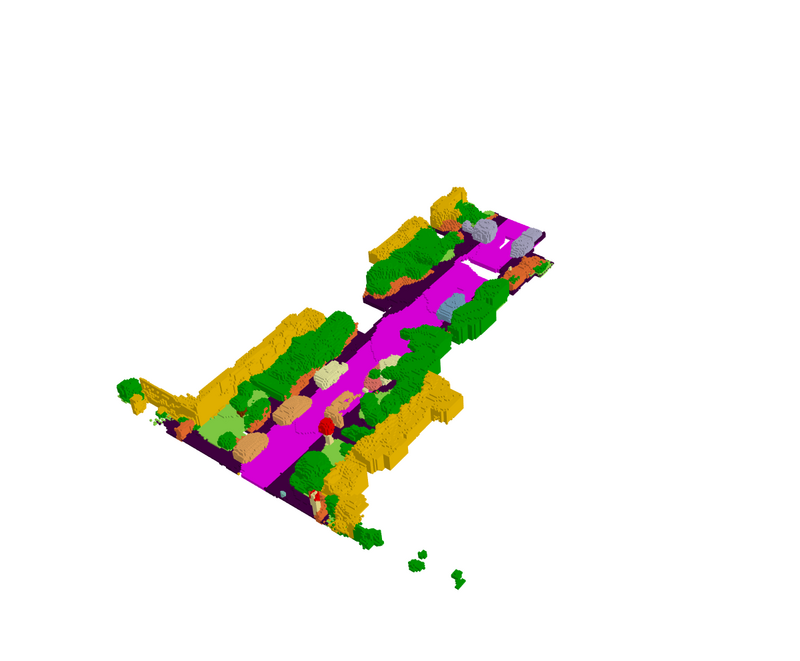} &
				\imfirst{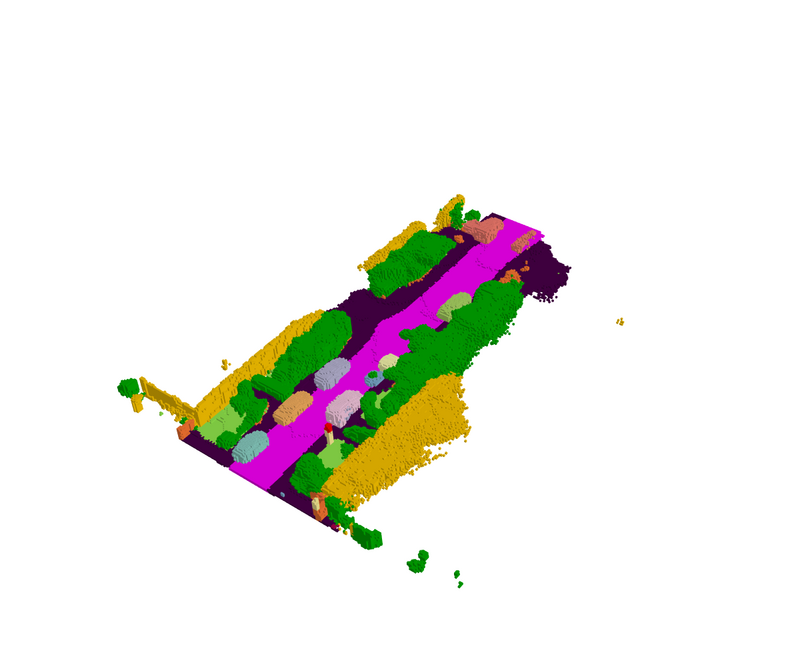} &
				\imfirst{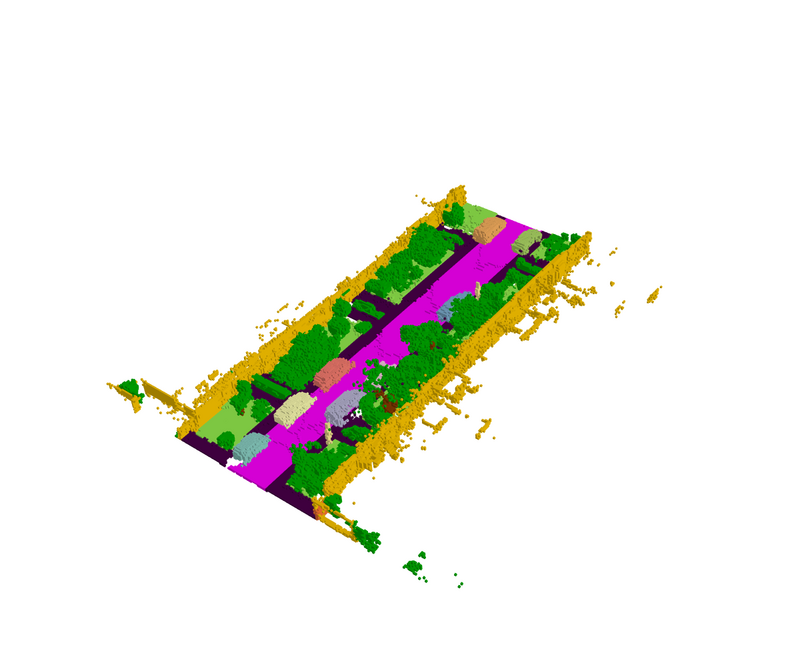}
				\\[-0.3em]

				&
				\imsecond{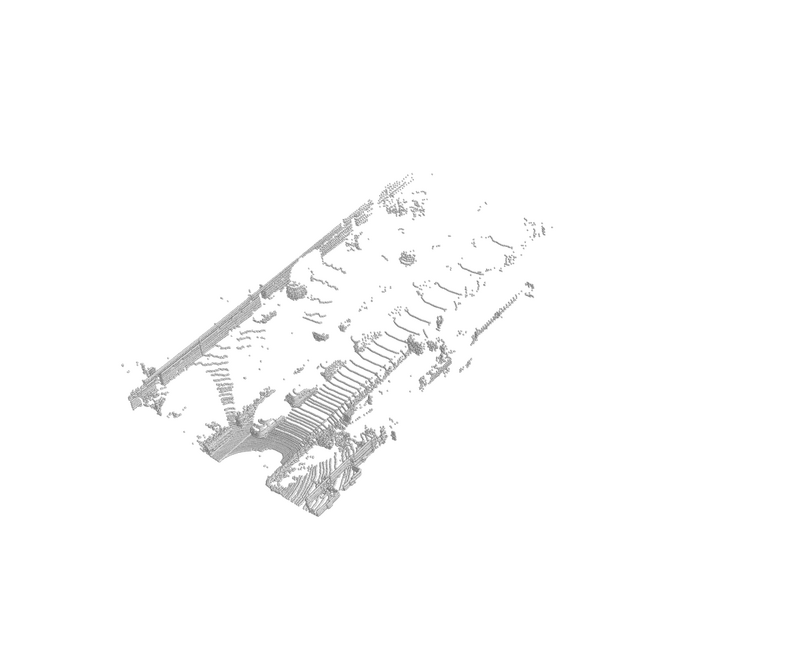} &
				\imsecond{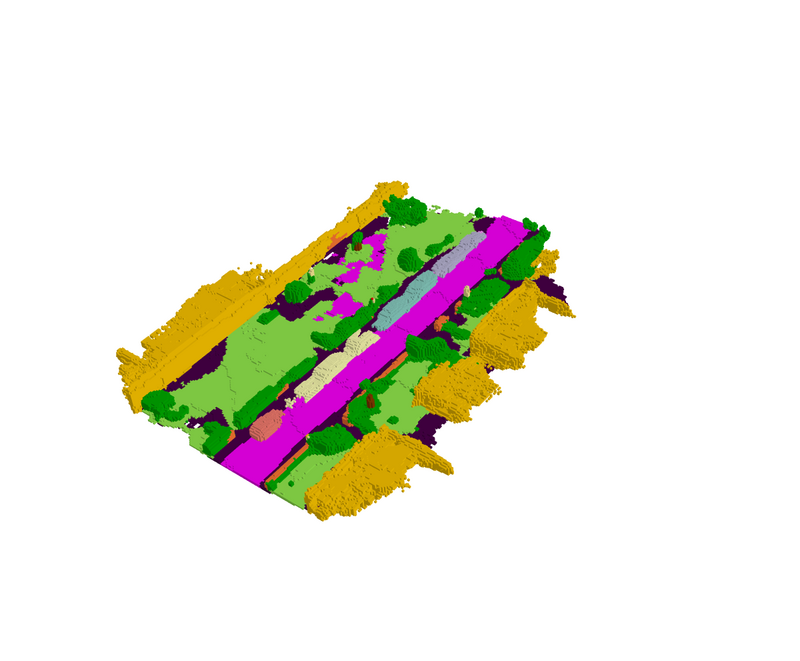} &
				\imsecond{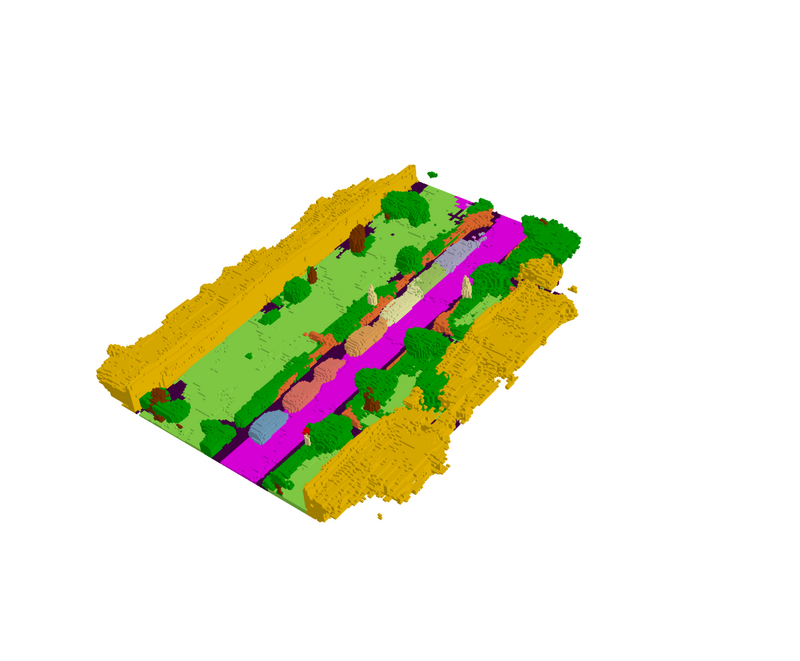} &
				\imsecond{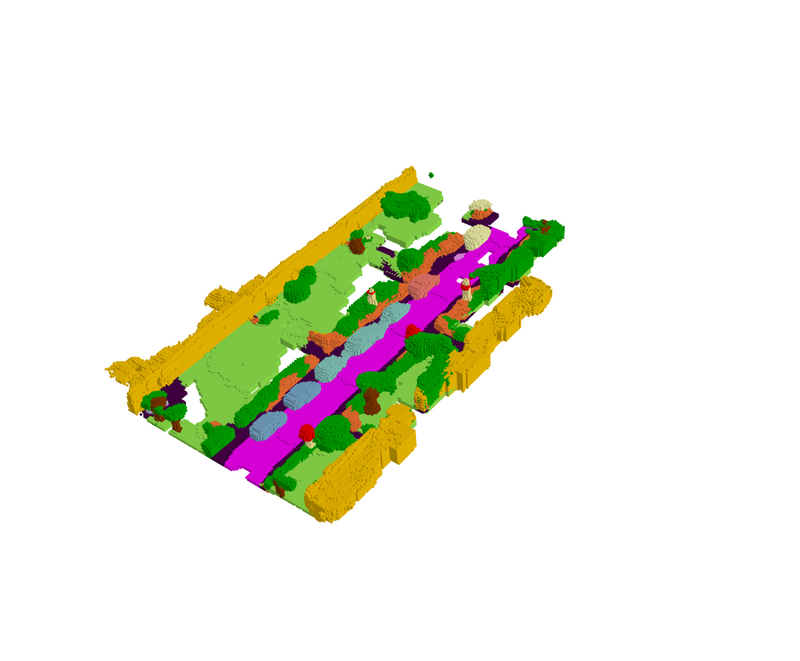} &
				\imsecond{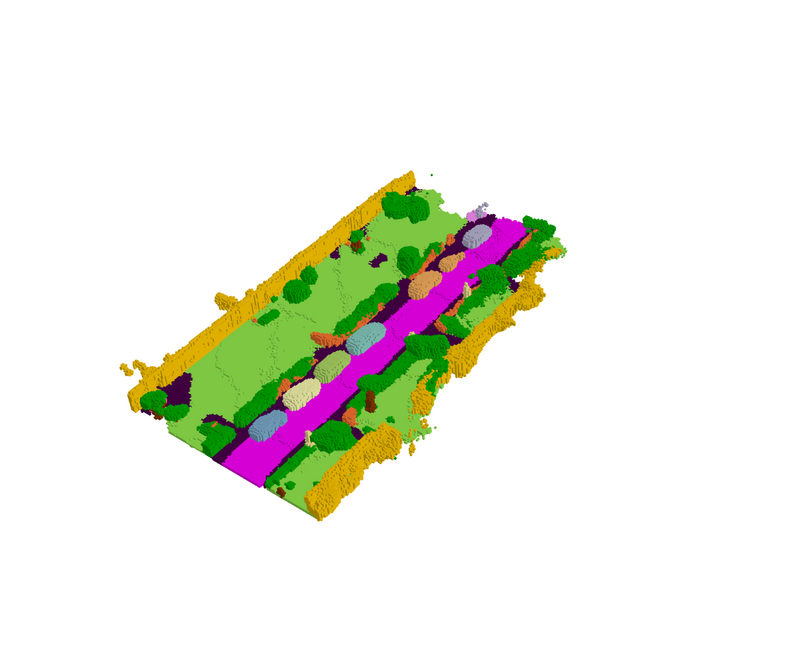} &
				\imsecond{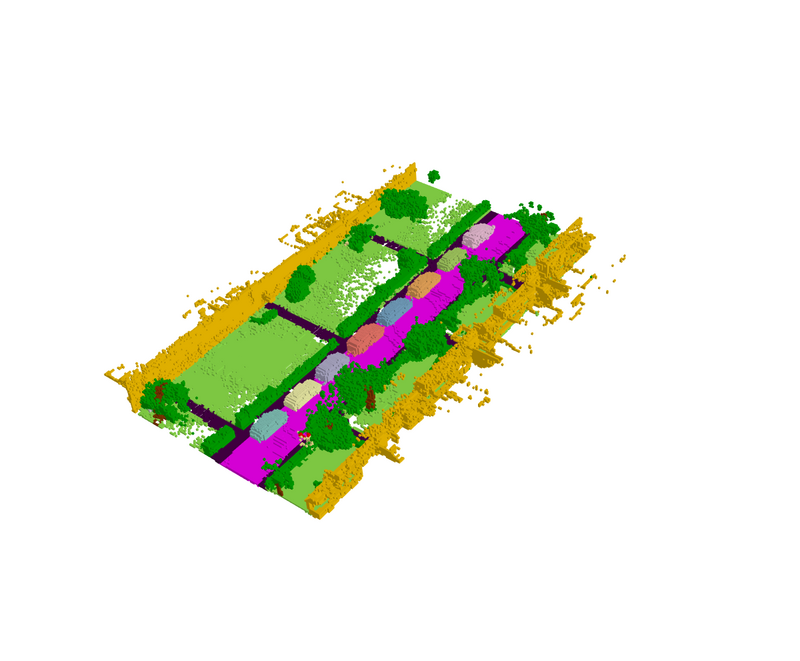}
				\\[-0.3em]
				\midrule
				
				\multirow{1}{*}{\rotatebox{90}{\textbf{SSCBench-KITTI360} (val set)\hspace{-2.3em}}}
				& \imthird{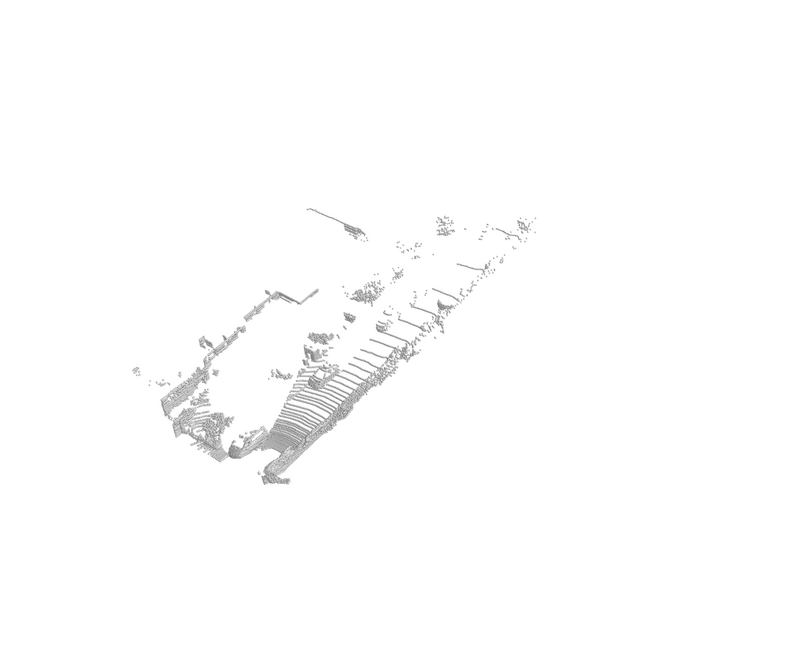} 
				& \imthird{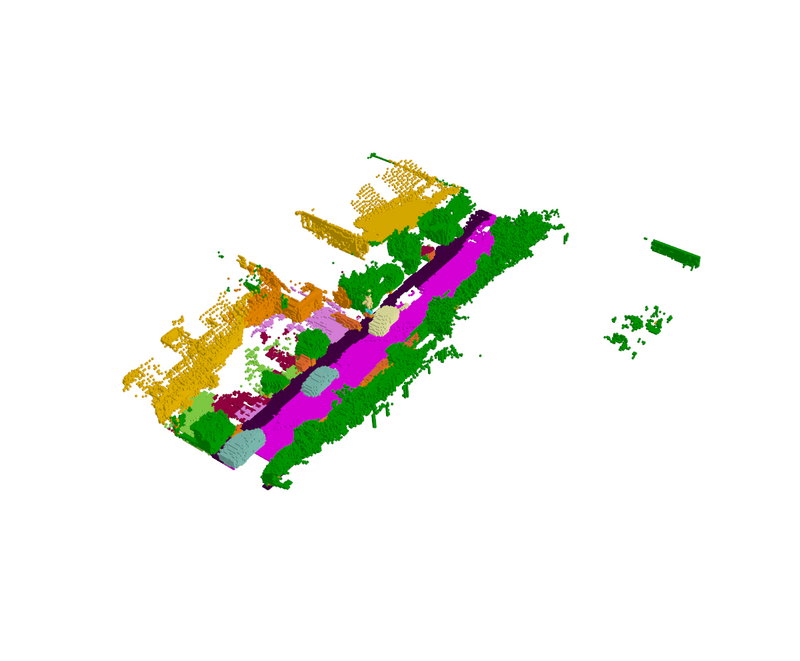} 
				& \imthird{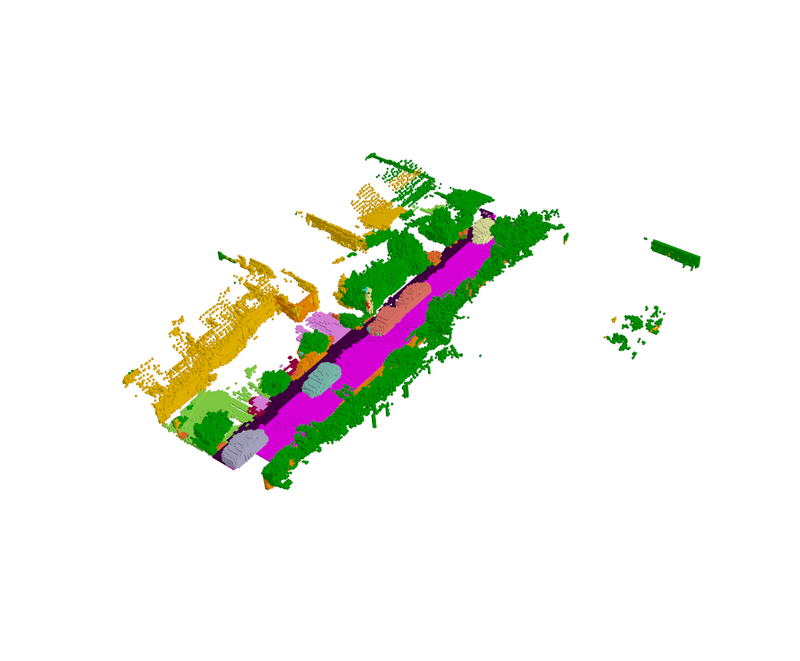} 
				& \imthird{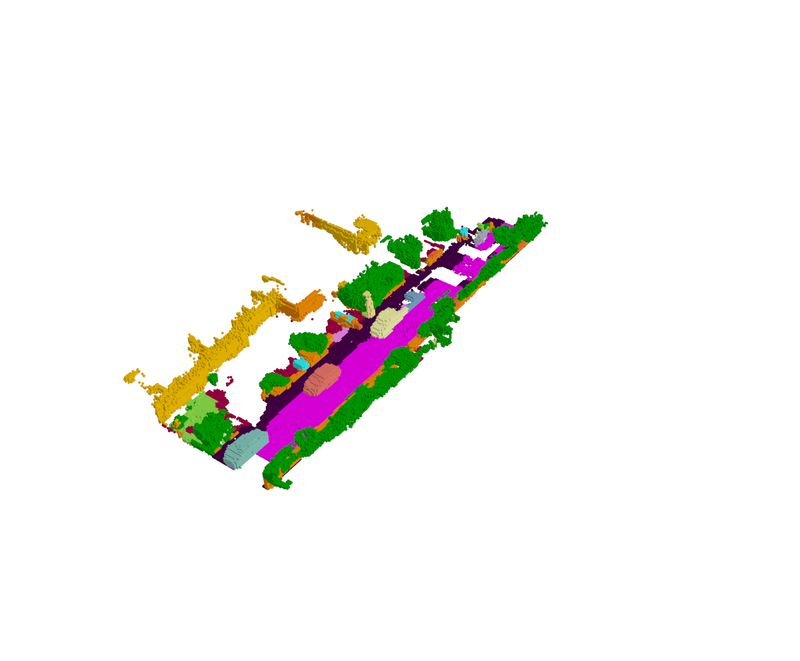} 
				& \imthird{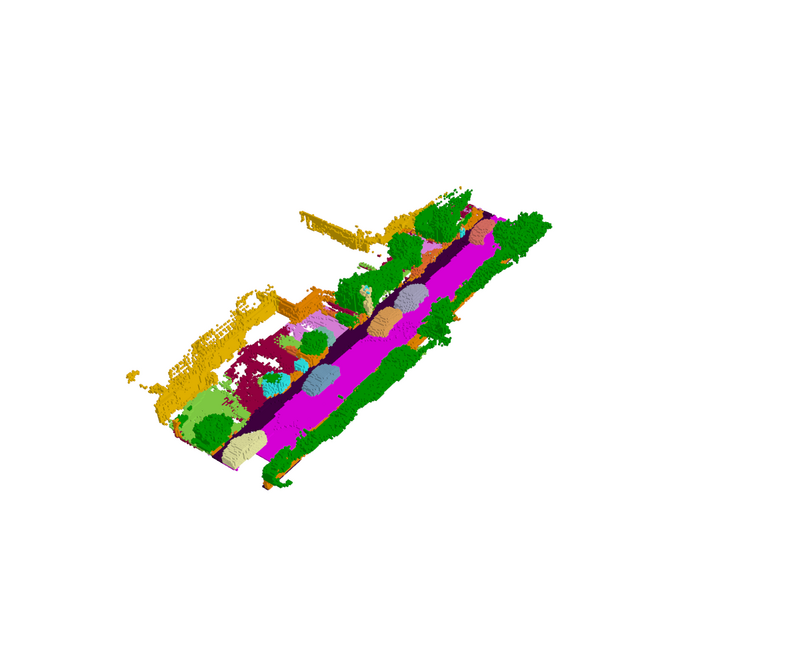} 
				& \imthird{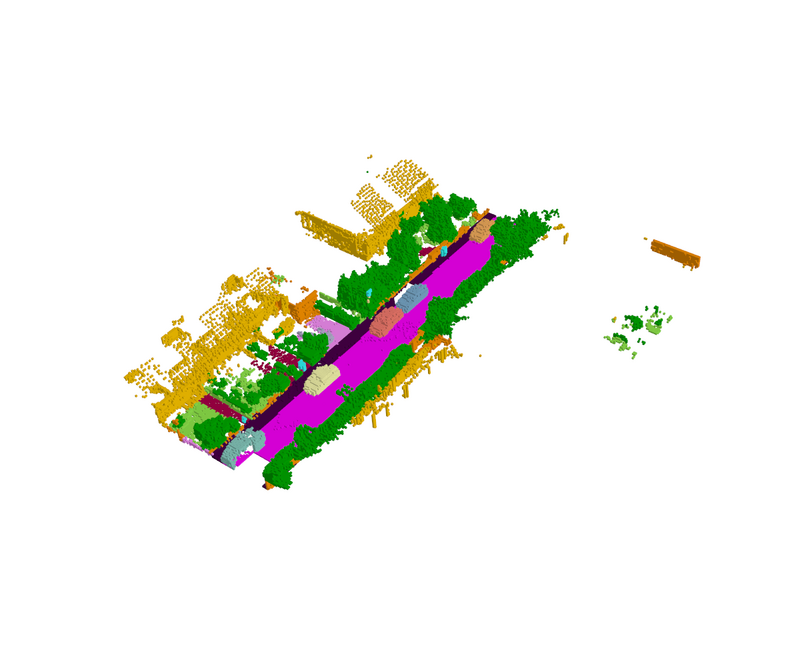}
				\\[-0.3em]

				& \imforth{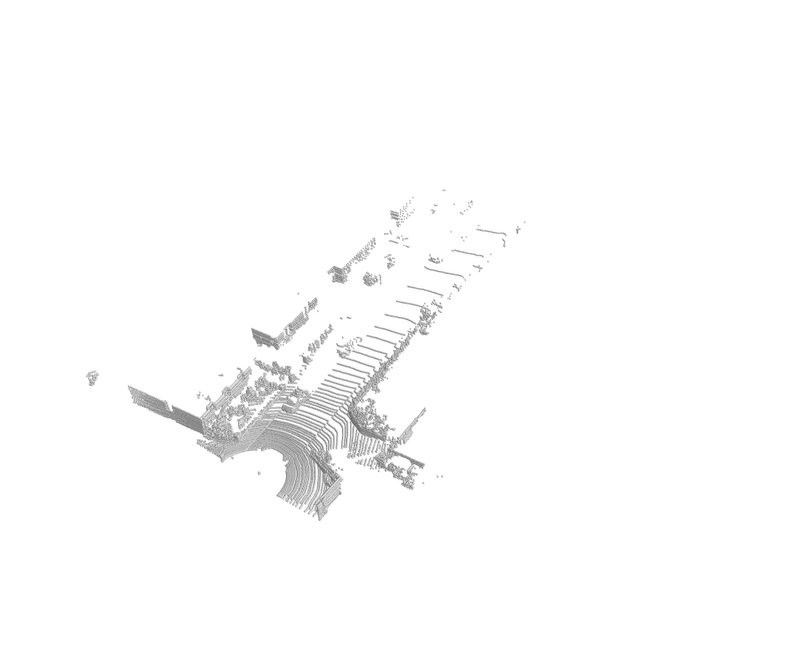} 
				& \imforth{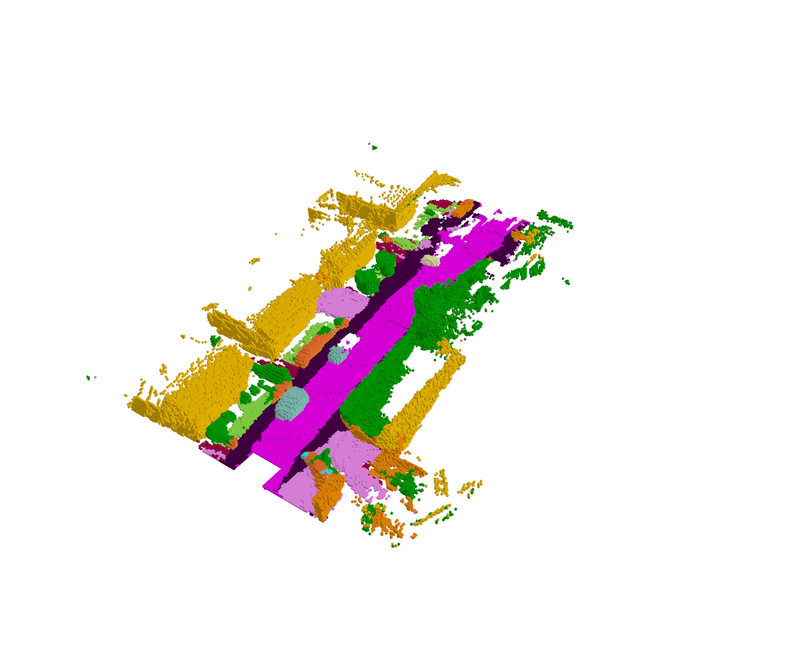} 
				& \imforth{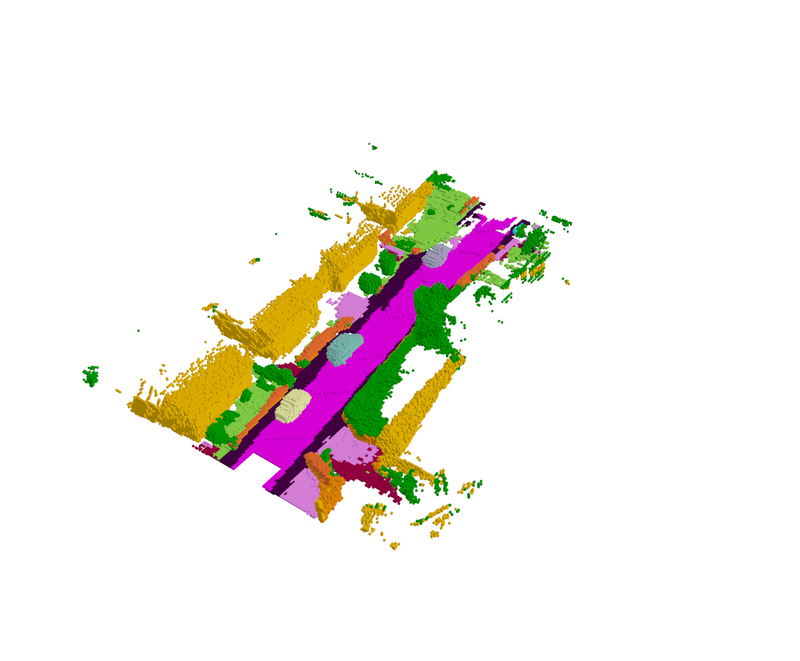} 
				& \imforth{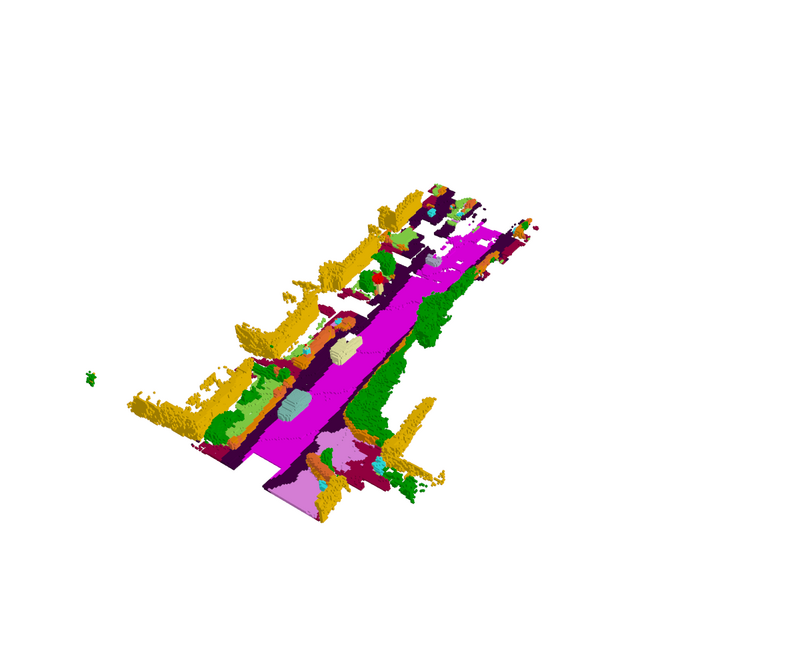} 
				& \imforth{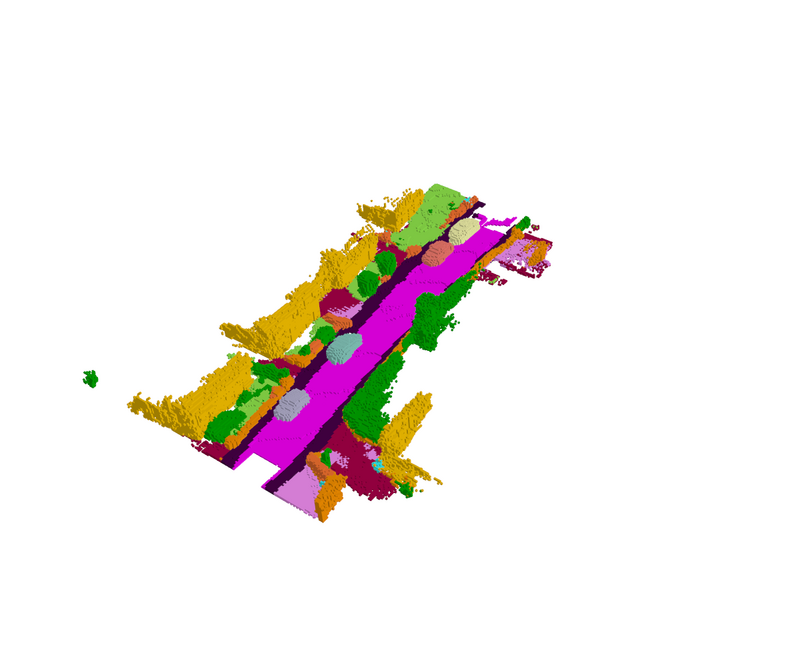} 
				& \imforth{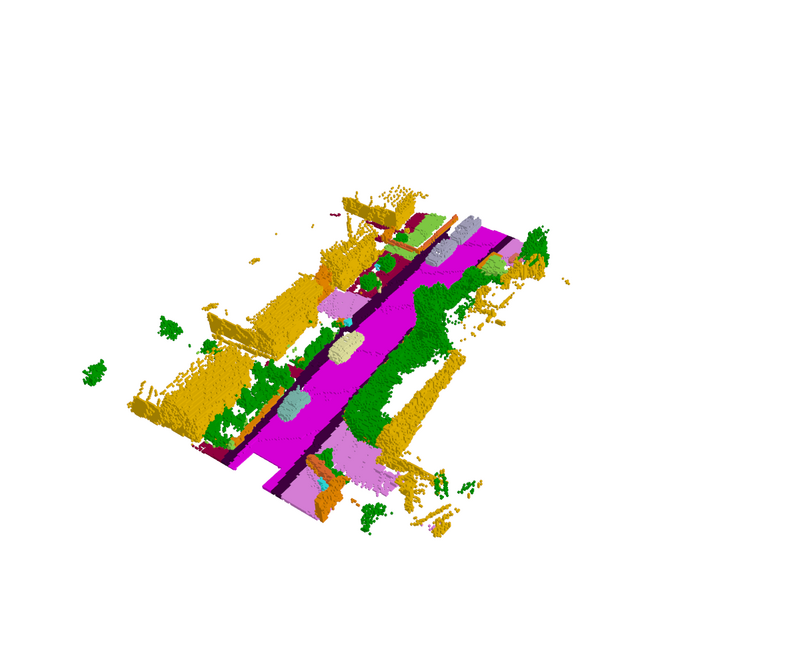}
				\\[-0.3em]

			\end{tabular}
		}\\
		{				
			\tiny
			\textcolor{bicycle}{$\blacksquare$}bicycle~%
			\textcolor{car}{$\blacksquare$}car~%
			\textcolor{motorcycle}{$\blacksquare$}motorcycle~%
			\textcolor{truck}{$\blacksquare$}truck~%
			\textcolor{other-vehicle}{$\blacksquare$}other vehicle~%
			\textcolor{person}{$\blacksquare$}person~%
			\textcolor{bicyclist}{$\blacksquare$}bicyclist~%
			\textcolor{motorcyclist}{$\blacksquare$}motorcyclist~%
			\textcolor{road}{$\blacksquare$}road~%
			\textcolor{parking}{$\blacksquare$}parking~%
			\textcolor{sidewalk}{$\blacksquare$}sidewalk~%
			\textcolor{other-ground}{$\blacksquare$}other ground~%
			\textcolor{building}{$\blacksquare$}building~%
			\textcolor{fence}{$\blacksquare$}fence~%
			\textcolor{vegetation}{$\blacksquare$}vegetation~%
			\textcolor{trunk}{$\blacksquare$}trunk~%
			\textcolor{terrain}{$\blacksquare$}terrain~%
			\textcolor{pole}{$\blacksquare$}pole~%
			\textcolor{traffic-sign}{$\blacksquare$}traffic sign~%
			\textcolor{other-struct}{$\blacksquare$}other structure~
			\textcolor{other-object}{$\blacksquare$}other object			
		}%
\caption{\textbf{Qualitative Panoptic Scene Completion}. We report PSC outputs for all baselines of~\cref{tab:psc_quantitative}. \ours{} shows better instance separation, with stronger instance shapes and scene structure, with fewer holes.
}
\label{fig:qualitative}
\end{figure*}

\subsection{Panoptic Scene Completion}
\label{sec:exp_psc}
To evaluate PSC, we first establish baselines for this new task, and then report results on the aforementioned datasets.\\

\condenseparagraph{Baselines.} We combine existing SSC methods with 3D panoptic segmentation. We select three SSC open-source methods: LMSCNet~\cite{lmscnet}, JS3CNet~\cite{js3cnet}, SCPNet~\cite{scpnet}, and also add SCPNet* --- our own reimplementation with much stronger performance.
For 3D panoptic segmentation, we use MaskPLS~\cite{maskpls}, well-suited for dense voxelized scene and the best open-source 3D panoptic segmentation to date. All baselines are retrained with their reported parameters.

We train the four PSC baselines using the SSC method to predict the complete semantic scene followed by the 3D panoptic segmentation method.

\condenseparagraph{Performance.} \cref{tab:psc_quantitative} compares \ours with the 4 baselines on Semantic KITTI and SSCBench-KITTI360. Our method is superior across all panoptic metrics (All, Things, Stuff) on both datasets. 
We see a major boost in All-PQ$^{\dagger}$/PQ of $+8.21$/$+5.62$ on Semantic KITTI and $+8.09$/$+3.45$ on SSCBench-KITTI360, due to our effective ensembling approach for PSC. Regarding inference time, PaSCo is only slower than LMSCNet+MaskPLS and SCPNet+MaskPLS but performs significantly better.
Additionally, \ours~outperforms baselines in individual metrics for both `things'/`stuff' categories, showing significant improvements in PQ with $+6.36$/$+5.08$ and $+0.9$/$+4.73$ on each dataset. On the subsidiary mIoU metric we perform on-par, being first on Semantic KITTI ($+2.22$) and 2nd in SSCBench-KITTI360 ($-0.27$).
Incidentally, we note that PSC and SSC metrics are not directly correlated since we improve the former drastically.

\cref{fig:qualitative} shows that our qualitative PSC results similarly show visual superiority. Overall, we observe that instances are much better separated by \ours compared to SCPNet* (our best competitor), with less holes in the geometry.

\subsection{Uncertainty estimation}
\label{sec:exp_uncertainty}
We further evaluate uncertainty as it correlates with model calibration~\cite{uncertaintyreview}, and is crucial for many applications.

\condenseparagraph{Baselines.} Using our architecture, we design three uncertainty estimation baselines based on state-of-the-art uncertainty literature. Each baseline provides multiple outputs, enabling similar computation of uncertainty to ours from the maximum softmax probability across inferences.
Test-Time Augmentation (TTA) is a classical strategy~\cite{tta} to improve robustness using multiple inferences of a unique network with input augmentations.
MC~Dropout~\cite{mcdropout} provides a bayesian approximation of the model uncertainty by randomly dropping activations (\ie, setting to 0) of neurons, applied with multiple inferences.
Finally, we report Deep Ensemble~\cite{deepensemble}, where duplicate networks solving the same task are trained independently and ensembled at test time for  better predictive uncertainty than a single network.

For fair comparison, all baselines use our architecture.
However, we note that in contrast to our approach, these baselines require more than one pass, either using multiple inferences for TTA and MC~Dropout or multiple networks for Deep Ensemble which translate in more parameters.\\
\condenseparagraph{Uncertainty estimation.} \cref{tab:uncertainty} reports uncertainties for all baselines using our architecture, as well as for \ours and \oursMIMO{1} which uses a single subnet. To ensure comparable performance, we set the number of inferences (for TTA and MC~Dropout) and number networks (for Deep Ensemble) equal to the number of subnets in \ours{} --- i.e., 3 on Semantic KITTI and 2 on SSCBench-KITTI360. Notably, baselines can only estimate voxel-wise uncertainties, for which we outperform by a large margin. Only the voxel ece of Deep Ensemble for Semantic KITTI is a close second (0.0428 vs 0.0426), though at the cost of ${\approx{}}3$ times our number of parameters and $3$ passes. 
Comparing \oursMIMO{1} and \ours highlights that our ensemble approach brings a clear boost on all metrics at a minor increase of number of parameters (111M vs 115M).

\begin{table}
	\centering
	\scriptsize
	\setlength{\tabcolsep}{0.004\linewidth}
	\resizebox{1.0\linewidth}{!}{
		\begin{tabular}{ l|cccc|cc|c|ccc }
			\multicolumn{10}{c}{\textbf{Semantic KITTI} (val set)}\\
			\toprule
			method  & ins ece$\downarrow$ & ins nll$\downarrow$ &  voxel ece$\downarrow$ & voxel nll$\downarrow$ & All PQ$^{\dagger}$$\uparrow$ & All PQ$\uparrow$ & mIoU$\uparrow$ &  Params$\downarrow$   &  Passes $\downarrow$ & Time(s) $\downarrow$ \\
			\midrule
			TTA & -	& - & 0.0456 & 0.7224 & - & - & 28.84 &  \textbf{111M}  & \second{3} & 1.78 \\
			MC~Dropout~\cite{mcdropout} & - & - & 0.0472 & 0.7437 & - & - & 28.82 &  \textbf{111M} & \second{3} & 1.70 \\
			Deep Ensemble~\cite{deepensemble} & - & - & \second{0.0428} & \second{0.6993} & -&  - & \second{30.10} &  333M   & \second{3} & 1.69 \\
			\oursMIMO{1} &  \second{0.6181} & \second{4.6559} & 0.0610 & 0.8250  & \second{26.49} & \second{15.36} & 28.22 &  \textbf{111M} &  \textbf{1} & \best{0.67} \\
			\ours~(ours) &  \textbf{0.4922} & \textbf{3.9155} & \best{0.0426} & \textbf{0.5835} & \textbf{31.42} & \textbf{16.51} & \textbf{30.11} & \second{120M}  & \textbf{1}  & \second{1.32}\\
			\bottomrule
		\end{tabular}
	}\\[.3em]
	\resizebox{1.0\linewidth}{!}{
		\begin{tabular}{l|cccc|cc|c|ccc }
			\multicolumn{10}{c}{\textbf{SSCBench-KITTI360} (test set)}\\
			\toprule
			method  & ins ece$\downarrow$ & ins nll$\downarrow$ &  voxel ece$\downarrow$ & voxel nll$\downarrow$  &  All PQ$^{\dagger}$$\uparrow$ & All PQ$\uparrow$ & mIoU$\uparrow$ &   Params$\downarrow$ &  Passes $\downarrow$ &  Time(s) $\downarrow$ \\
			\midrule
			TTA &  - & - & 0.1580	& 2.1282 & - & - & 21.78	& \textbf{111M}  & \second{2} & 0.85\\
			MC~Dropout~\cite{mcdropout}  & - & - &  0.1548 & 2.0737 & - & - & 21.73 &  \textbf{111M} & 2 & 0.79 \\
			Deep Ensemble~\cite{deepensemble} &  -	& - & \second{0.1540} &	\second{2.0653} & - & - & \best{22.51}  & 222M   & \second{2} & 0.90 \\
			\oursMIMO{1} &  \second{0.7899} &   \second{5.4405} &  0.1749 &  2.3556 & \second{19.53} & \second{9.91} & 21.17 & \best{111M} & \best{1} & \best{0.38} \\
			\ours~(ours) & \best{0.6015} & \best{4.1454} & \best{0.1348} & \best{1.6112} & \best{26.29} &  \best{10.92} & \second{22.39} &  \second{115M} & \best{1} & \second{0.65} \\
			\bottomrule
		\end{tabular}
	}
	\vspace{-1.0em}
	\caption{\textbf{Uncertainty evaluation.} We evaluate uncertainty on Semantic KITTI (top) and SSCBench-KITTI360 (bottom). Baselines only produce voxel uncertainty (`voxel ece,' `voxel nll') which we outperform while also estimating PSC uncertainty (All PQ/PQ$^\dagger$). 
	}
	\label{tab:uncertainty}
\end{table}

\begin{figure}
	\begin{subfigure}{\linewidth}		
		\centering
		\footnotesize
		\renewcommand{\arraystretch}{0.0}
		\setlength{\tabcolsep}{0.003\textwidth}
		\newcolumntype{P}[1]{>{\centering\arraybackslash}m{#1}}
		\begin{tabular}{ P{0.03\textwidth} P{0.21\textwidth} P{0.03\textwidth} P{0.21\textwidth} P{0.21\textwidth} P{0.21\textwidth}}
			\\
			& Input & & Panoptic & Voxel unc. &  Ins. unc.
			\\
			\multirow{1}{*}{\rotatebox{90}{\textbf{Semantic KITTI} (val. set)\hspace{-2.8em}}} & 
			\multirow{1}{*}{\adjincludegraphics[width=\linewidth, trim={.2\width} {.2\height} {.2\width} {.2\height}, clip]{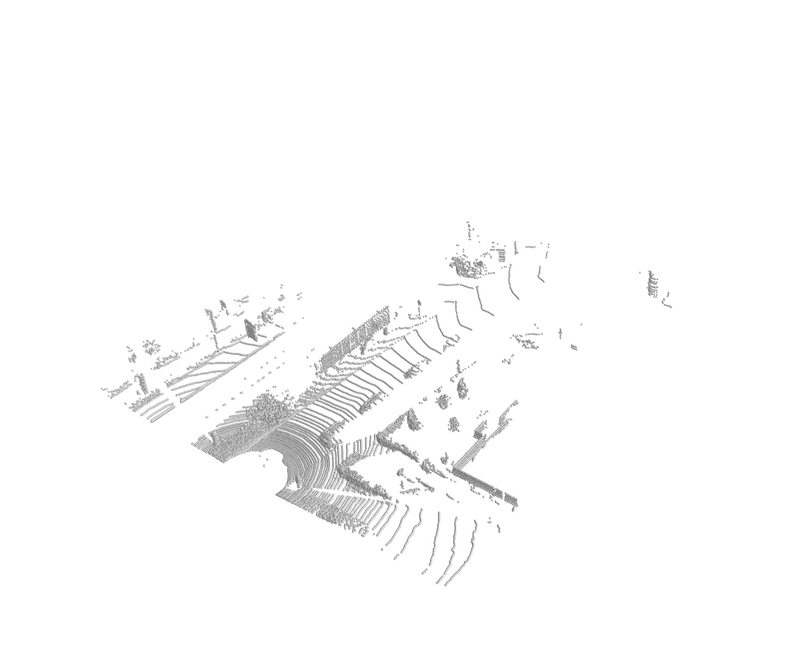}} & 
			\rotatebox{90}{\oursMIMO{1}} &
			\adjincludegraphics[width=\linewidth, trim={.2\width} {.2\height} {.2\width} {.2\height}, clip]{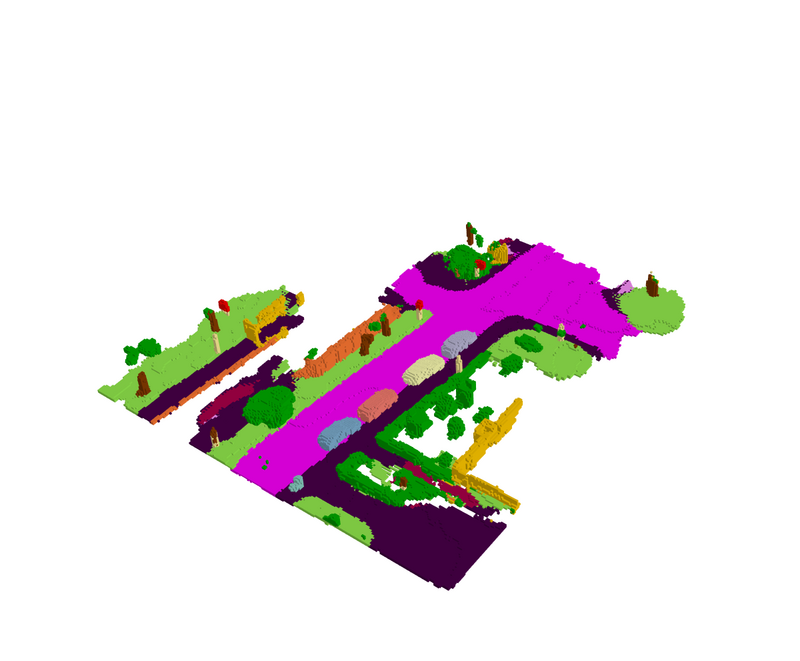} &
			\adjincludegraphics[width=\linewidth, trim={.2\width} {.2\height} {.2\width} {.2\height}, clip]{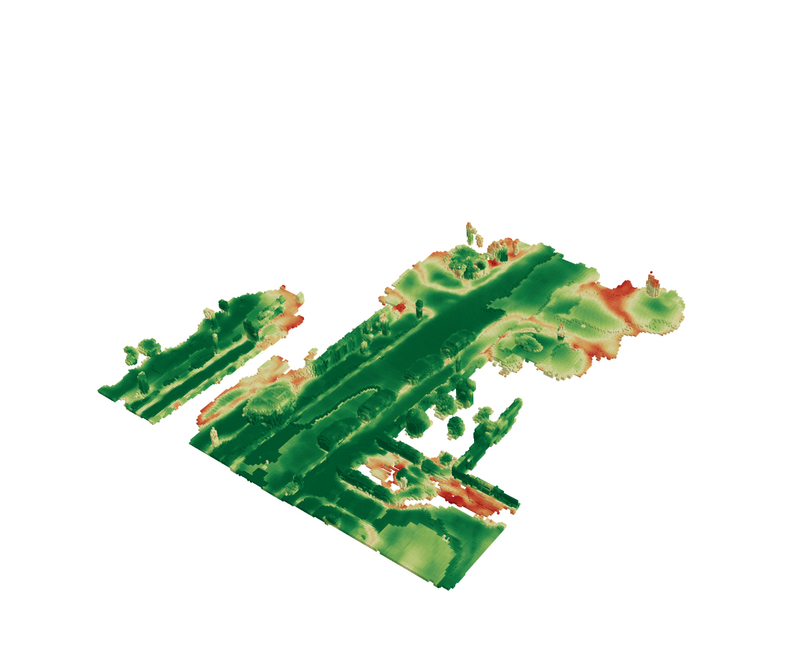} &
			\adjincludegraphics[width=\linewidth, trim={.2\width} {.2\height} {.2\width} {.2\height}, clip]{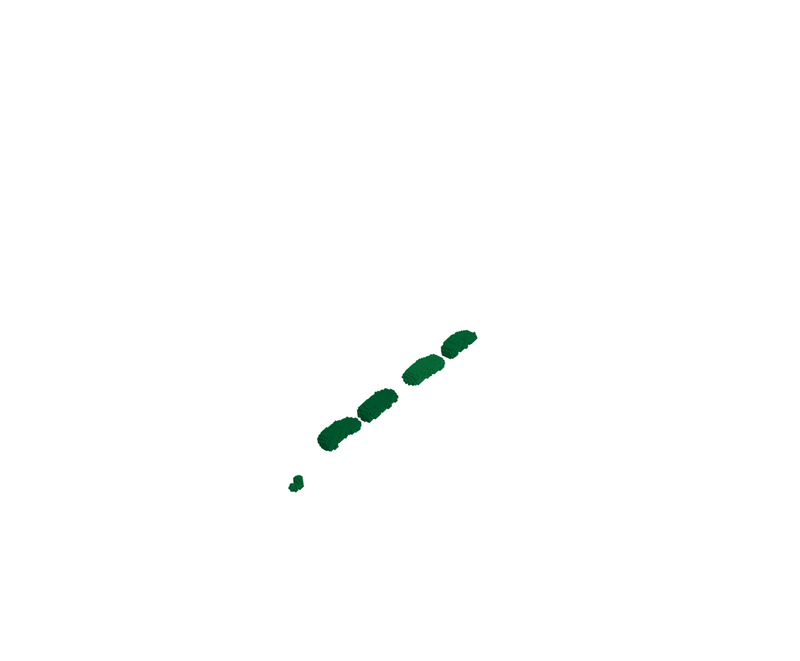} \\
			& 	
			& \rotatebox{90}{\ours} &
			\adjincludegraphics[width=\linewidth, trim={.2\width} {.2\height} {.2\width} {.2\height}, clip]{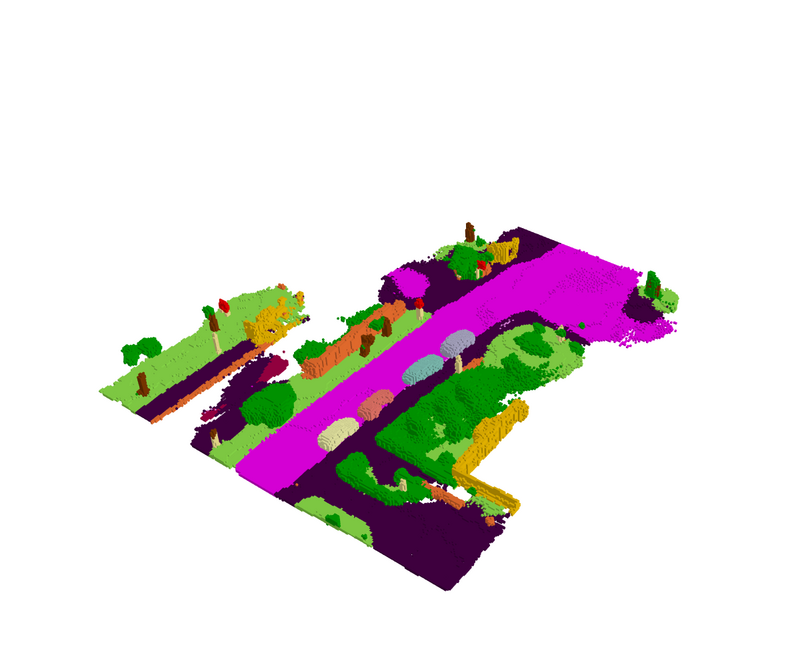} &
			\adjincludegraphics[width=\linewidth, trim={.2\width} {.2\height} {.2\width} {.2\height}, clip]{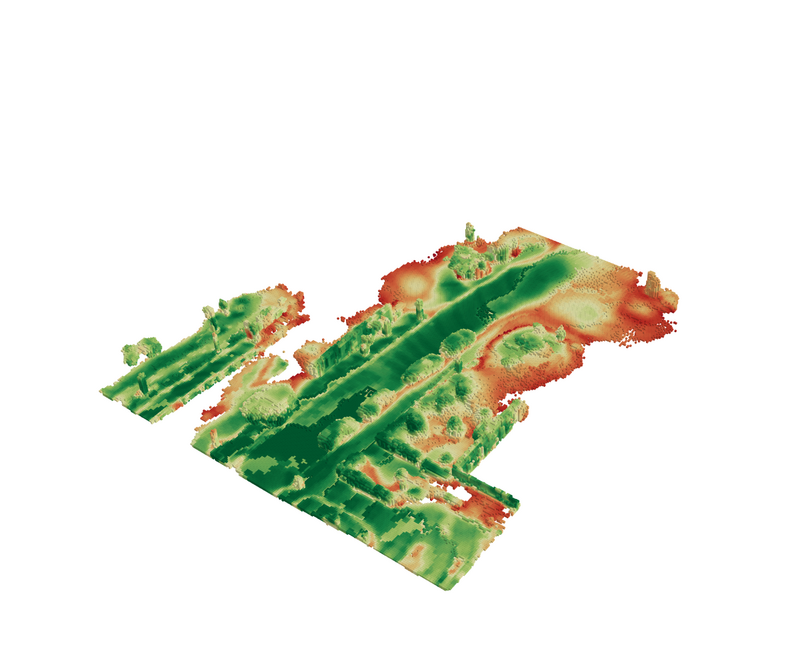} &
			\adjincludegraphics[width=\linewidth, trim={.2\width} {.2\height} {.2\width} {.2\height}, clip]{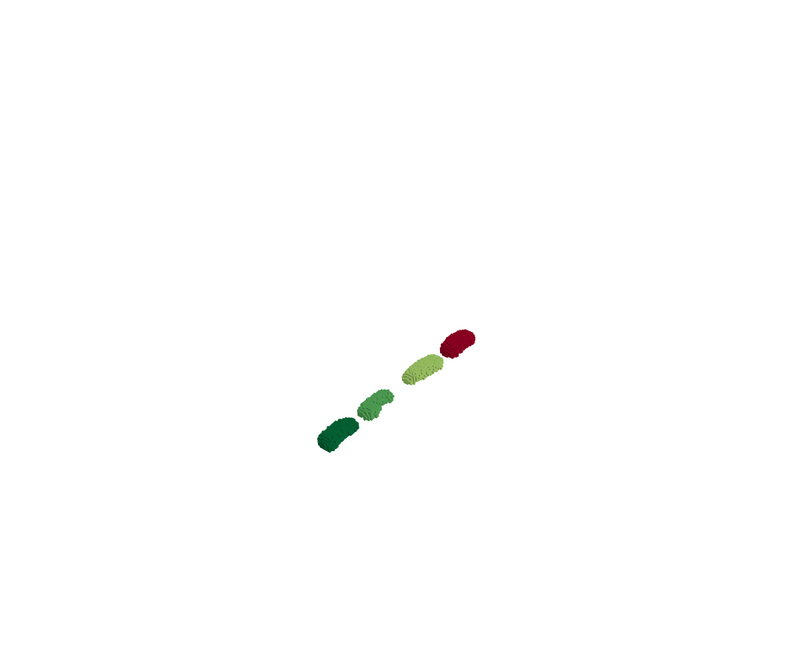} 
			\\
			\midrule
			\\
			\multirow{1}{*}{\rotatebox{90}{\textbf{SSCBench-KITTI360} (val. set)\hspace{-2.8em}}} & 
			\multirow{1}{*}{\adjincludegraphics[width=\linewidth, trim={.2\width} {.2\height} {.2\width} {.2\height}, clip]{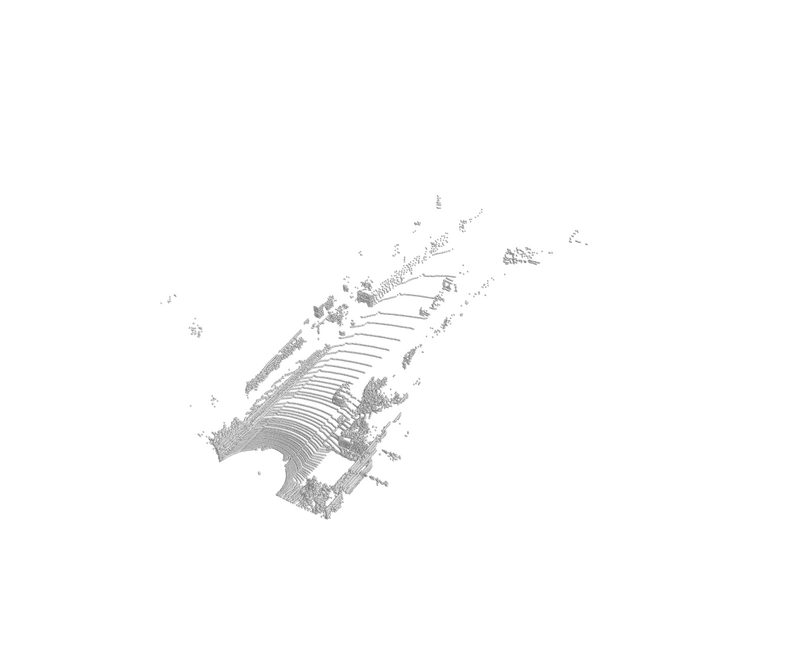}} & 
			\rotatebox{90}{\oursMIMO{1}} &
			\adjincludegraphics[width=\linewidth, trim={.2\width} {.2\height} {.2\width} {.2\height}, clip]{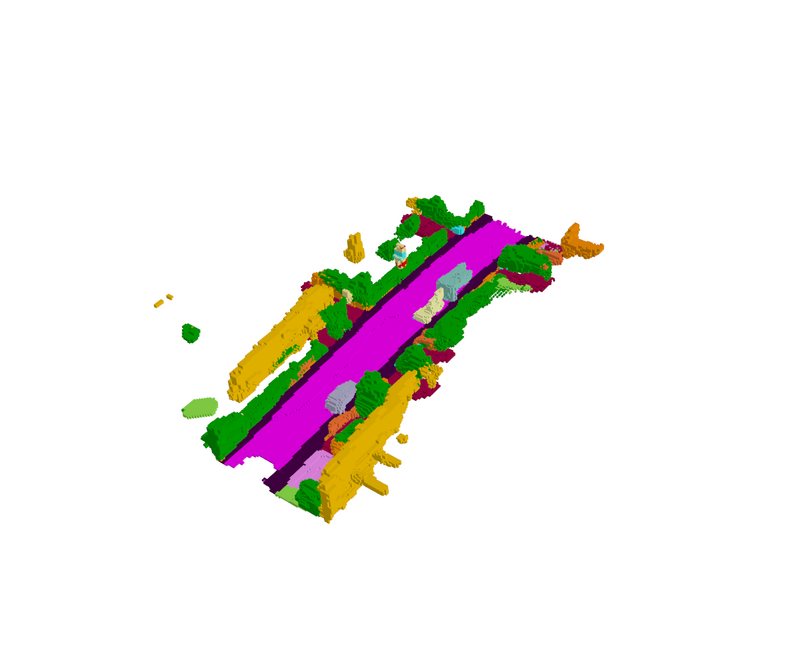} &
			\adjincludegraphics[width=\linewidth, trim={.2\width} {.2\height} {.2\width} {.2\height}, clip]{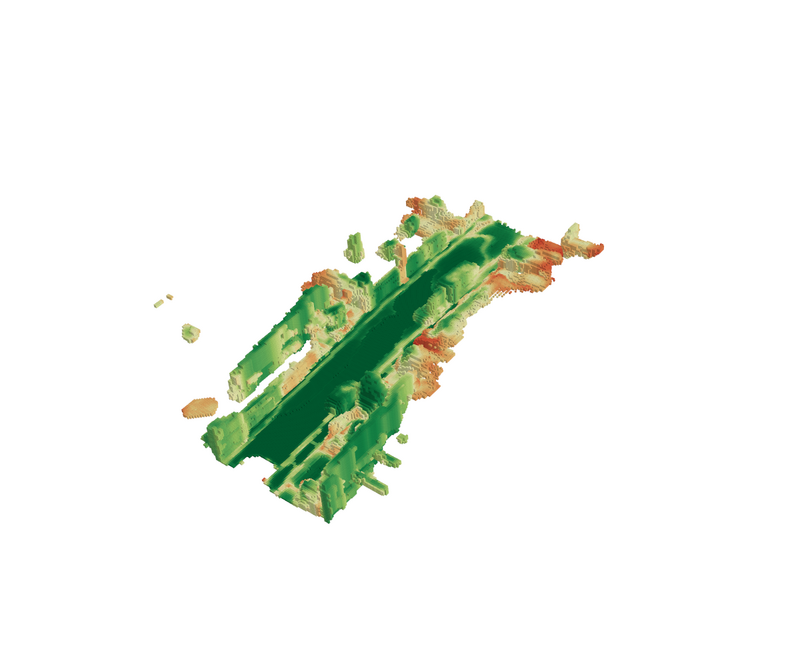} &
			\adjincludegraphics[width=\linewidth, trim={.2\width} {.2\height} {.2\width} {.2\height}, clip]{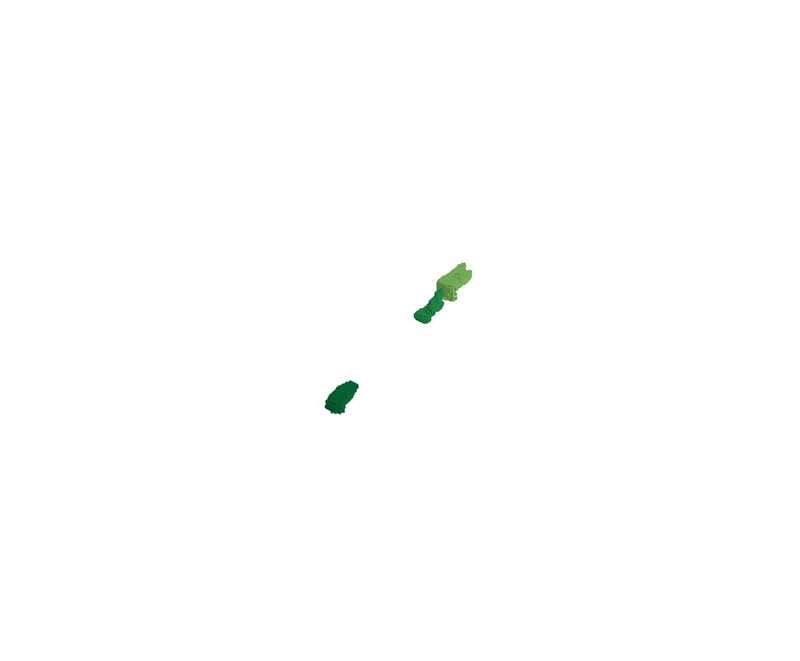} \\
			& & 	
			\rotatebox{90}{\ours} &
			\adjincludegraphics[width=\linewidth, trim={.2\width} {.2\height} {.2\width} {.2\height}, clip]{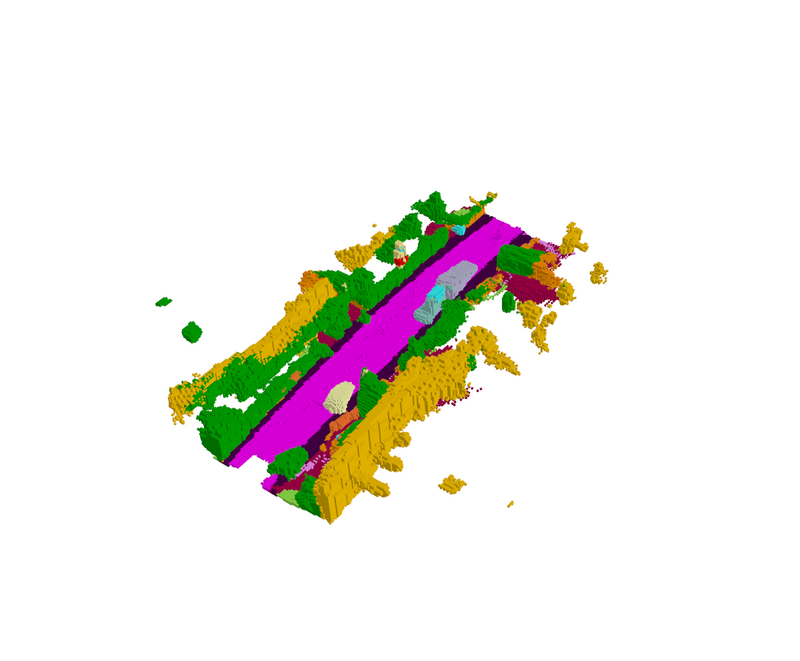} &
			\adjincludegraphics[width=\linewidth, trim={.2\width} {.2\height} {.2\width} {.2\height}, clip]{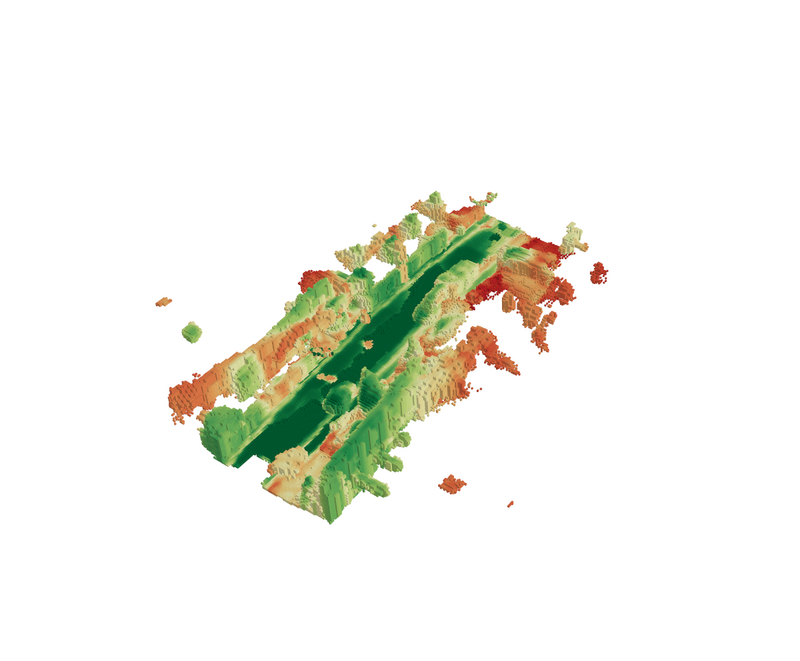} &
			\adjincludegraphics[width=\linewidth, trim={.2\width} {.2\height} {.2\width} {.2\height}, clip]{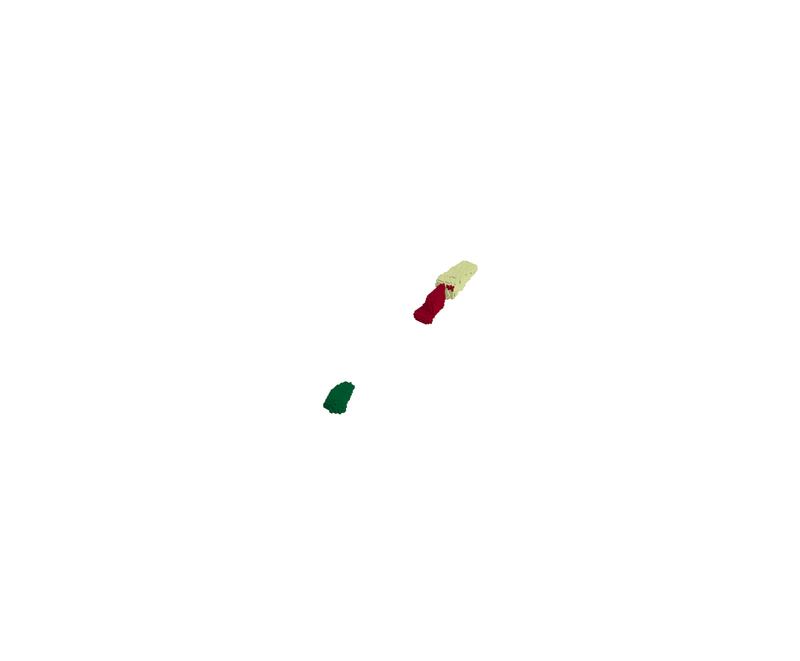} 
			\\
			& & & & \multicolumn{2}{c}{\adjincludegraphics[width=3cm]{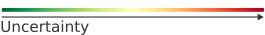}}  \\
			
		\end{tabular}			
	\end{subfigure}
	\caption{
  \textbf{Qualitative uncertainty comparison on SSCBench-KITTI360 and Semantic KITTI.} 
  Note that ``ins. unc." only shows examples from the ``thing'' class for clearer visualization. \oursMIMO{1} tends towards overconfidence in both voxel and ins. unc. In contrast, \ours~gives more intuitive uncertainty estimates, e.g., at segment boundaries, in areas with hallucinated scenery, and in regions with low input point density.}
	\label{fig:qualitative_uncertainty}
\end{figure} 

\cref{fig:qualitative_uncertainty} visualizes uncertainty estimation from~\oursMIMO{1} and~\ours on Semantic KITTI and SSCBench-KITTI360. For clarity, instance-wise uncertainty shows only ``thing" categories. \oursMIMO{1} often shows high confidence, likely due to deep networks' tendency for overconfidence~\cite{guo2017calibration}. For voxel-wise uncertainty, \ours{} exhibits increased uncertainty at segment boundaries (e.g., roads, sidewalks), low point density areas, and large missing regions. Instance-wise, \ours indicates more uncertainty in regions with ambiguous predictions, like sparse input points or close object proximity.

\begin{table}
	\centering
	\scriptsize
	\setlength{\tabcolsep}{0.005\linewidth}
	\resizebox{1.0\linewidth}{!}{
		\newcolumntype{H}{>{\setbox0=\hbox\bgroup}c<{\egroup}@{}}
		\begin{tabular}{ l |ccH|ccHHH|c || ccH|ccHHH|c }
			\toprule
			& \multicolumn{9}{c||}{\textbf{Semantic KITTI} (val set)} & \multicolumn{9}{c}{\textbf{SSCBench-KITTI360} (test set)}\\
			method  & 
			All PQ$^{\dagger}$$\uparrow$ & All PQ$\uparrow$ & mIoU$\uparrow$ & ins ece$\downarrow$ & ins nll$\downarrow$ &  voxel ece$\downarrow$ & voxel nll$\downarrow$ & Params$\downarrow$   &  Passes $\downarrow$ &
			All PQ$^{\dagger}$$\uparrow$ & All PQ$\uparrow$ & mIoU$\uparrow$ & ins ece$\downarrow$ & ins nll$\downarrow$ &  voxel ece$\downarrow$ & voxel nll$\downarrow$ & Params$\downarrow$   &  Passes $\downarrow$ \\
			\midrule
			TTA & 
			28.16 & 15.95 & 28.84 & 0.5295	& 4.3804 & 0.0456 & 0.7224 & \best{111M}  & \second{3} & 
			23.31 & 9.76 & 21.78	& 0.6953 & 4.8958 & 0.1580	& 2.1282 & \best{111M}  & \second{2}\\
			
			MC~Dropout & 
			29.62 & 16.11 & 28.82 & 0.5684 & 4.8617 & 0.0472 & 0.7437 & \best{111M} & \second{3} & 
			23.73 & \second{9.95} & 21.73 & 0.6804 & \second{4.6174} &  0.1548 & 2.0737 & \best{111M} & 2\\
			
			Deep Ensemble & 
			\second{30.71}	&  \second{16.41} & \second{30.10} & \second{0.5008} & \second{3.9181} & \second{0.0428} & \second{0.6993} & 333M  & \second{3} & 
			\second{23.85} & 9.88 & \best{22.51}  & \second{0.6673}	& 4.7809 & \second{0.1540} &	\second{2.0653} & 222M   & \second{2}\\
			
			\ours (Ours) & 
			\best{31.42} & \best{16.51} & \best{30.11} & \best{0.4922} & \best{3.9155} & \best{0.0426} & \best{0.5835} & \second{120M}  & \best{1} & 
			\best{26.29} &  \best{10.92} & \second{22.39} & \best{0.6015} & \best{4.1454} & \best{0.1348} & \best{1.6112} & \second{115M} & \best{1}\\
			\bottomrule
		\end{tabular}
	}
	\caption{\textbf{Effect of our ensembling.} We apply our permutation-invariant ensembling strategy (\cref{sec:met_uncertainty_ensemble}) to all baselines to enable PSC uncertainty estimation. Even when using our technique, we note \ours remains the best performing.}
	\label{tab:masks_ensemble}
\end{table}

\condenseparagraph{Mask ensembling.} 
As the uncertainty-aware baselines do not estimate instance uncertainties, we apply our permutation-invariant ensembling (\cref{sec:met_uncertainty_ensemble}) to all baselines, in order to enable instance-wise uncertainty estimation for all. \cref{tab:masks_ensemble} shows that our MIMO-strategy performs better than the baselines on all metrics, using a single pass.

\begin{figure}
	\centering
	\begin{minipage}[b]{0.50\linewidth}
		\includegraphics[width=\textwidth]{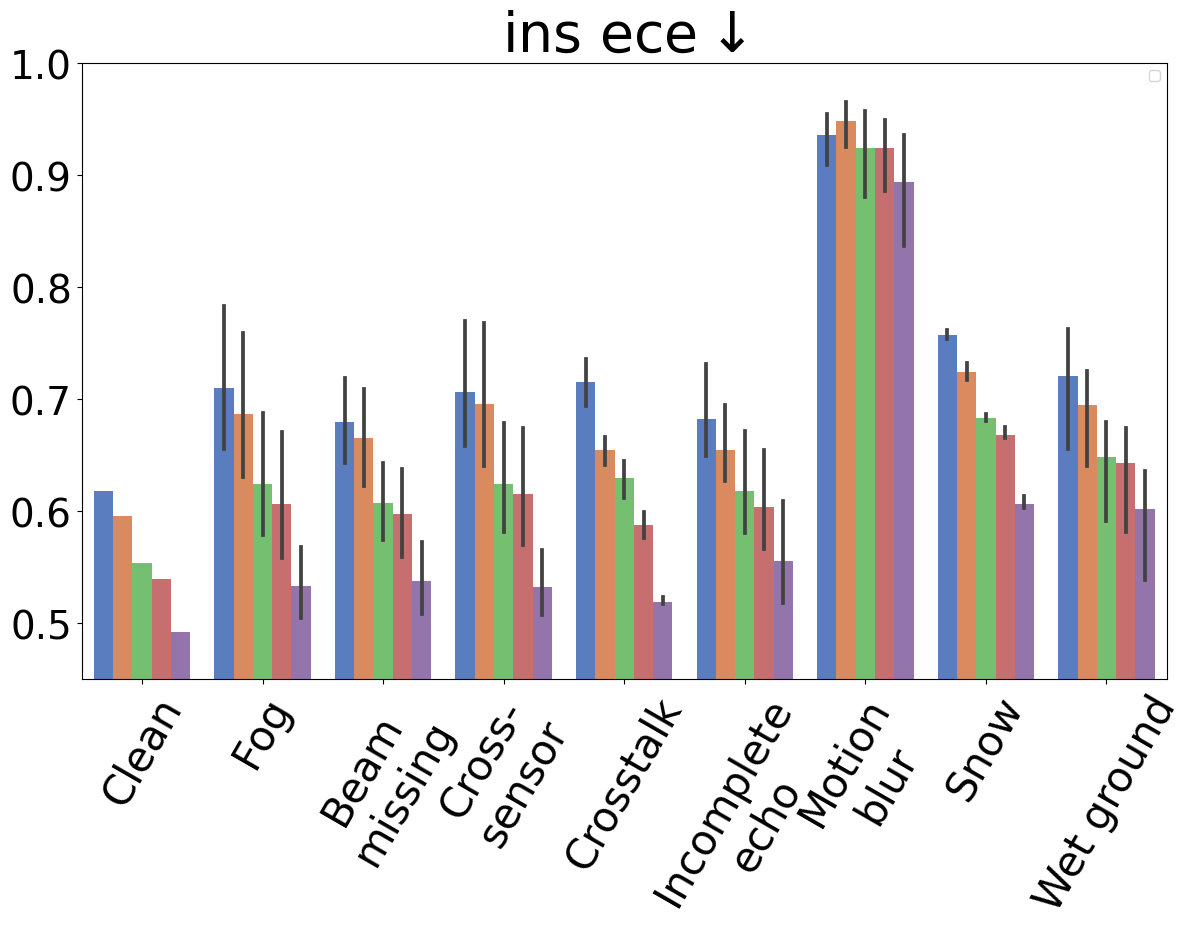}
	\end{minipage}\hfill%
	\begin{minipage}[b]{0.50\linewidth}
		\includegraphics[width=\textwidth]{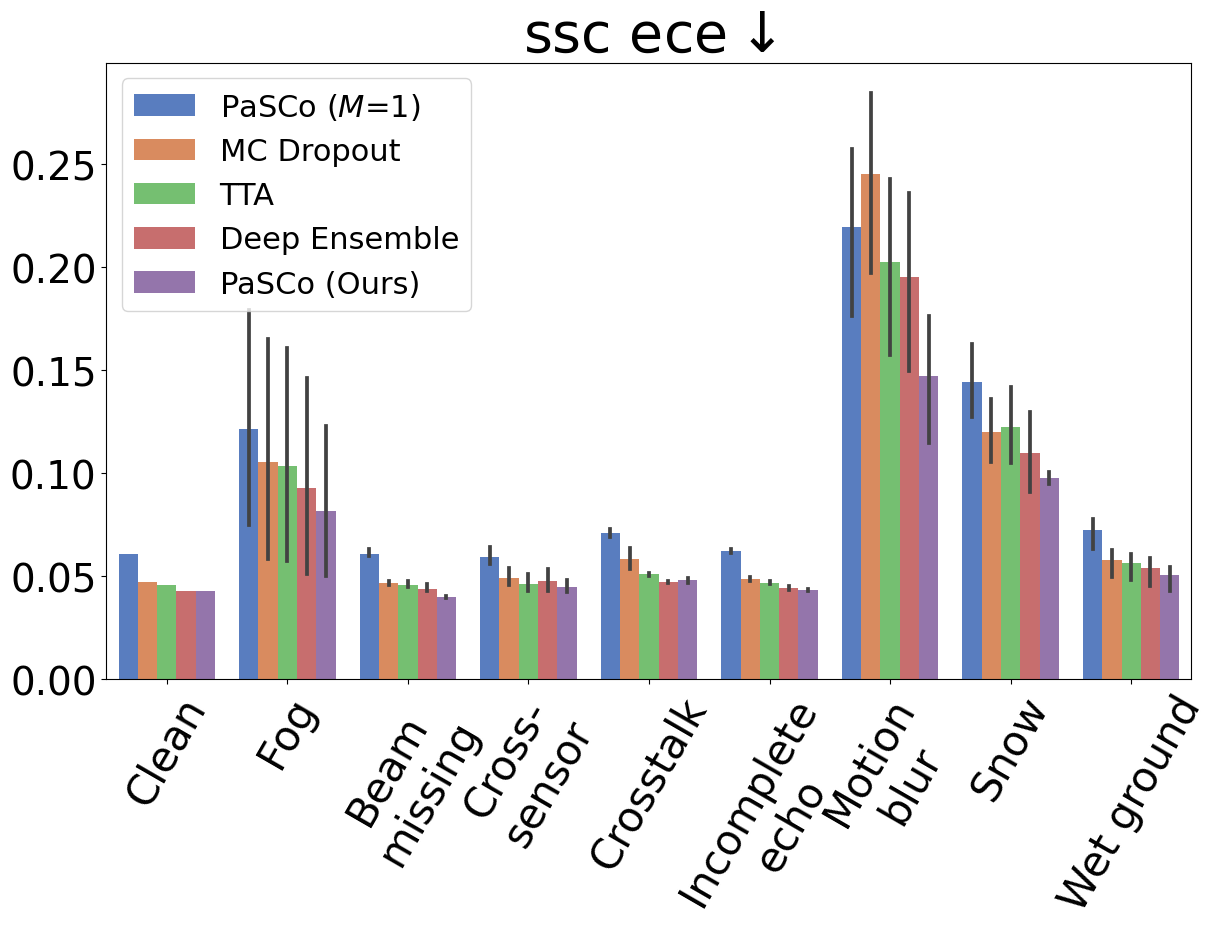}
	\end{minipage}%
	\vspace{-1em}
	\caption{\textbf{Effects of Out Of Distribution.} We evaluate uncertainties on corruptions of the Robo3D~\cite{robo3d}, shown in the x axis. Each bar reports the metric average per corruption while its error bar indicates the per-intensity minimum and maximum metric. \ours outperforms all methods by a large margin on all corruptions for instance-wise uncertainty (left) and better on 7 of 8 conditions (xcept `cross-talk')  on voxel-wise uncertainty (right).}
	\label{fig:robo3d}
\end{figure}

\begin{table*}
	\centering
	\scriptsize
	\setlength{\tabcolsep}{0.005\linewidth}
	\newcommand{\oursetting}{\cellcolor{gray!10}}
	\resizebox{1.0\linewidth}{!}{%
		\begin{tabular}{l|cc|c|cccc||cc|c|cccc|c}
			\toprule
			& \multicolumn{7}{c|}{\textbf{Semantic KITTI} (val. set)} & \multicolumn{7}{c||}{\textbf{SSCBench-KITTI360} (val. set)} & \multirow{2}{*}{\# Params$\downarrow$} \\
			\# subnets & 
			All PQ$^{\dagger}$$\uparrow$ & All PQ$\uparrow$ & mIoU$\uparrow$ & ins ece$\downarrow$ & ins nll$\downarrow$ & voxel ece$\downarrow$ & voxel nll$\downarrow$ & 
			All PQ$^{\dagger}$$\uparrow$ & All PQ$\uparrow$ & mIoU$\uparrow$ & ins ece$\downarrow$ & ins nll$\downarrow$ & voxel ece$\downarrow$ & voxel nll$\downarrow$ &  \\
			\midrule
			1 & 
			26.49 & 15.36 & 28.22 & 0.6181	& 4.6559	& 0.0610	& 0.8250 & 
			18.87 & \second{7.77} & 20.59 & 0.8355	& 6.2581 & 0.1744 &	2.5785 & \best{111M} \\
			2 & 
			30.34 & \best{17.23} & \second{30.04} & 0.5535 &	\second{4.1474} &	0.0530 &	0.6449  & 
			\oursetting\best{27.20} & \oursetting\best{8.36} & \oursetting\best{21.63} & \oursetting0.6022 & \oursetting4.4120 & \oursetting0.1285 & \oursetting1.8063 & \oursetting\second{115M} \\
			3 & 
			\oursetting\best{31.42} & \oursetting\second{16.51} & \oursetting\best{30.11} &  \oursetting\best{0.4922}	& \oursetting\best{3.9155}	& \oursetting\second{0.0426}	& \oursetting\second{0.5835} & 
			22.31 & 6.88 & \second{20.60} & \second{0.5293} & \second{3.4189} & \second{0.1233} & \second{1.6188} & 120M \\
			4 & 
			\second{31.20} & 16.33 & 29.41 &  \second{0.5304}	& 4.2681 & \best{0.0349} & \best{0.5572} & 
			\second{23.23} & 6.49 & 20.34 & \best{0.4098} & \best{2.4370} & \best{0.1011} & \best{1.4814} & 125M \\
			\bottomrule
		\end{tabular}%
	}
	\caption{\textbf{Performance when varying number of subnets on Semantic KITTI~\cite{semkitti} and SSCBench-KITTI360~\cite{sscbench}} validation sets. 
    PSC performance improves as the number of $\nsubnets$ increases, peaking at $\nsubnets{=}3$ for Semantic KITTI and at $\nsubnets{=}2$ for SSCBench-KITTI360. Further increasing the subnets can also help with uncertainty estimates.
    We choose $\nsubnets{=}3$ for Semantic KITTI and $\nsubnets=2$ for SSCBench-KITTI360 to balance high PSC performance and uncertainty estimation.
    }
	\label{tab:varied_n_subnets}
\end{table*}

\condenseparagraph{Effects of Out Of Distribution.} In the literature, uncertainty is classically used as a proxy of robustness to Out Of Distribution (OOD). To complement our study, we evaluate on the Robo3D~\cite{robo3d}, which provides point cloud under eight types of corruptions (\eg, fog, beam missing, cross-sensor, wet ground, \etc), each with three level of intensities (light, moderate, heavy). 
We evaluate on the complete set of 24 corruptions and plot instance and voxel uncertainties in~\cref{fig:robo3d}, showing that \ours demonstrates consistent improvement over baselines. 
Each bar shows the mean uncertainty of a method on a given corruption, while the error bar shows the per-level minimum and maximum uncertainties. 
Interestingly, we note that instance (\cref{fig:robo3d}, left) and voxel (\cref{fig:robo3d}, right) uncertainties are not strongly correlated, although methods' rankings remain rather stable across conditions.
For instance-wise uncertainty (`ins ece'), \ours is significantly better than \textit{all} baselines on \textit{all} corruptions, improving in 7 out of 8 on voxel-wise uncertainty (`ssc ece').

\subsection{Ablation Studies} 
\label{sec:exp_ablation}
\begin{table}
	\centering
	\scriptsize
	\setlength{\tabcolsep}{0.005\linewidth}
	\resizebox{1.0\linewidth}{!}{%
		\begin{tabular}{l |cc|c|ccccc}
			\toprule
			&  All PQ$^{\dagger}$$\uparrow$ & All PQ$\uparrow$ &  mIoU$\uparrow$ & ins ece$\downarrow$ & ins nll$\downarrow$ & voxel ece$\downarrow$ & voxel nll$\downarrow$  \\
			\midrule
			w/o augmentation & 27.89	& 14.07 & 28.30	& \second{0.5031} &	4.4245 &	0.0442 &	0.6713 \\
			w/o rotation augmentation & 28.84 & 14.95 & 28.95 & 0.5074	& 4.2987 & 0.0432 &	0.6309\\
			\midrule
			w/o voxel-query sem. loss & 28.82 & \second{15.55} & \second{29.87} & 0.5205 & 4.2909	& 0.0437 & \second{0.5878} \\
			w/o sem. pruning &  \second{30.12} & 15.04 & 29.04  & 0.5380 & \second{4.1814} & \second{0.0440} & 0.5980 \\
			\ours (Ours) & \textbf{31.42} & \textbf{16.51}  & \textbf{30.11} & \textbf{0.4922} & \textbf{3.9155} & \best{0.0426} & \textbf{0.5835} \\
			\bottomrule
		\end{tabular}%
	}\vspace{-1em}
	\caption{\textbf{Method ablation.} We ablate inference (top) and training (bottom) components of our method, showing that each contributes to the best performance.}
	\label{tab:ablation}
\end{table}
\condenseparagraph{Method ablation.} 
We ablate our method on SemanticKITTI~\cite{semkitti} in~\cref{tab:ablation} and report SSCBench-KITTI360 in~\cref{tab:supp:ablation_kitti360}. The upper table ablates our inference augmentations (\ie, rotation+translation), which  benefit overall performance, especially All-PQ$^\dagger$/PQ.  We attribute this to the increased variance profitable to the subnetworks as in MIMO~\cite{mimo}.
In the lower table, we retrain \ours while removing some components. We show that removing our voxel-query semantic loss $\SL'_{{\rm sem}}$ (\cref{eq:voxel-query}) harms training; such proxy supervision boosts performance at no additional cost. 
Finally, `w/o sem. pruning' replaces our semantic pruning with binary occupancy pruning~\cite{s3cnet, sgnn}, resulting in degraded performance due to loss of smaller classes.

\noindent\textbf{Subnets ablation.}~\cref{tab:varied_n_subnets} ablates different numbers of subnets $M \in \{1,2,3,4\}$ on the validation sets of our main datasets. Our main PQ metrics increase significantly with more subnets, though plateauing at $M=3$ for SemanticKITTI and $M=2$ for SSCBench-KITTI360.
This is due to the preserved constant computational cost implying that more subnets mean less per-subnet capacity, leading to more noise in the ensembling. Our finding confirms that of MIMO~\cite{mimo} in the classification setting, though we argue our plateau is reached before since PSC being is a much more complex task than classification.\\
		
\noindent\textbf{Mask matching.}
We ablate our mask matching, substituting our `soft matching' (sigmoid probabilities) with `hard matching' with binary mask IoU for assignment cost matrix calculation (\cref{sec:met_uncertainty_ensemble}). This results in a large drop in All-PQ$^\dagger$/PQ of -3.75/-1.12. Entirely removing mask matching severely impacts mask quality, dropping to 0.02/0.02.

\begin{figure}
  \centering
	\begin{minipage}[b]{0.25\linewidth}
		\includegraphics[width=\textwidth]{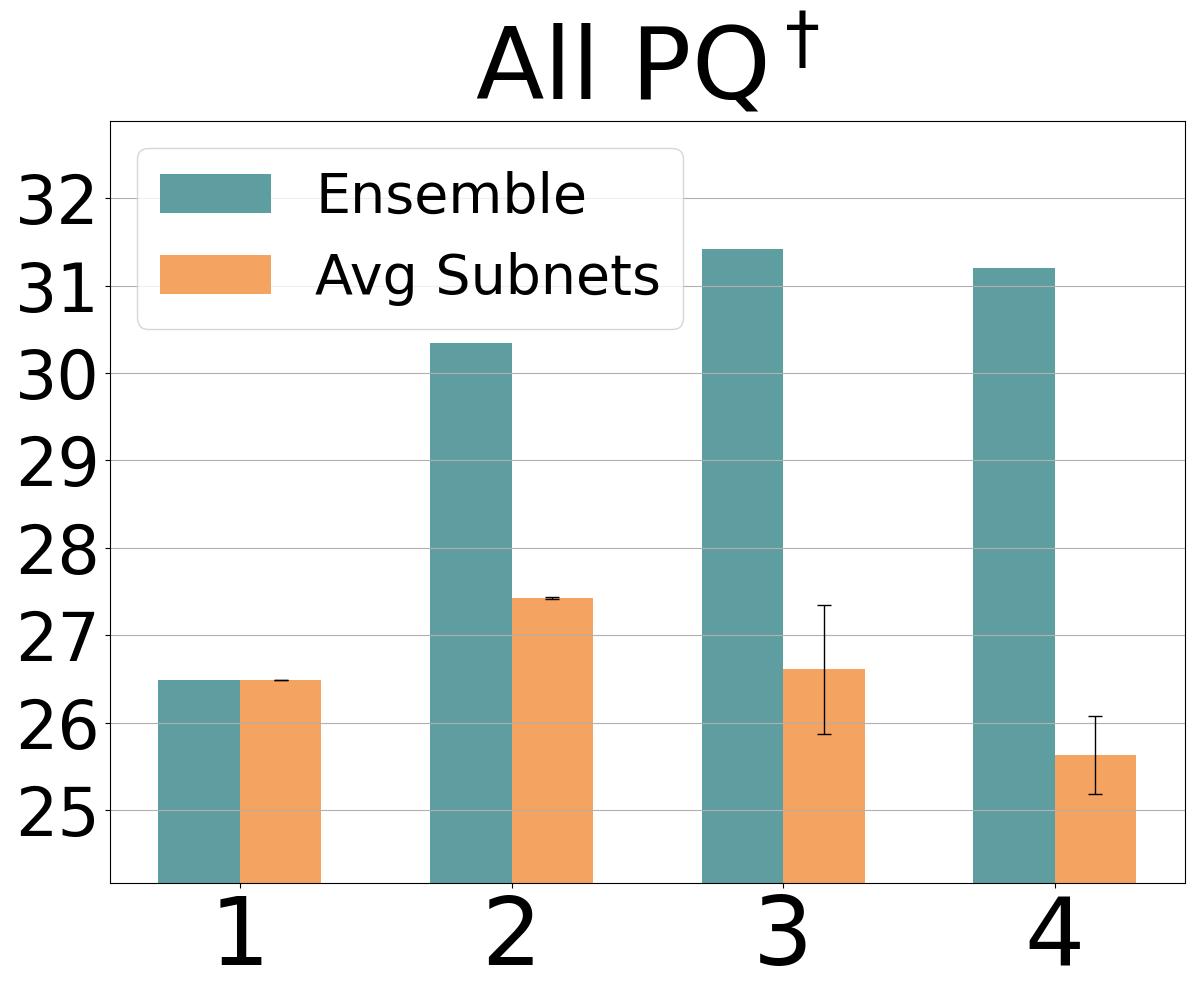}
		\label{fig:pq_dagger}
	\end{minipage}%
	\hfill%
	\begin{minipage}[b]{0.25\linewidth}
		\includegraphics[width=\textwidth]{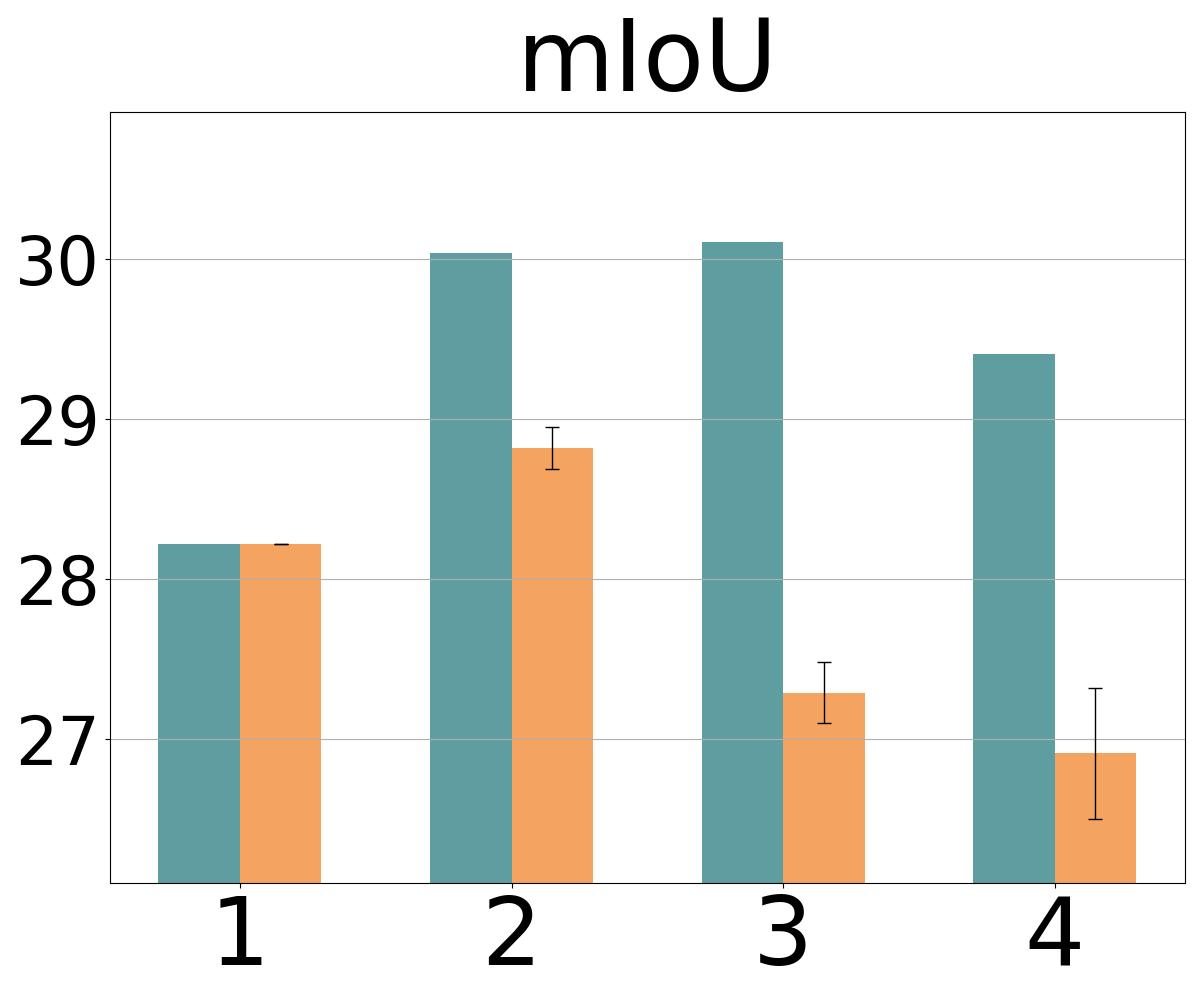}
		\label{fig:ssc_ece}
	\end{minipage}%
	\hfill%
	\begin{minipage}[b]{0.25\linewidth}
		\includegraphics[width=\textwidth]{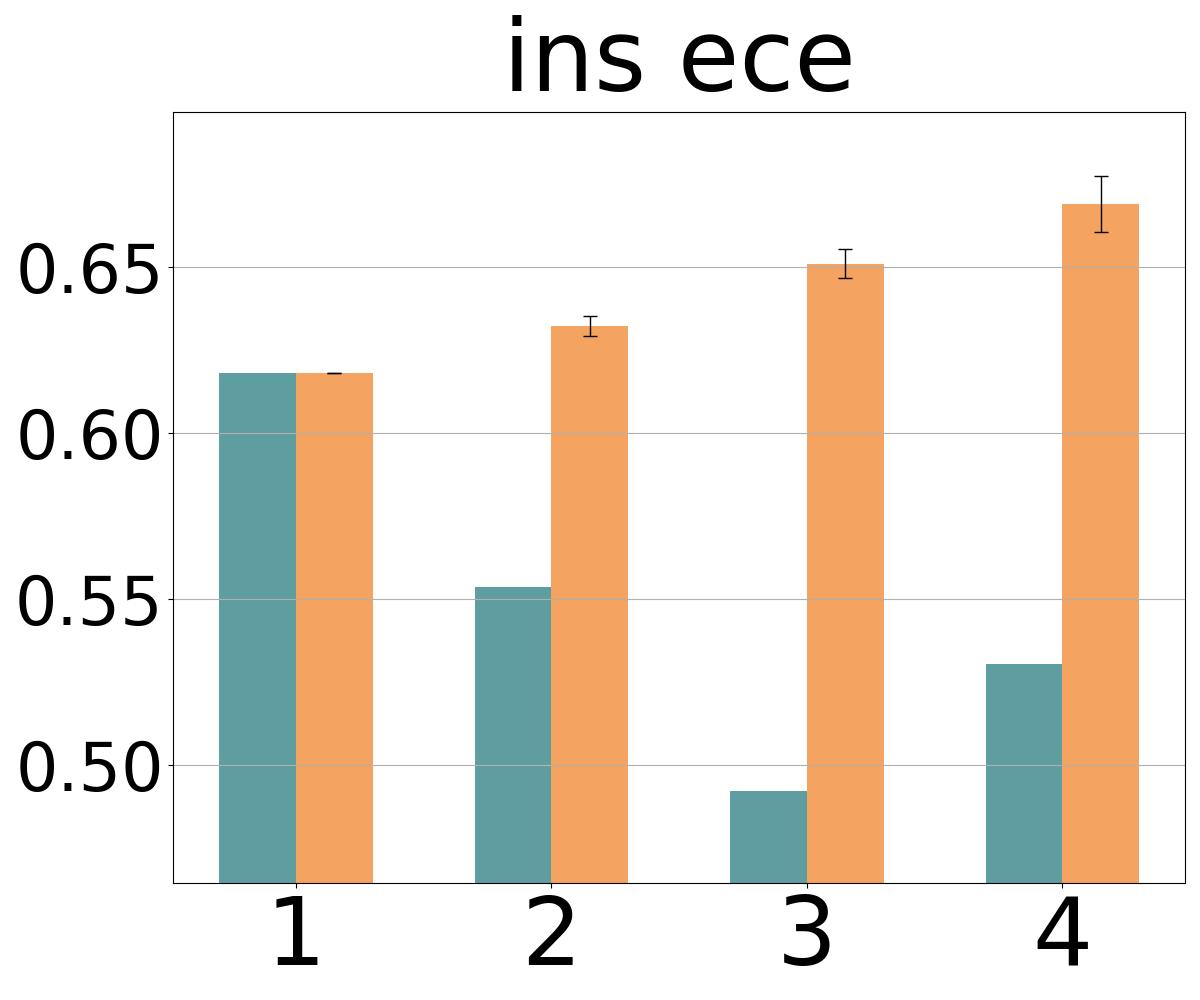}
		\label{fig:ins_ece}
	\end{minipage}%
	\hfill%
	\begin{minipage}[b]{0.25\linewidth}
		\includegraphics[width=\textwidth]{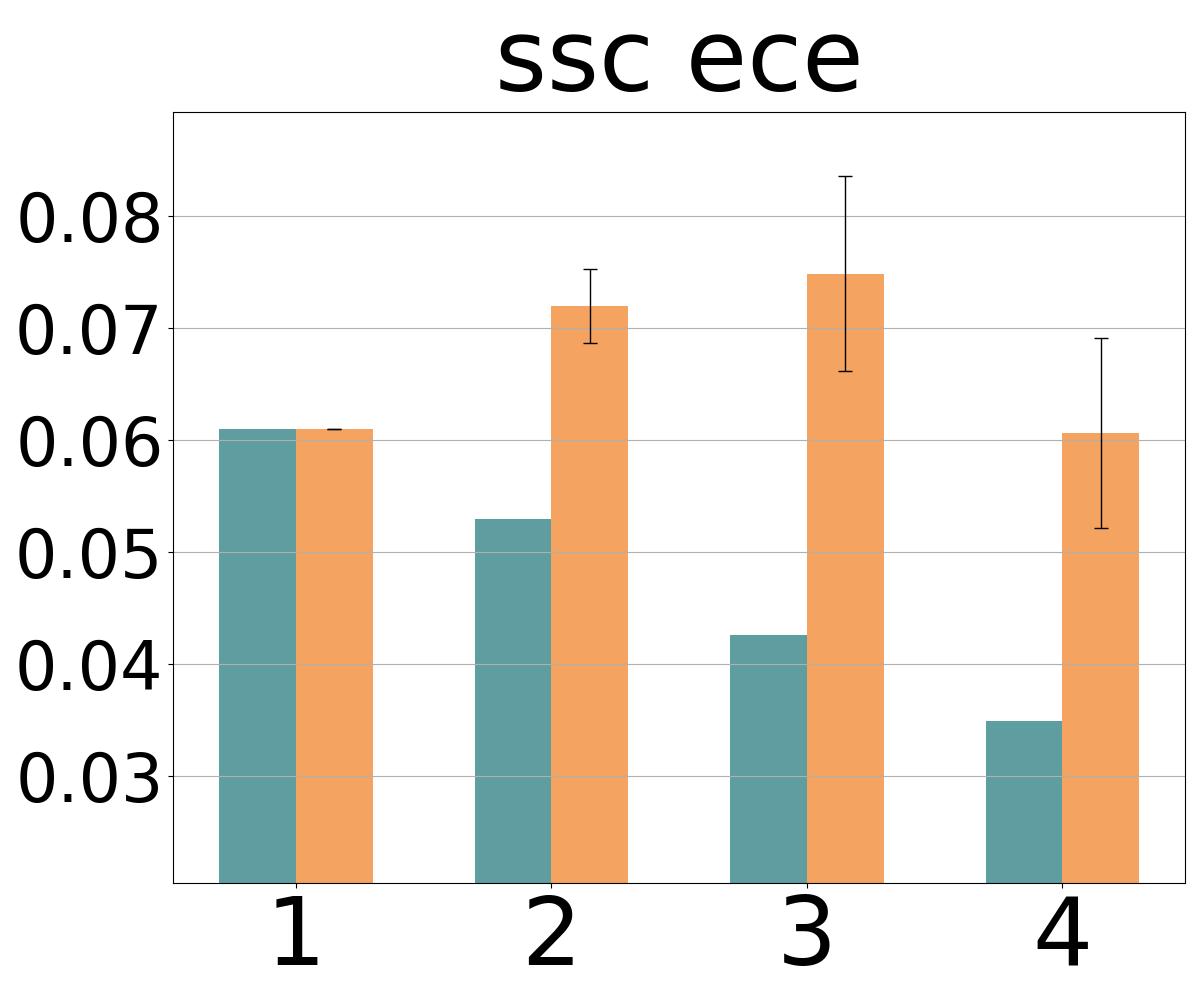}
		\label{fig:ins_ece}
	\end{minipage}%
\vspace{-0.7cm}
  \caption{\textbf{Ensemble vs subnets averaging.} 
  We compare our ensemble method with averaging individual subnets, across varying \# subnets (x-axis). Error bars show standard deviation across subnets. Peak performance is at $\nsubnets{=}3$, where our ensembling compensates for reduced per-subnet capacity with more subnets.
  }
  \label{fig:ensemble_vs_subnets}
\end{figure}

\noindent\textbf{Ensemble vs. Subnets averaging}. To further shed light on subnets performance, \cref{fig:ensemble_vs_subnets} displays metrics of SemanticKITTI as a function of number of subnets $M$ for our ensembling (\cref{sec:met_uncertainty_ensemble}) or the averaging of the individual subnet performance. 
When averaging subnets, optimal performance is reached at $M=2$, with larger $M$ unable to solve PSC efficiently. However, ensembling reaches improved performance at $M=3$, showing that our ensembling effectively leverages weaker subnets. 

\noindent\textbf{Limitations.} Like MIMO~\cite{mimo}, our method can accommodate a limited number of subnets, depending on the task nature and the network's capacity. Our approach may overlook objects or mix up nearby objects, particularly when they are small and exhibit semantic resemblance. Our method does not distinguish between types of uncertainty, such as epistemic or aleatoric. Exploring this aspect could be a valuable direction for future research.

\section{Conclusion}
We first address Panoptic Scene Completion (PSC) which aims to complete scene geometry, semantics, and instances from a sparse observation. We introduce an efficient ensembling method complemented by a novel technique that combines predictions of unordered sets, enhancing the overall prediction accuracy and reliability in terms of uncertainty.

{\footnotesize
\condenseparagraph{Acknowledgment.} 
{The research was funded by the ANR project SIGHT (ANR-20-CE23-0016), the ERC Starting Grant SpatialSem (101076253), and SAMBA collaborative project co-funded by BpiFrance in the Investissement d’Avenir Program. Computation was performed using HPC resources from GENCI–IDRIS (2023-AD011014102, AD011012808R2). We thank all \href{https://astra-vision.github.io/}{Astra-Vision} members for their valuable feedbacks, including \mbox{Andrei~Bursuc} and \mbox{Gilles Puy} for excellent suggestions and \mbox{Tetiana~Martyniuk} for her kind proofreading.}\par}

\clearpage
\setcounter{page}{1}
\maketitlesupplementary

We begin by reporting implementation details of \ours and baselines in~\cref{sec:supp:implementation}, and present additional ablations in~\cref{sec:supp:abl}.
Finally, we provide additional experiments on uncertainty estimation and panoptic scene completion in~\cref{sec:supp:exp}, showcasing \ours performance.

We refer to the \textbf{supplementary video} for better qualitative judgment of \ours performance.

\section{Implementation details}
\label{sec:supp:implementation}

\paragraph{\ours.}
\label{sec:supp:our_details}
Our network employs sparse convolution from the MinkowskiEngine library~\cite{mink}.
The architecture of our Dense 3D CNN is similar to the 3D Completion Sub-network of SCPNet~\cite{scpnet}. Additionally, the implementation of the MLP and voxelization is based on Cylinder3D~\cite{zhu2020cylindrical}. 

Training \ours~required three days for the Semantic KITTI dataset and five days for the SSCBench-KITTI360 dataset using 2 A100 GPUs (1 item per GPU).

\paragraph{Baselines.}
\label{sec:supp:baselines_detail}

We employed the official implementations of \textbf{LMSCNet}\footnote{\url{https://github.com/cv-rits/LMSCNet}}, \textbf{JS3CNet}\footnote{\url{https://github.com/yanx27/JS3C-Net}}, \textbf{SCPNet}\footnote{\url{https://github.com/SCPNet/Codes-for-SCPNet}}, and \textbf{MaskPLS}\footnote{\url{https://github.com/PRBonn/MaskPLS}} with their provided parameters. 

For SCPNet, despite many email exchanges with the authors we were unable to reproduce their reported performance using their official code as also mentioned by other users~\footnote{\url{https://github.com/SCPNet/Codes-for-SCPNet/issues/8}}.
Hence, we put extra effort to reimplement their method following authors' recommendation, which resulted in SCPNet*. Note that the latter is several points better than the official implementation.

\section{Additional ablations}
\label{sec:supp:abl}
\paragraph{Method ablation on SSCBench-KITTI360.}
We provide additional ablations of our method on SSCBench-KITTI360 validation set in~\cref{tab:supp:ablation_kitti360}, which align with the results reported in~\cref{tab:ablation}. 
The first two rows of \cref{tab:supp:ablation_kitti360} ablate the augmentations used during inference (\ie rotation + translation). The subsequent rows present the performance of \ours, retrained w/o our proposed components. 

Notably, substituting semantic pruning with binary occupancy pruning~\cite{s3cnet, sgnn} leads to a substantial decline in performance, particularly in PSC metrics with -4.82/-1.86 in All PQ$^\dagger$/All PQ. 
This result is expected as semantic pruning not only balances supervision across classes, particularly smaller ones, but also provides additional information to the network. 
Removing our voxel-query semantic loss (\cref{eq:voxel-query}) also leads to a remarkable decrease in performance, with a -2.24/-0.77 drop in All PQ$^\dagger$/All PQ, demonstrating its effectiveness without incurring additional computational costs. 
Lastly, the augmentations applied during inference (top rows) contribute to increased variation among subnetworks, thereby enhancing overall performance.

\begin{table}
	\centering
	\scriptsize
	\setlength{\tabcolsep}{0.005\linewidth}
	\resizebox{1.0\linewidth}{!}{%
		\begin{tabular}{l |cc|c|ccccc}
			\toprule
			&  All PQ$^{\dagger}$$\uparrow$ & All PQ$\uparrow$ &  mIoU$\uparrow$ & ins ece$\downarrow$ & ins nll$\downarrow$ & voxel ece$\downarrow$ & voxel nll$\downarrow$  \\
			\midrule
			w/o augmentation & 26.42 & \second{8.09} & 20.40 & 0.6114	& 4.7203 & 0.1296 & 2.0639\\
			w/o rotation augmentation & \second{26.48} & 7.90 & \second{21.20} & \second{0.6028} & 4.6321	& \second{0.1293} & 1.9261\\
			\midrule
			w/o voxel-query sem. loss & 24.96 & 7.59 & 20.90 & 0.6381 & \second{4.4526} & 0.1307 & \second{1.8124} \\
			w/o sem. pruning & 22.38 & 6.50 & 19.59 & 0.6244 & 4.5468 & 0.1364 & 2.0997 \\
			\ours (Ours) & \best{27.20} &  \best{8.36} & \best{21.63} & \best{0.6022} & \best{4.4119} & \best{0.1285} & \best{1.8063} \\
			\bottomrule
		\end{tabular}%
	}\vspace{-1em}
	\caption{\textbf{Method ablation on SSCBench-KITTI360 (val. set.)} We ablate different components of our method during inference~(top) and training (bottom), demonstrating that each plays a significant role in achieving the best performance.}
	\label{tab:supp:ablation_kitti360}
\end{table}

\begin{table*}
	\begin{subtable}{1.0\textwidth}
		\scriptsize
		\setlength{\tabcolsep}{0.004\linewidth}
		\newcommand{\classfreq}[1]{{~\tiny(\semkitfreq{#1}\%)}}  %
		\centering
		\begin{tabular}{l|l|c c c c c c c c c c c c c c c c c c c c}
			\toprule
			& Method
			& \rotatebox{90}{\textcolor{car}{$\blacksquare$} car\classfreq{car}} 
			& \rotatebox{90}{\textcolor{bicycle}{$\blacksquare$} bicycle\classfreq{bicycle}} 
			& \rotatebox{90}{\textcolor{motorcycle}{$\blacksquare$} motorcycle\classfreq{motorcycle}} 
			& \rotatebox{90}{\textcolor{truck}{$\blacksquare$} truck\classfreq{truck}} 
			& \rotatebox{90}{\textcolor{other-vehicle}{$\blacksquare$} other-veh.\classfreq{othervehicle}} 
			& \rotatebox{90}{\textcolor{person}{$\blacksquare$} person\classfreq{person}} 
			& \rotatebox{90}{\textcolor{bicyclist}{$\blacksquare$} bicyclist\classfreq{bicyclist}} 
			& \rotatebox{90}{\textcolor{motorcyclist}{$\blacksquare$} motorcyclist.\classfreq{motorcyclist}} 
			& \rotatebox{90}{\textcolor{road}{$\blacksquare$} road\classfreq{road}} 
			& \rotatebox{90}{\textcolor{parking}{$\blacksquare$} parking\classfreq{parking}} 
			& \rotatebox{90}{\textcolor{sidewalk}{$\blacksquare$} sidewalk\classfreq{sidewalk}}
			& \rotatebox{90}{\textcolor{other-ground}{$\blacksquare$} other-grnd\classfreq{otherground}} 
			& \rotatebox{90}{\textcolor{building}{$\blacksquare$} building\classfreq{building}} 
			& \rotatebox{90}{\textcolor{fence}{$\blacksquare$} fence\classfreq{fence}} 
			& \rotatebox{90}{\textcolor{vegetation}{$\blacksquare$} vegetation\classfreq{vegetation}} 
			& \rotatebox{90}{\textcolor{trunk}{$\blacksquare$} trunk\classfreq{trunk}} 
			& \rotatebox{90}{\textcolor{terrain}{$\blacksquare$} terrain\classfreq{terrain}} 
			& \rotatebox{90}{\textcolor{pole}{$\blacksquare$} pole\classfreq{pole}} 
			& \rotatebox{90}{\textcolor{traffic-sign}{$\blacksquare$} traf.-sign\classfreq{trafficsign}} 
			& \rotatebox{90}{mean}
			\\
			\midrule
			\multirow{2}{*}{\rotatebox{90}{\textbf{PQ}}} & w/o sem. pruning & 22.13 & 9.82 & 17.97 & 11.17 & 7.11 & 2.91 & 0.00 & 0.00 & 75.00 & 11.38 & 24.20 & 0.00 & 3.53 & \best{0.61} & 8.21 & \best{4.55} & 31.05 & 8.21 & 2.61 & 15.04 \\

			& \ours (Ours) & \best{24.55} & \best{7.82} & \best{18.09} & \best{44.89} & \best{11.32} & \best{3.00} & 0.00 & 0.00 & \best{76.22} & \best{28.12} & \best{30.42} & \best{1.33} & \best{4.85} & 0.27 & \best{12.97} & 4.22 & \best{32.61} & \best{9.69} & \best{3.26} & \best{16.51} \\
			
			\bottomrule
		\end{tabular}
		\caption{Semantic KITTI (val. set)}
		\label{tab:supp:perclass_semvsoc_semkitti}
	\end{subtable}
	\begin{subtable}{1.0\textwidth}
		\scriptsize
		\setlength{\tabcolsep}{0.004\linewidth}
		\newcommand{\classfreq}[1]{{~\tiny(\kittithreesixtyfreq{#1}\%)}}  %
		\centering
		\begin{tabular}{l|l|c c c c c c c c c c c c c c c c c c c c}
			\toprule
			& Method
			& \rotatebox{90}{\textcolor{car}{$\blacksquare$} car\classfreq{car}} 
			& \rotatebox{90}{\textcolor{bicycle}{$\blacksquare$} bicycle\classfreq{bicycle}} 
			& \rotatebox{90}{\textcolor{motorcycle}{$\blacksquare$} motorcycle\classfreq{motorcycle}} 
			& \rotatebox{90}{\textcolor{truck}{$\blacksquare$} truck\classfreq{truck}} 
			& \rotatebox{90}{\textcolor{other-vehicle}{$\blacksquare$} other-veh.\classfreq{othervehicle}} 
			& \rotatebox{90}{\textcolor{person}{$\blacksquare$} person\classfreq{person}}
			& \rotatebox{90}{\textcolor{road}{$\blacksquare$} road\classfreq{road}} 
			& \rotatebox{90}{\textcolor{parking}{$\blacksquare$} parking\classfreq{parking}} 
			& \rotatebox{90}{\textcolor{sidewalk}{$\blacksquare$} sidewalk\classfreq{sidewalk}}
			& \rotatebox{90}{\textcolor{other-ground}{$\blacksquare$} other-grnd\classfreq{otherground}} 
			& \rotatebox{90}{\textcolor{building}{$\blacksquare$} building\classfreq{building}} 
			& \rotatebox{90}{\textcolor{fence}{$\blacksquare$} fence\classfreq{fence}} 
			& \rotatebox{90}{\textcolor{vegetation}{$\blacksquare$} vegetation\classfreq{vegetation}} 
			& \rotatebox{90}{\textcolor{terrain}{$\blacksquare$} terrain\classfreq{terrain}} 
			& \rotatebox{90}{\textcolor{pole}{$\blacksquare$} pole\classfreq{pole}} 
			& \rotatebox{90}{\textcolor{traffic-sign}{$\blacksquare$} traf.-sign\classfreq{trafficsign}} 
			& \rotatebox{90}{\textcolor{other-struct}{$\blacksquare$} other-structure\classfreq{otherstruct}} 
			& \rotatebox{90}{\textcolor{other-object}{$\blacksquare$} other-object\classfreq{otherobject}} 
			& \rotatebox{90}{mean}
			\\
			\midrule
			\multirow{2}{*}{\rotatebox{90}{\textbf{PQ}}} & w/o sem. pruning & 12.46 & 0.00 & 2.20 & 5.00 & 1.34 & \best{1.54} & 68.78 & 1.63 & 18.73 & 0.00 & 0.71 & 0.06 & 0.17 & 0.00 & 0.37 & 3.29 & 0.00 & 0.68 & 6.50\\
			
			& \ours (Ours) & \best{16.93} & 0.00 & \best{3.00} & \best{7.67} & \best{4.56} & 0.91 & \best{69.17} & \best{2.22} & \best{22.29} & \best{0.06} & \best{7.08} & \best{0.06} & \best{3.19} & 0.00 & \best{2.59} & \best{4.85} & 0.00 & \best{5.84} &  \best{8.36} \\
			\bottomrule
		\end{tabular}\\
		\caption{SSCBench-KITTI360 (val. set)}
		\label{tab:supp:perclass_semvsocc_sscbenchkitti360}
	\end{subtable}
	\caption{\textbf{Ablation of sem. pruning on \subref{tab:supp:perclass_semkitti} Semantic KITTI (val. set) and \subref{tab:supp:perclass_sscbenchkitti360} SSCBench-KITTI360 (val. set) for Panoptic Scene Completion class-wise performance.} Semantic pruning improves the performance of the majority of classes on both datasets.}
	\label{tab:supp:perclass_semvsoc_performance}
\end{table*}

\paragraph{Semantic pruning.}
We further ablate the use of our semantic pruning in \cref{tab:supp:perclass_semvsoc_performance}, by the replacing it with binary pruning as employed in~\cite{sgnn, s3cnet}. 
From the table, semantic pruning significantly enhances performance across most classes, notably for rare classes such as truck (+33.72/+2.67 PQ on Semantic KITTI/SSCBench-KITTI360), other-vehicle (+4.21/3.22 PQ on Semantic KITTI/SSCBench-KITTI360), and pole (+1.48/+2.22 PQ on Semantic KITTI/SSCBench-KITTI360). 
These improvements can be attributed to a more balanced supervision among classes and the inclusion of additional semantic information.

\section{Additional experiments}
\label{sec:supp:exp}

We provide further experiments for uncertainty estimation in~\cref{sec:supp:exp_uncertainty} and panoptic scene completion in~\cref{sec:supp:exp_psc}.

\subsection{Uncertainty estimation}
\label{sec:supp:exp_uncertainty}

\paragraph{Robustness to Out Of Distribution (OOD).}

\begin{figure}[!h]
	\centering
	\begin{minipage}[b]{0.50\linewidth}
		\includegraphics[width=\textwidth]{plots/robo3d/all_metrics_ins_ece.png}
	\end{minipage}\hfill%
	\begin{minipage}[b]{0.50\linewidth}
		\includegraphics[width=\textwidth]{plots/robo3d/all_metrics_ssc_ece.png}
	\end{minipage}%

	\begin{minipage}[b]{0.50\linewidth}
		\includegraphics[width=\textwidth]{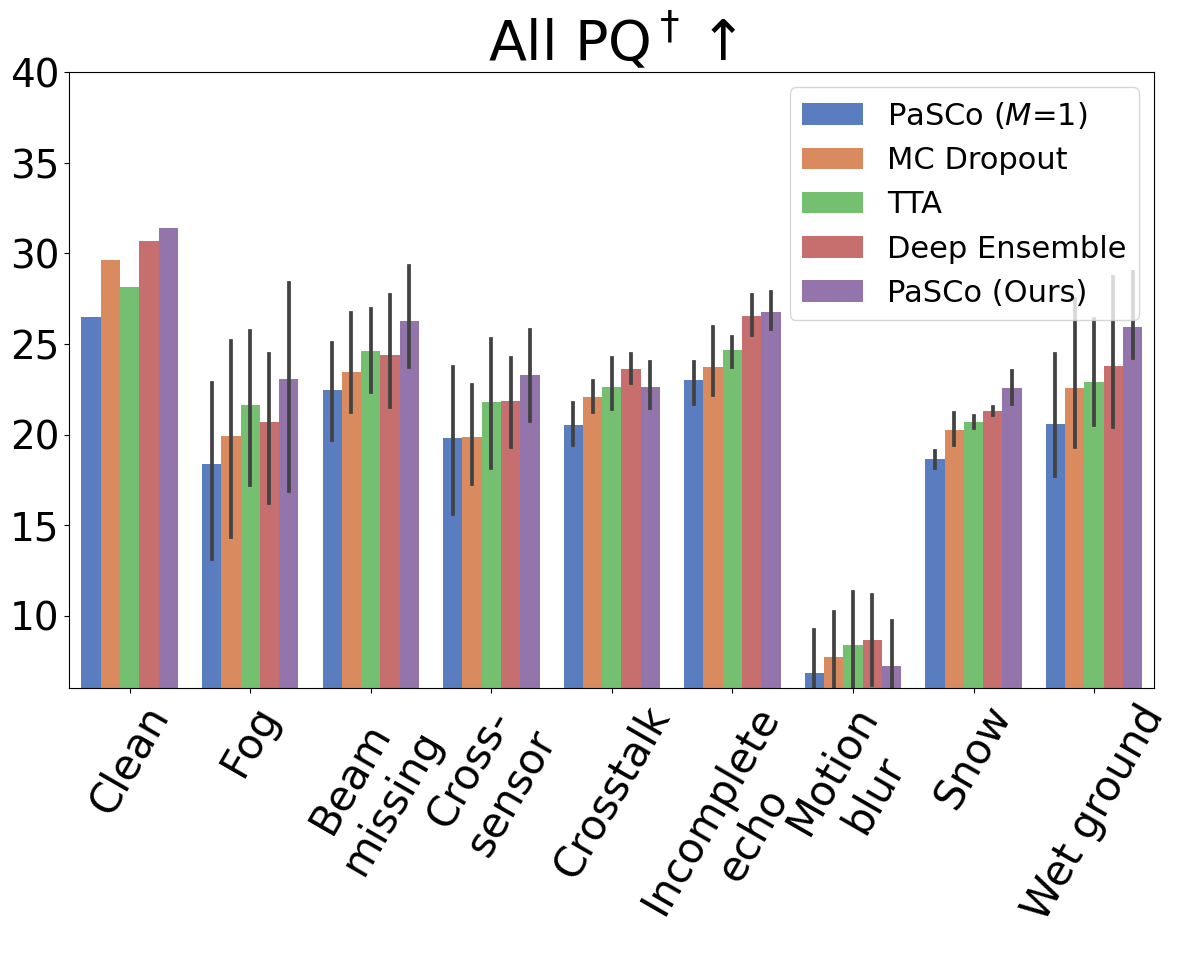}
	\end{minipage}\hfill%
	\begin{minipage}[b]{0.50\linewidth}
		\includegraphics[width=\textwidth]{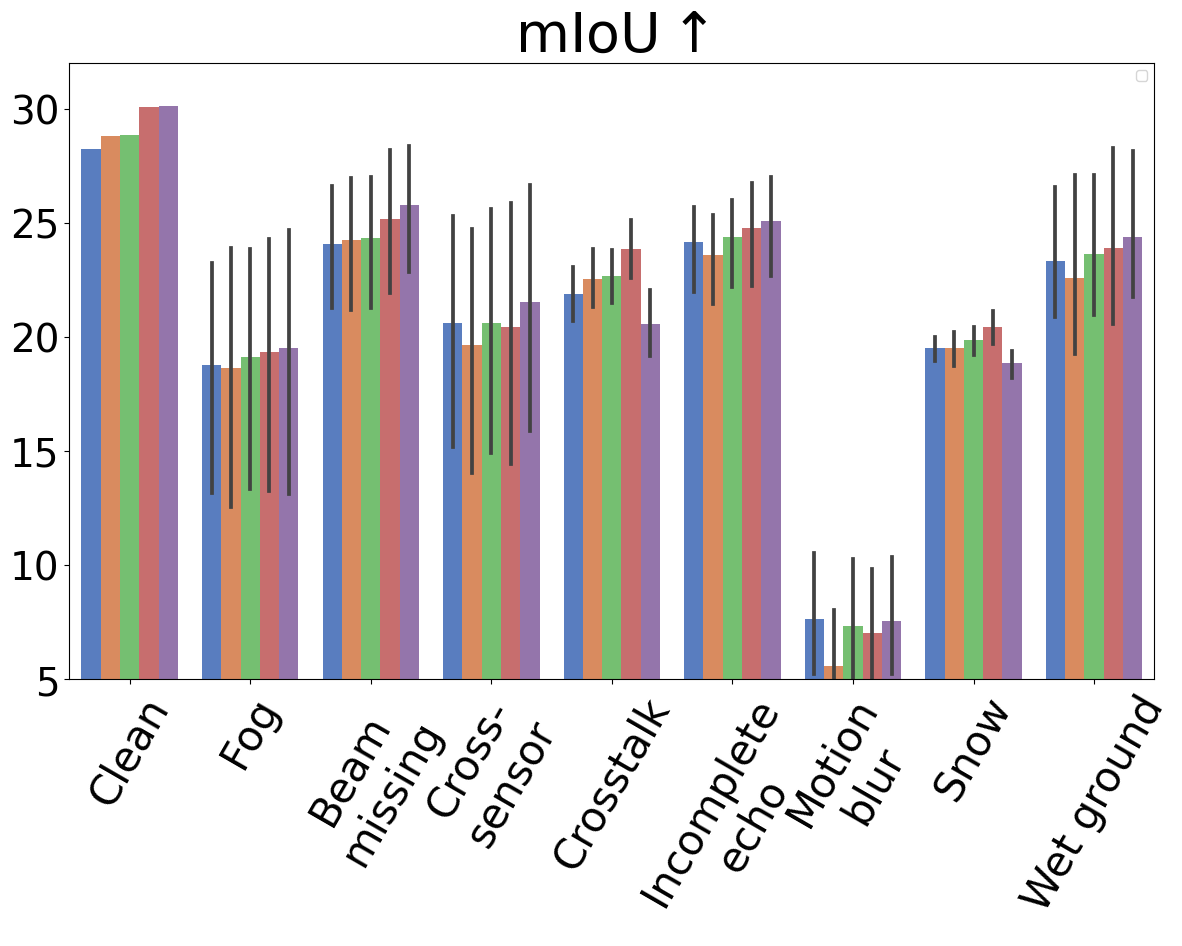}
	\end{minipage}%

	\vspace{-1em}
	\caption{\textbf{Impact of Out Of Distribution data on PSC and uncertainty performance.} We evaluate \ours and baselines, trained on `clean' Semantic KITTI on corrupted versions of the same set from the Robo3D~\cite{robo3d} dataset. Top part reports, uncertainty ece metrics (\textit{lower is better}), while bottom part reports performance metrics (\textit{higher is better}).
		The types of corruptions are shown on the x-axis. Each bar represents the average metric for each type of corruption, with the error bars showing the minimum and maximum metric across intensities. 
		\ours surpasses all baselines in terms of uncertainty measurement (\ie, {lower} ece) and demonstrates comparable or better performance metrics (\textit{higher} All PQ$^\dagger$ and mIoU). 
	}
	\label{fig:supp:robo3d_uncertainty}
\end{figure}

We extend our evaluation for Out Of Distribution, initially reported in~\cref{sec:exp_uncertainty} and~\cref{fig:robo3d}.
In \cref{fig:supp:robo3d_uncertainty}, we again evaluate baselines MC Dropout~\cite{mcdropout}, TTA~\cite{tta}, Deep Ensemble~\cite{deepensemble} along with \oursMIMO{1} and \ours, on the Robo3D dataset~\cite{robo3d} which contains corrupted version of SemanticKITTI. All methods are trained on the clean version of SemanticKITTI.

Different from the main paper, we report performance on the `clean' set (\ie, the original SemanticKITTI) to better assess the effect of OOD.
It's also important to note that better calibration may come at the cost of worse performance. 

Hence, \cref{fig:supp:robo3d_uncertainty} presents not only uncertainty metrics (instance ece, SSC ece -- \textit{lower is better}) but also \textit{two performance metrics} (All PQ$^\dagger$ and mIoU -- \textit{higher is better}). 
The latter demonstrates that our better calibration (\cref{fig:supp:robo3d_uncertainty}, top) comes along with better performance (\cref{fig:supp:robo3d_uncertainty}, bottom) in almost all corruptions. This further demonstrates the superiority of \ours compared to both its one-subnet variation, \oursMIMO{1}, and all other baselines.

\paragraph{Qualitative results.}
\cref{fig:supp:qualitative_uncertainty_semkitti} and  \cref{fig:supp:qualitative_uncertainty_sscbenchkitti360} present additional qualitative results of uncertainty estimation on Semantic KITTI and SSCBench-KITTI360 validation sets. 
We also illustrate the uncertainty of ``stuff" class which we omitted for brevity in the main paper. 

Overall, \oursMIMO{1} exhibits higher level of confidence across both voxel and instance uncertainties. Instances of small classes such as person, motorcycle, pole, and traffic light show higher uncertainty levels compared to larger objects like buildings, road and sidewalk. Furthermore, uncertainty tends to increase around the edges of instances, in areas with occluded views, and in regions with lower density of input points.

\subsection{Panoptic Scene Completion (PSC)}
\label{sec:supp:exp_psc}

\paragraph{Quality of pseudo panoptic labels.}
DBSCAN is a classical strategy to approximates panoptic labels when unavailable~\cite{hong2024unified, pnal}.
In the absence of full panoptic ground truth ~(GT), we validate the quality of our pseudo labels against the \textit{single-scan} point-wise panoptic GT of SemKITTI for the val set, voxelizing both and evaluating where both are defined, in~\cref{tab:label-quality}.
This confirms that DBSCAN provides a good approximation of the true labels.

\begin{table}[h!]
	\centering
	\resizebox{1.0\linewidth}{!}{
		\setlength{\tabcolsep}{0.008\linewidth}
		\newcolumntype{H}{>{\setbox0=\hbox\bgroup}c<{\egroup}@{}}
			\begin{tabular}{c|cccccccccc}
				\toprule
				\textbf{labels} & \textbf{All PQ}$^{\dagger}$ & \textbf{All PQ} & \textbf{All SQ} & \textbf{All RQ} & \textbf{Thing PQ} & \textbf{Thing SQ} & \textbf{Thing RQ} & \textbf{Stuff PQ} & \textbf{Stuff SQ} & \textbf{Stuff RQ} \\
				\midrule
				HDBSCAN         & 80.07           & 80.05           & 90.41           & 83.07           & 63.24            & 89.03            & 70.29            & \multirow{2}{*}{\best{91.26}}            & \multirow{2}{*}{\best{91.32}}            & \multirow{2}{*}{\best{91.59}}            \\ 
				DBSCAN (ours)          & \best{88.17}           & \best{88.15}           & \best{92.70}            & \best{90.12}           & \best{83.50}             & \best{94.76}            & \best{87.91}            & \multicolumn{3}{c}{\scriptsize{}note that stuff classes are not clustered}\\ 
				\bottomrule
			\end{tabular}
		}%
		
		\caption{Quality of Pseudo Labels on the SemKITTI (val set)}
		\label{tab:label-quality}
		
	\end{table}
	\noindent{}In \makebox{Table~I of MaskPLS~\cite{maskpls}}, authors demonstrates that clustering with HDBSCAN (a density-aware DBSCAN) leads to reasonable panoptic segmentation on SemKITTI. 
	
	In~\cref{tab:label-quality},
	we find DBSCAN labels to be even more accurate than HDBSCAN since we operate on fairly homogenous density data (aggregation of multiple scans), while MaskPLS clusters single scans exhibiting high-varying density. 
	
\begin{figure}[!h]
		\begin{subfigure}{\linewidth}		
			\centering
			\footnotesize
			\renewcommand{\arraystretch}{0.0}
			\setlength{\tabcolsep}{0.003\textwidth}
			\newcolumntype{P}[1]{>{\centering\arraybackslash}m{#1}}
			\begin{tabular}{ P{0.24\textwidth} P{0.24\textwidth} P{0.24\textwidth} P{0.24\textwidth}}
				\\
				\multicolumn{2}{c}{Large objects } & \multicolumn{2}{c}{Small objects } \\
				\adjincludegraphics[width=\linewidth, trim={.5\width} {.5\height} {.3\width} {.3\height}, clip]{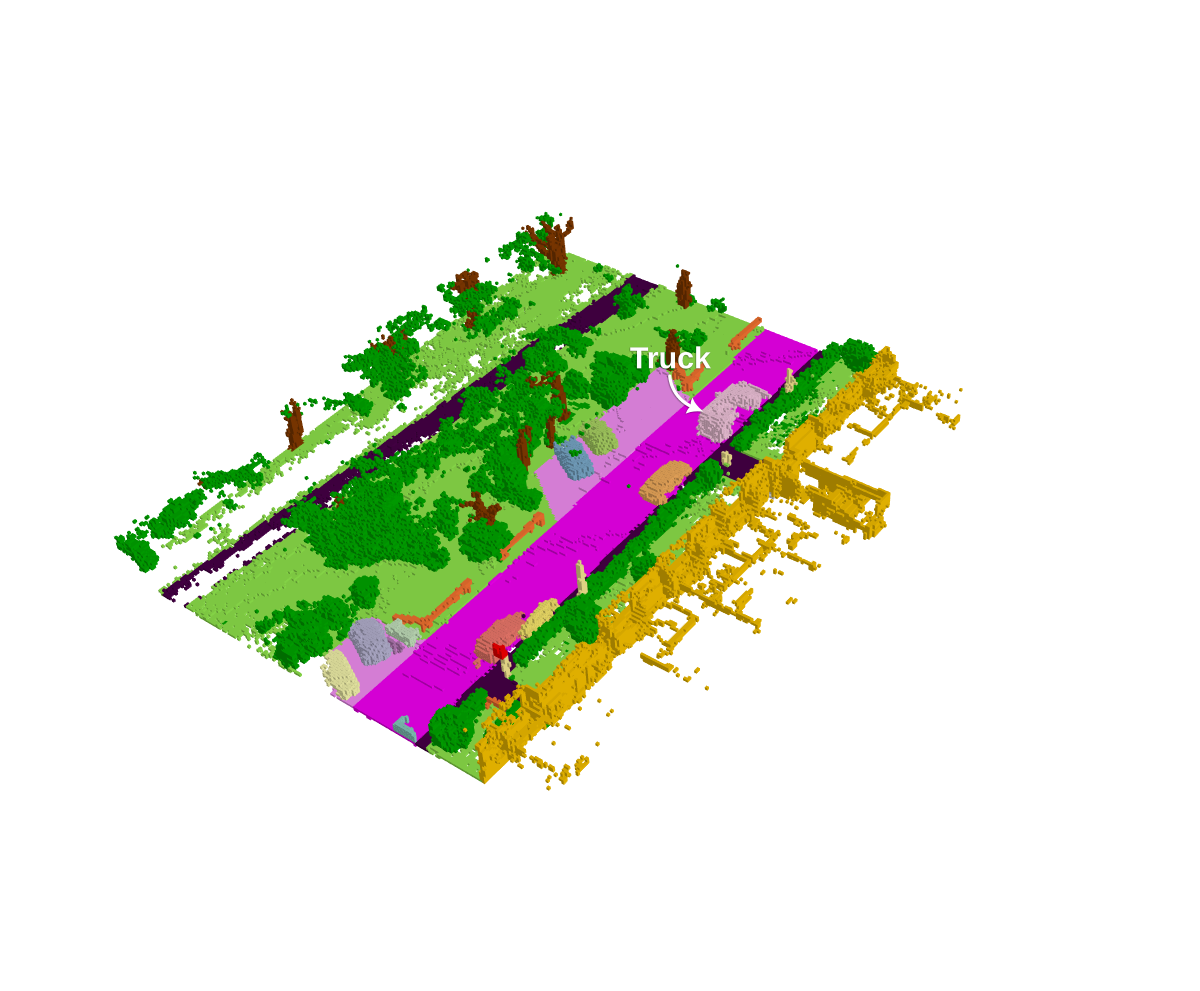} & \adjincludegraphics[width=\linewidth, trim={.3\width} {.25\height} {.45\width} {.50\height}, clip]{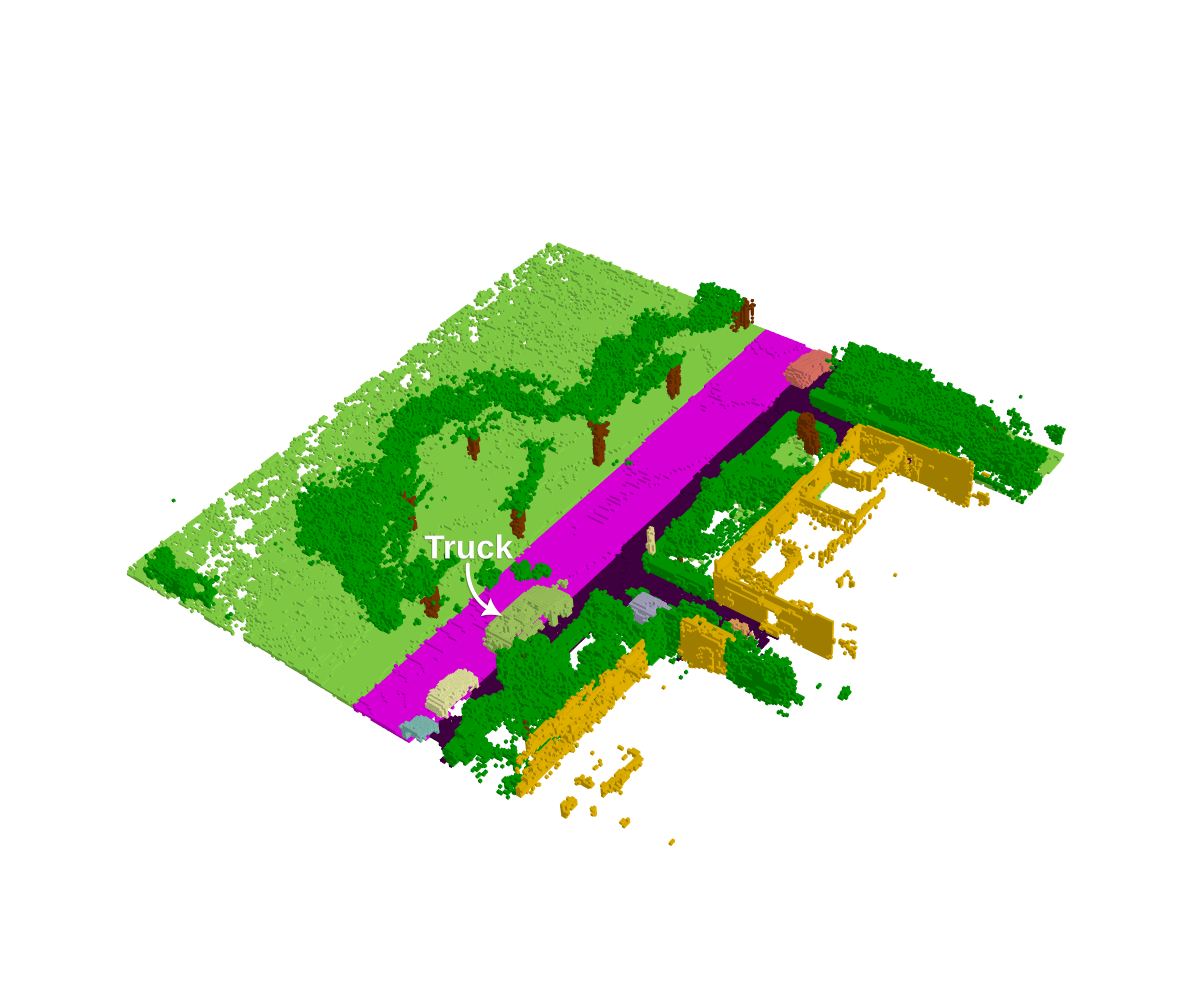} & 
				\adjincludegraphics[width=\linewidth, trim={.6\width} {.4\height} {.2\width} {.4\height}, clip]{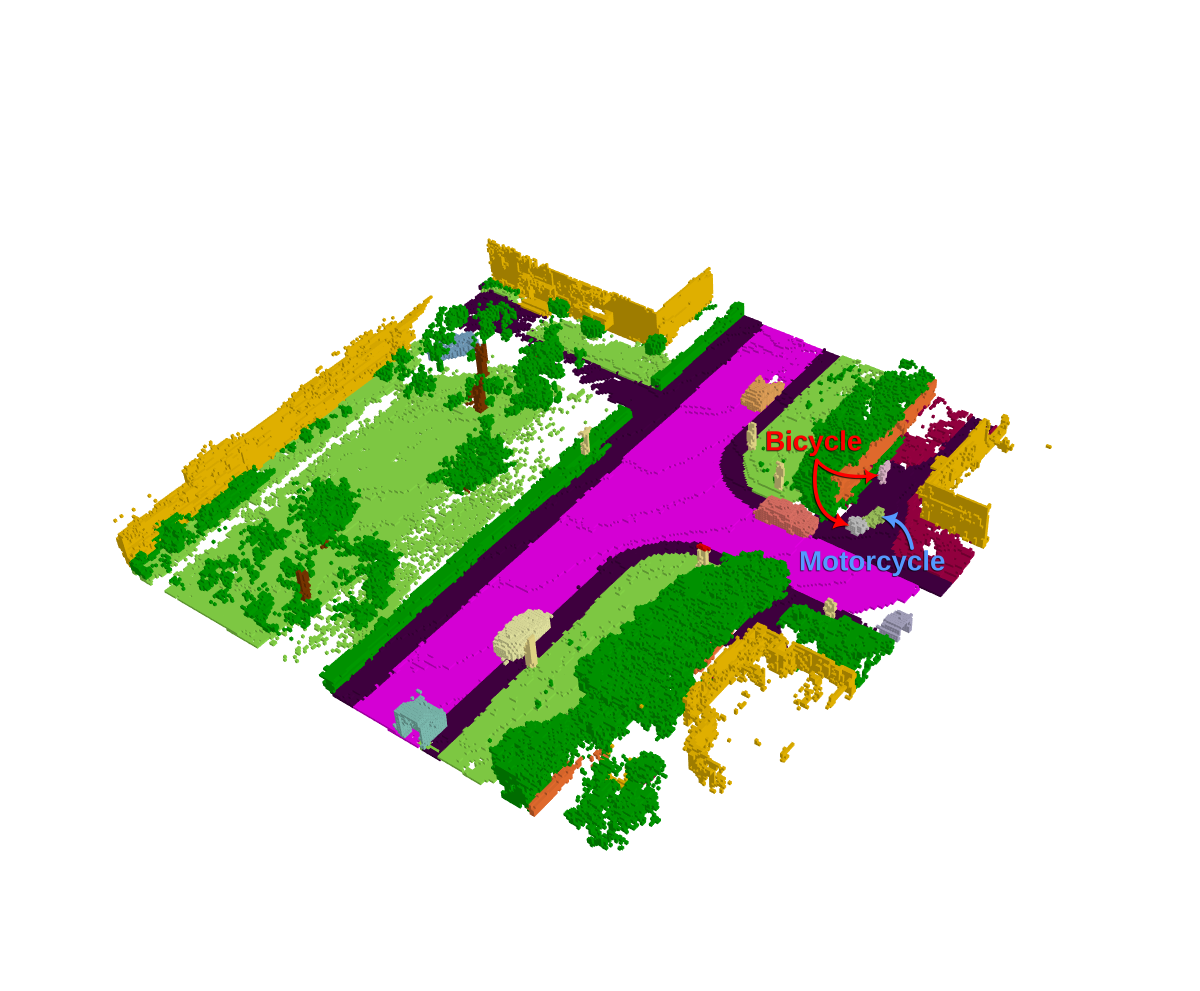} &
				\adjincludegraphics[width=\linewidth, trim={.15\width} {.33\height} {.65\width} {.47\height}, clip]{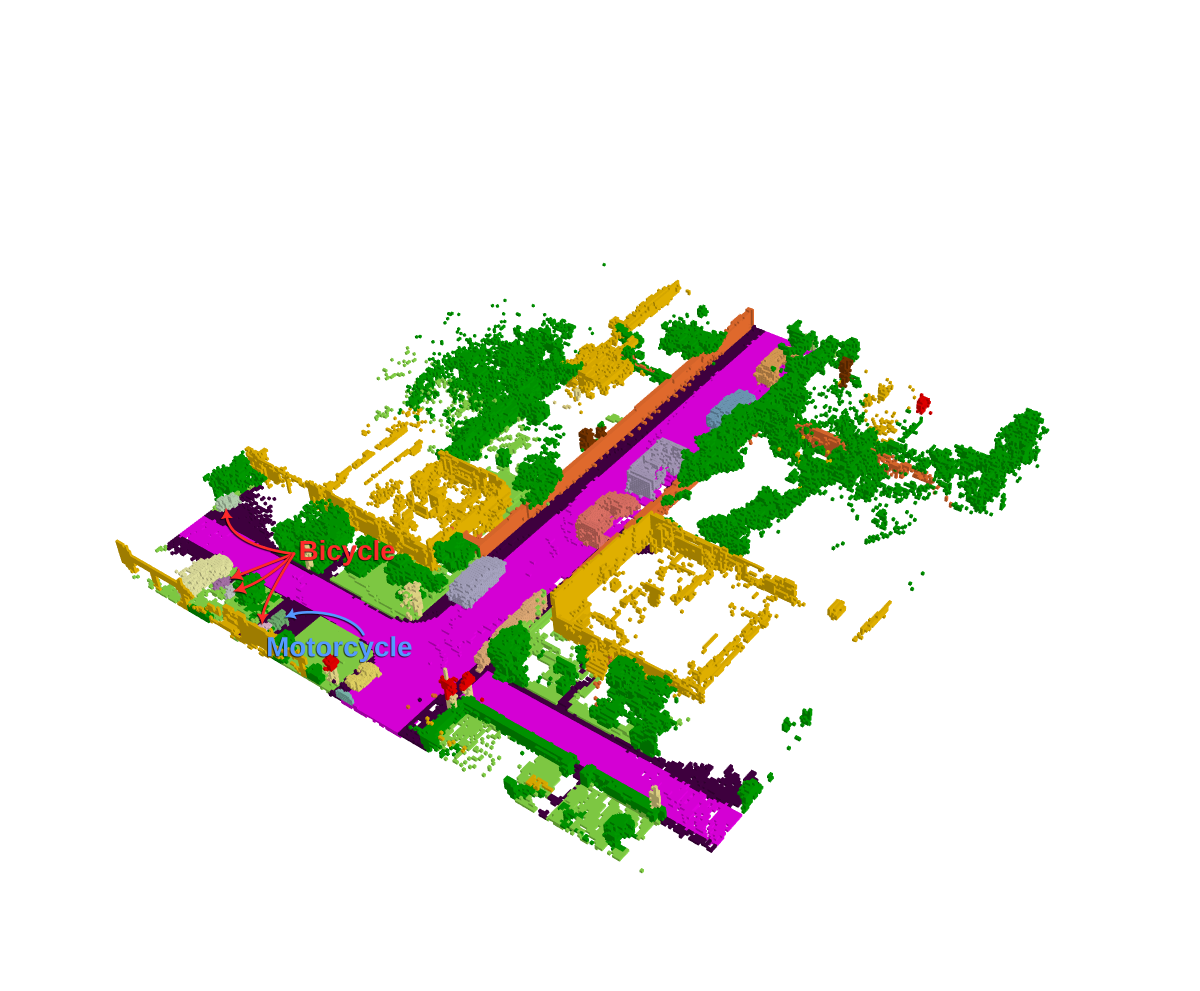} 
				\\
			\end{tabular}			
		\end{subfigure}%
		\caption{Examples of DBSCAN labels}%
		\label{fig:qualitative_label}
	\end{figure}

	From~\cref{fig:qualitative_label}, large objects and small objects are reasonably clustered by DBSCAN.

\paragraph{Class-wise performance.}
\cref{tab:supp:perclass_performance} presents a class-wise comparison of our method, \ours, against baselines on \subref{tab:supp:perclass_semkitti} the Semantic KITTI (val. set) and \subref{tab:supp:perclass_sscbenchkitti360} the SSCBench-KITTI360 (test set). 
For the important PQ metric, \ours outperforms all baselines in most classes on Semantic KITTI, with the exception of the fence category. 
Note that no method successfully predicted the bicyclist and motorcyclist classes. 
The SSCBench-KITTI360 dataset demonstrates its greater complexity with generally lower performance across all classes when compared to Semantic KITTI. Nonetheless, \ours still demonstrates strong performance, ranking first in 10 out of 18 classes and second in 3 out of 18. The effectiveness of \ours is further illustrated in its superior ability to detect masks and produce high-quality masks, as reflected by its first or second highest performance in most classes based on SQ and RQ metrics on both datasets.

\begin{table*}[!t]
	\begin{subtable}{1.0\textwidth}
		\scriptsize
		\setlength{\tabcolsep}{0.004\linewidth}
		\newcommand{\classfreq}[1]{{~\tiny(\semkitfreq{#1}\%)}}  %
		\centering
		\begin{tabular}{l|l|c c c c c c c c c c c c c c c c c c c c}
			\toprule
			& Method
			& \rotatebox{90}{\textcolor{car}{$\blacksquare$} car\classfreq{car}} 
			& \rotatebox{90}{\textcolor{bicycle}{$\blacksquare$} bicycle\classfreq{bicycle}} 
			& \rotatebox{90}{\textcolor{motorcycle}{$\blacksquare$} motorcycle\classfreq{motorcycle}} 
			& \rotatebox{90}{\textcolor{truck}{$\blacksquare$} truck\classfreq{truck}} 
			& \rotatebox{90}{\textcolor{other-vehicle}{$\blacksquare$} other-veh.\classfreq{othervehicle}} 
			& \rotatebox{90}{\textcolor{person}{$\blacksquare$} person\classfreq{person}} 
			& \rotatebox{90}{\textcolor{bicyclist}{$\blacksquare$} bicyclist\classfreq{bicyclist}} 
			& \rotatebox{90}{\textcolor{motorcyclist}{$\blacksquare$} motorcyclist.\classfreq{motorcyclist}} 
			& \rotatebox{90}{\textcolor{road}{$\blacksquare$} road\classfreq{road}} 
			& \rotatebox{90}{\textcolor{parking}{$\blacksquare$} parking\classfreq{parking}} 
			& \rotatebox{90}{\textcolor{sidewalk}{$\blacksquare$} sidewalk\classfreq{sidewalk}}
			& \rotatebox{90}{\textcolor{other-ground}{$\blacksquare$} other-grnd\classfreq{otherground}} 
			& \rotatebox{90}{\textcolor{building}{$\blacksquare$} building\classfreq{building}} 
			& \rotatebox{90}{\textcolor{fence}{$\blacksquare$} fence\classfreq{fence}} 
			& \rotatebox{90}{\textcolor{vegetation}{$\blacksquare$} vegetation\classfreq{vegetation}} 
			& \rotatebox{90}{\textcolor{trunk}{$\blacksquare$} trunk\classfreq{trunk}} 
			& \rotatebox{90}{\textcolor{terrain}{$\blacksquare$} terrain\classfreq{terrain}} 
			& \rotatebox{90}{\textcolor{pole}{$\blacksquare$} pole\classfreq{pole}} 
			& \rotatebox{90}{\textcolor{traffic-sign}{$\blacksquare$} traf.-sign\classfreq{trafficsign}} 
			& \rotatebox{90}{mean}
			\\
			\midrule
			\multirow{5}{*}{\rotatebox{90}{\textbf{PQ}}} & LMSCNet~\cite{lmscnet} + MaskPLS~\cite{maskpls}  & 9.43 & 0.00 & 0.76 & 2.32 & 0.00 & 0.47 & 0.00 & 0.00 & 53.53 & 1.82 & 5.63 & \second{0.00} & 0.26 & 0.19 & 0.00 & 0.27 & 3.52 & 1.00 & 0.00  & 4.17 \\
			& JS3CNet~\cite{js3cnet} + MaskPLS~\cite{maskpls} & 9.57 & 1.07 & 4.19 & 17.54 & 0.91 & 0.12 & 0.00 & 0.00 & 58.45 & 5.32 & 15.89 & \second{0.00} & 1.02 & 1.33 & 0.00 & 0.76 & 13.63 & 0.28 & 0.00 & 6.85 \\
			& SCPNet~\cite{scpnet} + MaskPLS~\cite{maskpls} & \second{18.44} & \second{4.84} & 6.72 & 4.42 & 2.79 & \second{1.81} & 0.00 & 0.00 & 63.89 & 7.92 & 19.92 & \second{0.00} & \second{3.11} & \second{3.28} & 0.13 & 2.29 & 21.55 & 1.99 & 0.17 & 8.59
			\\
			& SCPNet*~\cite{scpnet} + MaskPLS~\cite{maskpls} & 11.72 & 1.80 & \second{14.70} & \second{26.44} & \second{3.83} & 0.33 & 0.00 & 0.00 & \second{66.44} & \second{18.71} & \second{25.29} & \second{0.00} & 2.06 & \best{4.12} & \second{0.39} & \second{3.11} & \second{22.24} & \second{3.97} & \second{1.72} & \second{10.89} \\
			
			& \ours (Ours) & \best{24.55} & \best{7.82} & \best{18.09} & \best{44.89} & \best{11.32} & \best{3.00} & 0.00 & 0.00 & \best{76.22} & \best{28.12} & \best{30.42} & \best{1.33} & \best{4.85} & 0.27 & \best{12.97} & \best{4.22} & \best{32.61} & \best{9.69} & \best{3.26} & \best{16.51} \\
			\midrule
			\multirow{5}{*}{\rotatebox{90}{SQ}} & LMSCNet~\cite{lmscnet} + MaskPLS~\cite{maskpls}  & 62.65 & 0.00 & 53.44 & 53.87 & 0.00 & \best{69.00} & 0.00 & 0.00 & 63.30 & 57.83 & 52.70 & \second{0.00} & 53.93 & 52.58 & 0.00 & \best{59.76} & 54.12 & 53.37 & 0.00 & 36.13 \\
			
			& JS3CNet~\cite{js3cnet} + MaskPLS~\cite{maskpls} & 59.88 & 53.79 & 55.17 & \second{57.73} & 55.70 & \second{62.50} & 0.00 & 0.00 & 65.98 & 55.70 & 54.53 & \second{0.00} & 52.62 & 53.41 & 0.00 & 55.01 & 56.38 & \best{57.68} & 0.00 & 41.90 \\
			
			& SCPNet~\cite{scpnet} + MaskPLS~\cite{maskpls} & \second{66.69} & \second{57.78} & \second{65.30} & 55.30 & \best{65.15} & 61.01 & 0.00 & 0.00 & 68.56 & 58.72 & 55.81 & \second{0.00} & \second{54.94} & \second{54.45} & 51.04 & 55.58 & \second{59.86} & 52.97 & \best{57.14} & \second{49.49} \\
			
			& SCPNet*~\cite{scpnet} + MaskPLS~\cite{maskpls} & 63.98 & 54.46 & 60.54 & 54.03 & 58.55 & 52.31 & 0.00 & 0.00 & \second{70.61} & \second{59.25} & \second{56.69} & \second{0.00} & 53.61 & \best{55.70} & \second{52.84} & 56.05 & 58.76 & 53.55 & \second{56.62} & 48.29\\
			
			& \ours (Ours) & \best{70.10} & \best{57.84} & \best{67.00} & \best{67.33} & \second{62.15} & 60.14 & 0.00 & 0.00 & \best{77.52} & \best{62.62} & \best{59.95} & \best{54.71} & \best{55.87} & 51.29 & \best{52.85} & \second{57.50} & \best{63.88} & \second{54.78} & 55.17 & \best{54.25}
			\\
			\midrule
			\multirow{5}{*}{\rotatebox{90}{RQ}} & LMSCNet~\cite{lmscnet} + MaskPLS~\cite{maskpls}  & 15.05 & 0.00 & 1.42 & 4.30 & 0.00 & 0.67 & 0.00 & 0.00 & 84.56 & 3.15 & 10.69 & \second{0.00} & 0.48 & 0.37 & 0.00 & 0.45 & 6.50 & 1.87 & 0.00 & 6.82 \\
			
			& JS3CNet~\cite{js3cnet} + MaskPLS~\cite{maskpls} & 15.98 & 2.00 & 7.59 & 30.38 & 1.63 & 0.20 & 0.00 & 0.00 & 88.59 & 9.55 & 29.14 & \second{0.00} & 1.93 & 2.49 & 0.00 & 1.39 & 24.17 & 0.48 & 0.00 & 11.34 \\
			
			& SCPNet~\cite{scpnet} + MaskPLS~\cite{maskpls} & \second{27.65} & \second{8.38} & 10.29 & 8.00 & 4.28 & \second{2.96} & 0.00 & 0.00 & 93.18 & 13.50 & 35.69 & \second{0.00} & \second{5.66} & \second{6.03} & 0.25 & 4.12 & 36.00 & 3.76 & 0.30  & 13.69
			\\
			
			& SCPNet*~\cite{scpnet} + MaskPLS~\cite{maskpls} & 18.32 & 3.31 & \second{24.29} & \second{48.94} & \second{6.54} & 0.63 & 0.00 & 0.00 & \second{94.09} & \second{31.58} & \second{44.61} & \second{0.00} & 3.84 & \best{7.39} & \second{0.73} & \second{5.56} & \second{37.84} & \second{7.42} & \second{3.04} & \second{17.80}\\
			
			& \ours (Ours) & \best{35.03} & \best{13.51} & \best{27.00} & \best{66.67} & \best{18.21} & \best{4.98} & 0.00 & 0.00 & \best{98.32} & \best{44.91} & \best{50.73} & \best{2.44} & \best{8.69} & 0.52 & \best{24.54} & \best{7.33} & \best{51.05} & \best{17.70} & \best{5.91} & \best{25.13}\\
			\bottomrule
		\end{tabular}
		\caption{Semantic KITTI (val. set)}
		\label{tab:supp:perclass_semkitti}
	\end{subtable}
	\begin{subtable}{1.0\textwidth}
		\scriptsize
		\setlength{\tabcolsep}{0.004\linewidth}
		\newcommand{\classfreq}[1]{{~\tiny(\kittithreesixtyfreq{#1}\%)}}  %
		\centering
		\begin{tabular}{l|l|c c c c c c c c c c c c c c c c c c c c}
			\toprule
			& Method
			& \rotatebox{90}{\textcolor{car}{$\blacksquare$} car\classfreq{car}} 
			& \rotatebox{90}{\textcolor{bicycle}{$\blacksquare$} bicycle\classfreq{bicycle}} 
			& \rotatebox{90}{\textcolor{motorcycle}{$\blacksquare$} motorcycle\classfreq{motorcycle}} 
			& \rotatebox{90}{\textcolor{truck}{$\blacksquare$} truck\classfreq{truck}} 
			& \rotatebox{90}{\textcolor{other-vehicle}{$\blacksquare$} other-veh.\classfreq{othervehicle}} 
			& \rotatebox{90}{\textcolor{person}{$\blacksquare$} person\classfreq{person}}
			& \rotatebox{90}{\textcolor{road}{$\blacksquare$} road\classfreq{road}} 
			& \rotatebox{90}{\textcolor{parking}{$\blacksquare$} parking\classfreq{parking}} 
			& \rotatebox{90}{\textcolor{sidewalk}{$\blacksquare$} sidewalk\classfreq{sidewalk}}
			& \rotatebox{90}{\textcolor{other-ground}{$\blacksquare$} other-grnd\classfreq{otherground}} 
			& \rotatebox{90}{\textcolor{building}{$\blacksquare$} building\classfreq{building}} 
			& \rotatebox{90}{\textcolor{fence}{$\blacksquare$} fence\classfreq{fence}} 
			& \rotatebox{90}{\textcolor{vegetation}{$\blacksquare$} vegetation\classfreq{vegetation}} 
			& \rotatebox{90}{\textcolor{terrain}{$\blacksquare$} terrain\classfreq{terrain}} 
			& \rotatebox{90}{\textcolor{pole}{$\blacksquare$} pole\classfreq{pole}} 
			& \rotatebox{90}{\textcolor{traffic-sign}{$\blacksquare$} traf.-sign\classfreq{trafficsign}} 
			& \rotatebox{90}{\textcolor{other-struct}{$\blacksquare$} other-structure\classfreq{otherstruct}} 
			& \rotatebox{90}{\textcolor{other-object}{$\blacksquare$} other-object\classfreq{otherobject}} 
			& \rotatebox{90}{mean}
			\\
			\midrule
			\multirow{5}{*}{\rotatebox{90}{\textbf{PQ}}} & LMSCNet~\cite{lmscnet} + MaskPLS~\cite{maskpls}  &  4.64 & \second{0.00} & 0.00 & 0.00 & 0.00 & 0.67 & 49.87 & 0.31 & 5.75 & 0.00 & 0.00 & 0.00 & \second{0.00} & 1.59 & 0.13 & 4.24 & 0.00 & 0.00 & 4.14 \\
			
			& JS3CNet~\cite{js3cnet} + MaskPLS~\cite{maskpls} & 13.77 & \second{0.00} & 0.81 & 3.58 & 0.48 & 1.50 & \second{63.13} & \second{1.63} & 23.99 & \second{0.12} & \second{0.14} & 0.19 & \second{0.00} & 4.36 & 1.51 & 6.55 & 0.09 & \second{0.44} & 6.79 \\
			
			& SCPNet~\cite{scpnet} + MaskPLS~\cite{maskpls} & \best{17.77} & \second{0.00} & \best{2.56} & 1.45 & \second{1.69} & \best{1.91} & 46.77 & 0.54 & 22.06 & 0.04 & \second{0.14} & 0.37 & \second{0.00} & \second{5.28} & 2.02 & 7.48 & \best{0.14} & 0.29 & 6.14 \\
			
			& SCPNet*~\cite{scpnet} + MaskPLS~\cite{maskpls} & \second{15.87} & \second{0.00} & 1.53 & \second{3.80} & 0.97 & \second{1.72} & 57.55 & \best{3.34} & \second{30.65} & \best{0.16} & 0.04 & \best{0.68} & \second{0.00} & \best{5.66} & \second{2.46} & \second{9.62} & 0.08 & 0.26 & \second{7.47} \\
			
			& \ours (Ours) &  14.85 & \best{0.11} & \second{1.73} & \best{9.01} & \best{1.97} & 1.60 & \best{72.05} & 1.47 & \best{35.82} & 0.00 & \best{24.29} & \second{0.53} & \best{8.96} & 5.07 & \best{3.15} & \best{14.47} & \second{0.11} & \best{1.38} & \best{10.92} \\
			
			\midrule
			\multirow{5}{*}{\rotatebox{90}{SQ}} & LMSCNet~\cite{lmscnet} + MaskPLS~\cite{maskpls}  & 54.47 & \second{0.00} & 0.00 & 0.00 & 0.00 & \best{67.99} & 66.89 & \second{58.36} & 53.46 & 0.00 & 0.00 & 0.00 & \second{0.00} & 54.17 & \second{56.27} & \second{65.93} & 0.00 & 0.00 & 26.52 \\
			
			& JS3CNet~\cite{js3cnet} + MaskPLS~\cite{maskpls} & 58.34 & \second{0.00} & \second{59.77} & 52.01 & 52.67 & \second{67.64} & \second{69.77} & 56.09 & 57.48 & \second{53.95} & 52.22 & 53.28 & \second{0.00} & 54.62 & 54.94 & 64.73 & \second{55.20} & \second{58.21} &  51.16 \\
			
			& SCPNet~\cite{scpnet} + MaskPLS~\cite{maskpls} & \best{60.89} & \second{0.00} & \best{61.42} & 52.88 & \best{57.20} & 58.39 & 63.06 & 55.22 & 56.97 & \best{55.56} & \second{52.45} & 55.16 & \second{0.00} & 54.96 & 54.54 & 65.31 & \best{61.72} & 55.60 & \second{51.18} \\
			
			& SCPNet*~\cite{scpnet} + MaskPLS~\cite{maskpls} & \second{59.32} & \second{0.00} & 54.73 & \best{55.96} & 53.73 & 65.01 & 69.17 & \best{59.13} & \second{58.09} & 52.13 & 50.13 & \second{55.66} & \second{0.00} & \second{55.33} & 55.10 & 63.70 & 52.62 & 52.18 & 50.67 \\

			& \ours (Ours) & 58.22 & \best{53.03} & 57.14 & \second{55.89} & \second{55.53} & 65.39 & \best{76.15} & 58.04 & \best{58.51} & 0.00 & \best{56.18} & \best{61.65} & \best{55.28} & \best{57.99} & \best{56.83} & \best{67.85} & 54.74 & \best{61.42} & \best{56.10} \\
			
			\midrule
			\multirow{5}{*}{\rotatebox{90}{RQ}} & LMSCNet~\cite{lmscnet} + MaskPLS~\cite{maskpls}  & 8.51 & \second{0.00} & 0.00 & 0.00 & 0.00 & 0.99 & 74.56 & 0.53 & 10.75 & 0.00 & 0.00 & 0.00 & \second{0.00} & 2.94 & 0.24 & 6.43 & 0.00 & 0.00 & 6.45 \\
			
			& JS3CNet~\cite{js3cnet} + MaskPLS~\cite{maskpls} &  23.61 & \second{0.00} & 1.36 & \second{6.88} & 0.91 & 2.22 & \second{90.49} & \second{2.91} & 41.72 & \second{0.22} & \second{0.28} & 0.36 & \second{0.00} & 7.98 & 2.75 & 10.11 & 0.15 & \second{0.76} & 10.71 \\
			
			& SCPNet~\cite{scpnet} + MaskPLS~\cite{maskpls} & \best{29.18} & \second{0.00} & \best{4.17} & 2.75 & \second{2.95} & \best{3.27} & 74.16 & 0.98 & 38.72 & 0.08 & \second{0.28} & 0.67 & \second{0.00} & \second{9.60} & 3.71 & 11.45 & \best{0.23} & 0.53 & 10.15 \\
			
			& SCPNet*~\cite{scpnet} + MaskPLS~\cite{maskpls} & 26.76 & \second{0.00} & 2.79 & 6.79 & 1.81 & \second{2.65} & 83.20 & \best{5.65} & \second{52.77} & \best{0.30} & 0.09 & \best{1.23} & \second{0.00} & \best{10.24} & \second{4.47} & \second{15.10} & 0.16 & 0.50 & \second{11.92} \\
			
			& \ours (Ours) & 25.51 & \best{0.21} & \second{3.03} & \best{16.13} & \best{3.55} & 2.45 & \best{94.62} & 2.53 & \best{61.22} & 0.00 & \best{43.24} & \second{0.86} & \best{16.21} & 8.75 & \best{5.54} & \best{21.33} & \second{0.19} & \best{2.24} & \best{17.09} \\
			\bottomrule
		\end{tabular}\\
		\caption{SSCBench-KITTI360 (test set)}
		\label{tab:supp:perclass_sscbenchkitti360}
	\end{subtable}
	\caption{\textbf{Class-wise performance on \subref{tab:supp:perclass_semkitti} Semantic KITTI (val. set) and \subref{tab:supp:perclass_sscbenchkitti360} SSCBench-KITTI360 (test set) for Panoptic Scene Completion.} We report the performance of our method and baselines for each class across the two datasets. Our approach exceeds the performance of baseline methods in most classes, particularly in the crucial PQ metric. The SSCBench-KITTI360 dataset exhibits a higher level of complexity compared to Semantic KITTI, as evidenced by its overall lower performance metrics. \ours also shows superior performance in SQ and RQ by being either first or second in most classes.}
	\label{tab:supp:perclass_performance}
\end{table*}

\begin{figure*}
	\centering
	\setlength{\fboxsep}{0.008pt}
	\newcommand{\tmpframe}[1]{\fbox{#1}}
	\newcommand{\im}[6]{\stackinset{r}{-0.0cm}{b}{0pt}{\tmpframe{\adjincludegraphics[width=#6\linewidth, trim={#2\width} {#3\height} {#4\width} {#5\height}, clip]{#1}}}{\adjincludegraphics[width=\linewidth, trim={.2\width} {.2\height} {.2\width} {.2\height}, clip]{#1}}}
	
	\newcommand{\imv}[9]{\stackinset{r}{-0.0cm}{b}{0pt}{\tmpframe{\adjincludegraphics[width=0.5\linewidth, trim={#2\width} {#3\height} {#4\width} {#5\height}, clip]{#1}}}{\adjincludegraphics[width=\linewidth, trim={#6\width} {#7\height} {#8\width} {#9\height}, clip]{#1}}}
	
	\newcommand{\imleftzoom}[6]{\stackinset{l}{0.0cm}{b}{0.02cm}{\tmpframe{\adjincludegraphics[width=#6\linewidth, trim={#2\width} {#3\height} {#4\width} {#5\height}, clip]{#1}}}{\adjincludegraphics[width=\linewidth, trim={.2\width} {.2\height} {.3\width} {.3\height}, clip]{#1}}}
	
	\newcommand{\imtopleftzoom}[6]{\stackinset{l}{0.0cm}{t}{0.0cm}{\tmpframe{\adjincludegraphics[width=#6\linewidth, trim={#2\width} {#3\height} {#4\width} {#5\height}, clip]{#1}}}{\adjincludegraphics[width=\linewidth, trim={.2\width} {.2\height} {.3\width} {.3\height}, clip]{#1}}}
	\newcommand{\imfirst}[1]{\imtopleftzoom{#1}{0.56}{0.51}{0.33}{0.36}{0.48}}
	\newcommand{\imsecond}[1]{\im{#1}{0.58}{0.58}{0.32}{0.31}{0.49}}
	\newcommand{\imthird}[1]{\imleftzoom{#1}{0.40}{0.44}{0.50}{0.44}{0.48}}
	\newcommand{\imforth}[1]{\im{#1}{0.55}{0.56}{0.36}{0.35}{0.48}}
	\newcommand{\imfive}[1]{\imtopleftzoom{#1}{0.51}{0.40}{0.39}{0.48}{0.48}}
	\newcommand{\imsix}[1]{\imtopleftzoom{#1}{0.49}{0.42}{0.36}{0.40}{0.56}}
	
		\resizebox{0.9\linewidth}{!}{
			\centering
			\newcolumntype{P}[1]{>{\centering\arraybackslash}m{#1}}
			\setlength{\tabcolsep}{0.001\textwidth}
			\renewcommand{\arraystretch}{0.8}
			\footnotesize
			\begin{tabular}{cP{0.15\textwidth} P{0.15\textwidth} P{0.15\textwidth} P{0.15\textwidth} P{0.15\textwidth} P{0.15\textwidth}}		
				&Input & LMSCNet + MaskPLS & JS3CNet + MaskPLS & SCPNet* + MaskPLS & \ours~(ours) & ground truth 
				\\
				
				\multirow{1}{*}{\rotatebox{90}{\textbf{Semantic KITTI} (val set)\hspace{-0.3em}}}
				&
				
				\imfirst{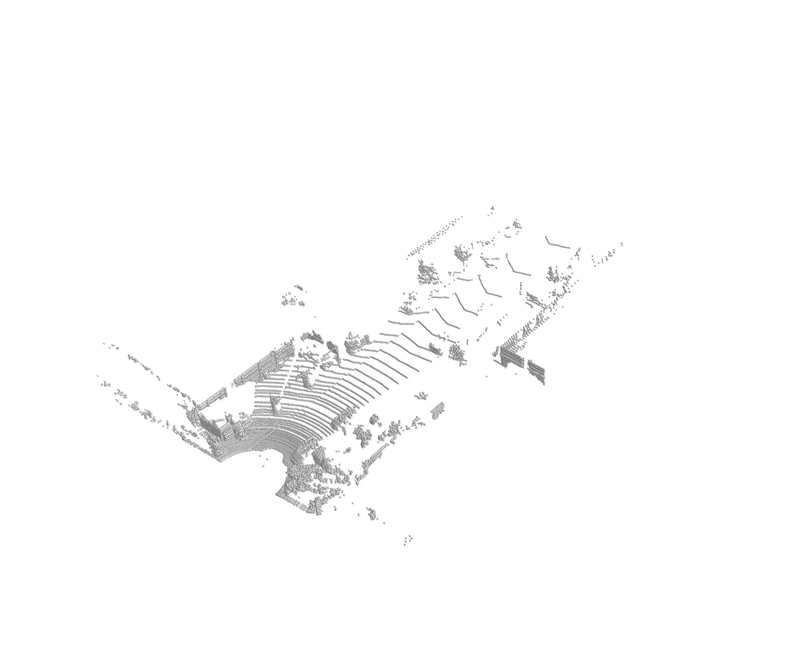} &
				\imfirst{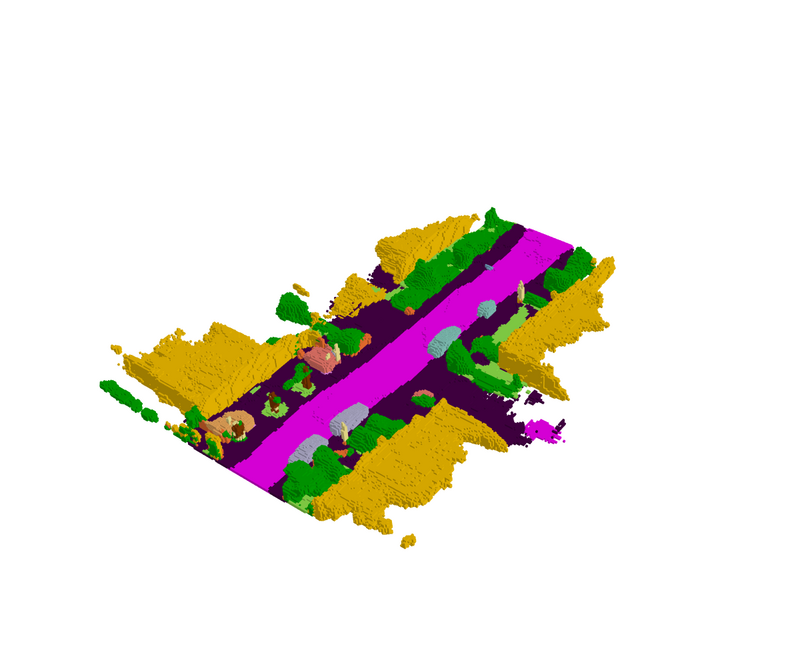} &
				\imfirst{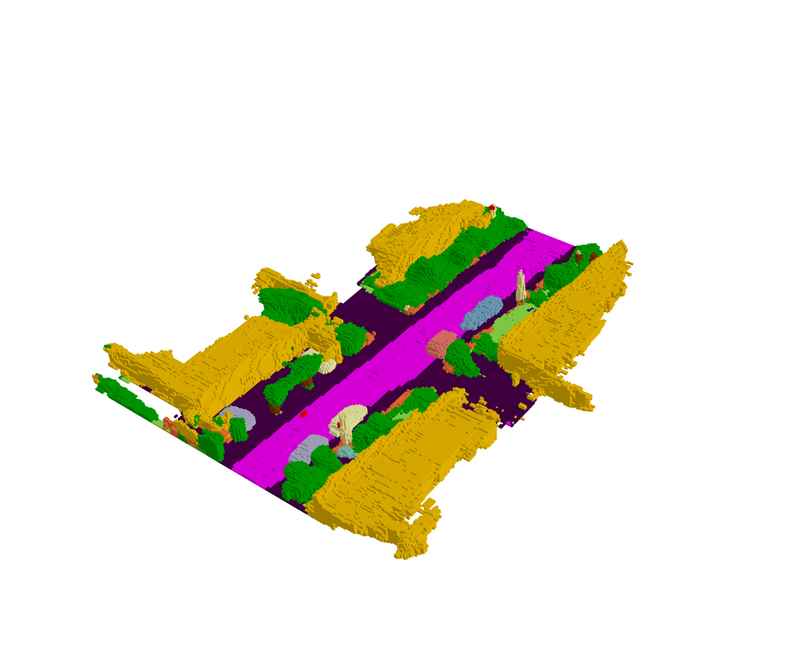} &
				\imfirst{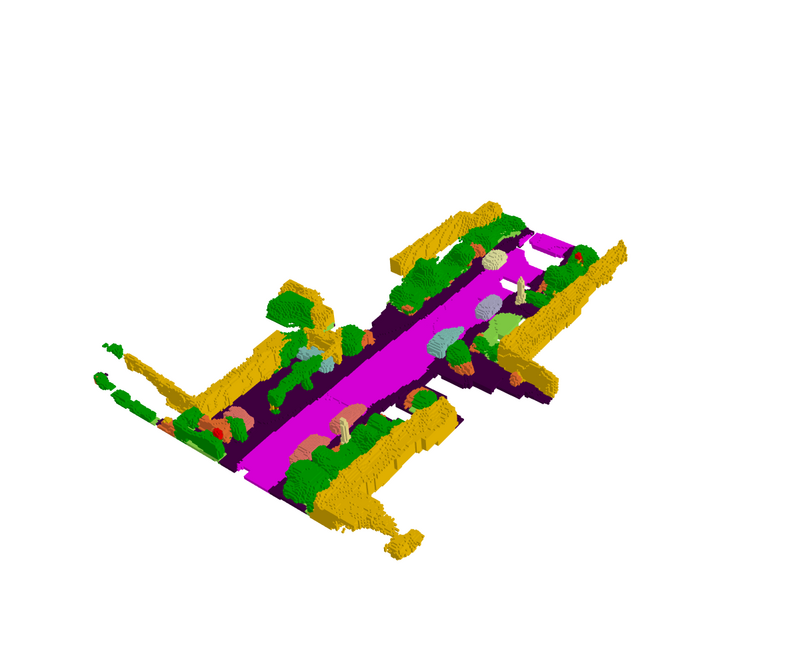} &
				\imfirst{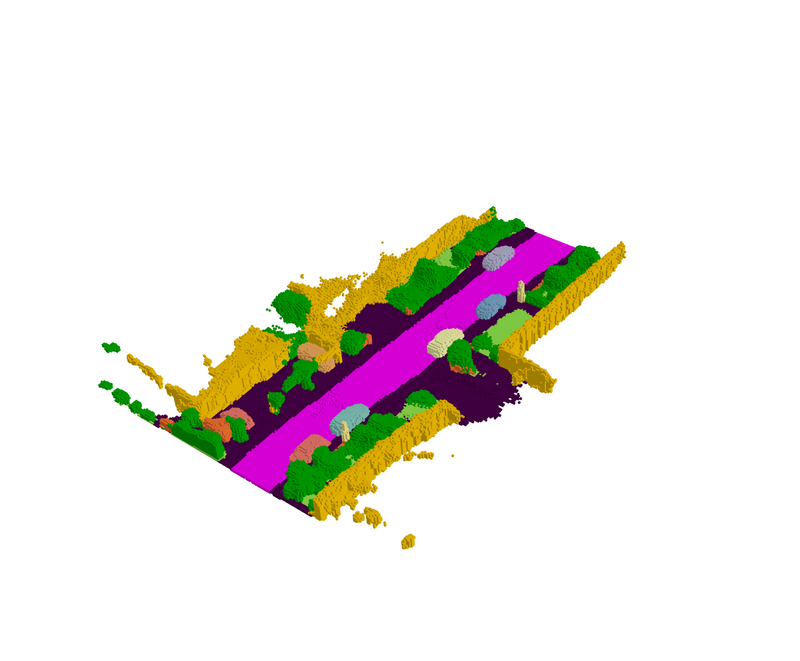} &
				\imfirst{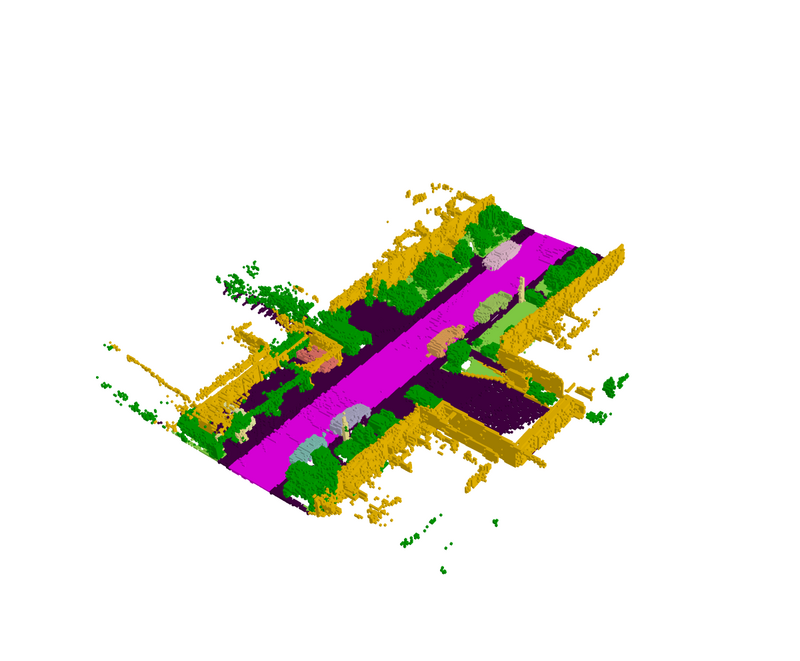}
				\\[-0.3em]

				&
				\imsix{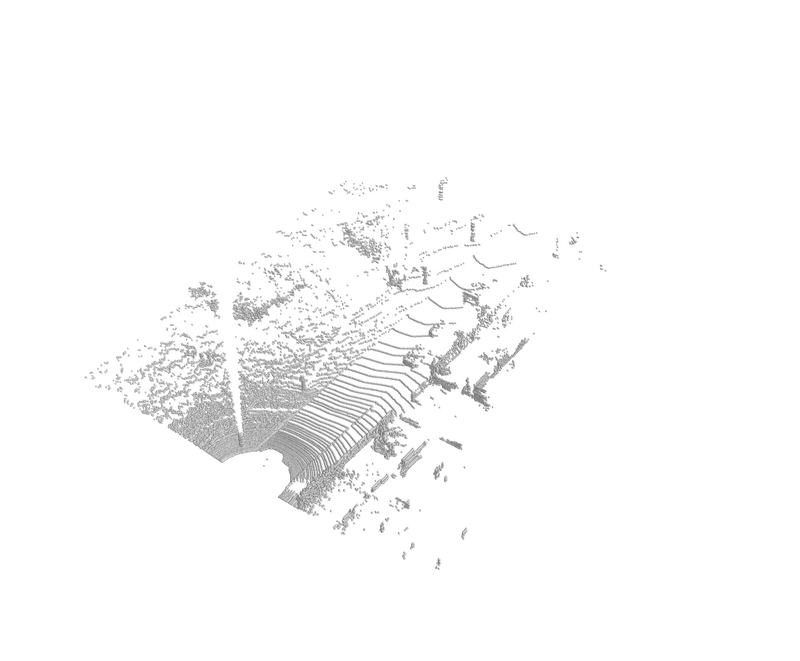} &
				\imsix{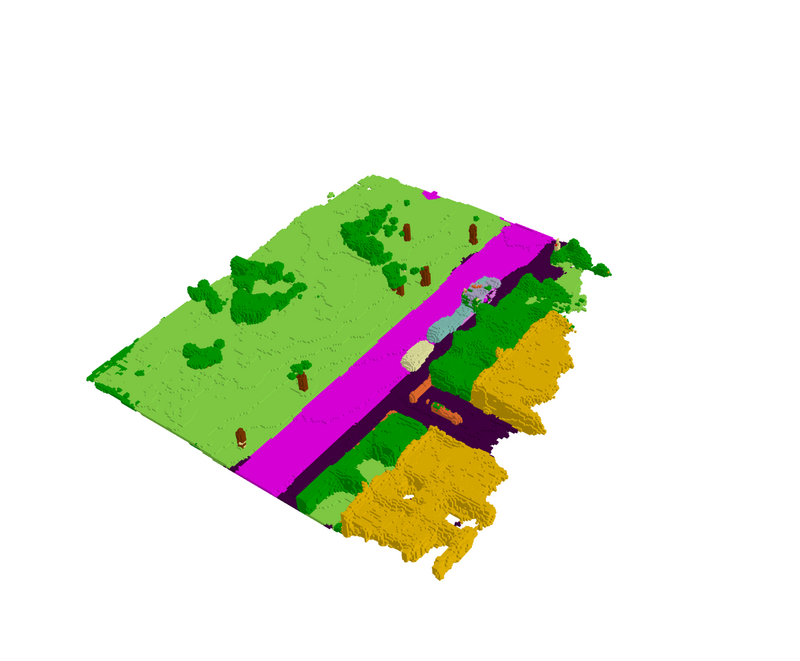} &
				\imsix{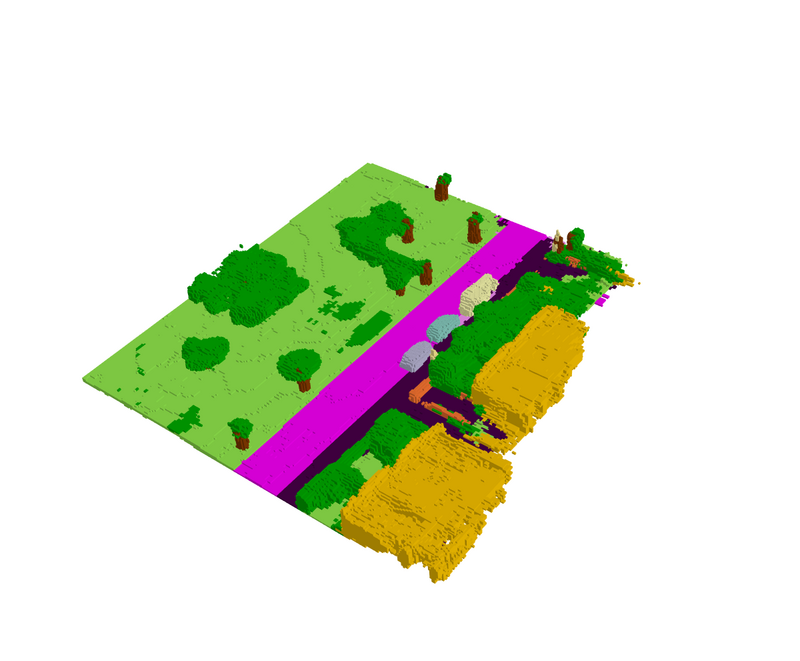} &
				\imsix{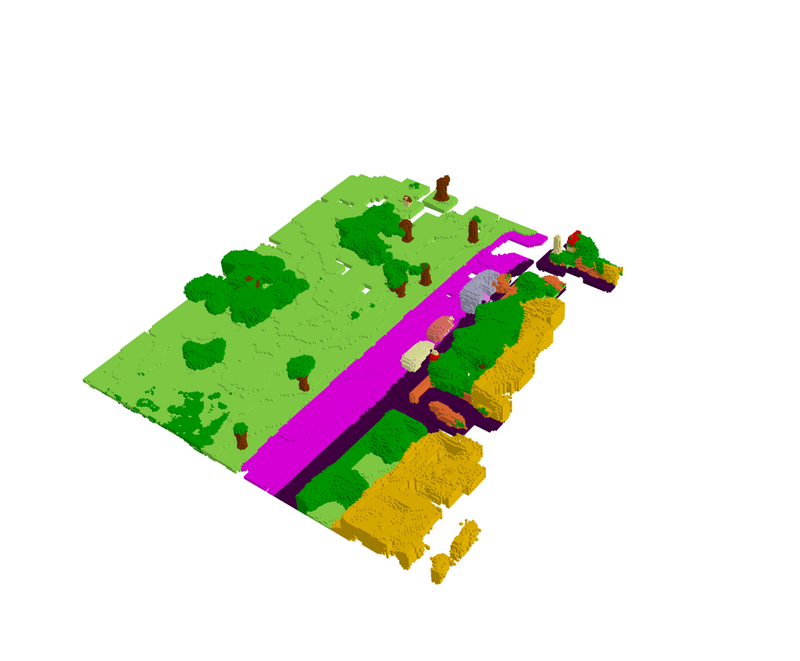} &
				\imsix{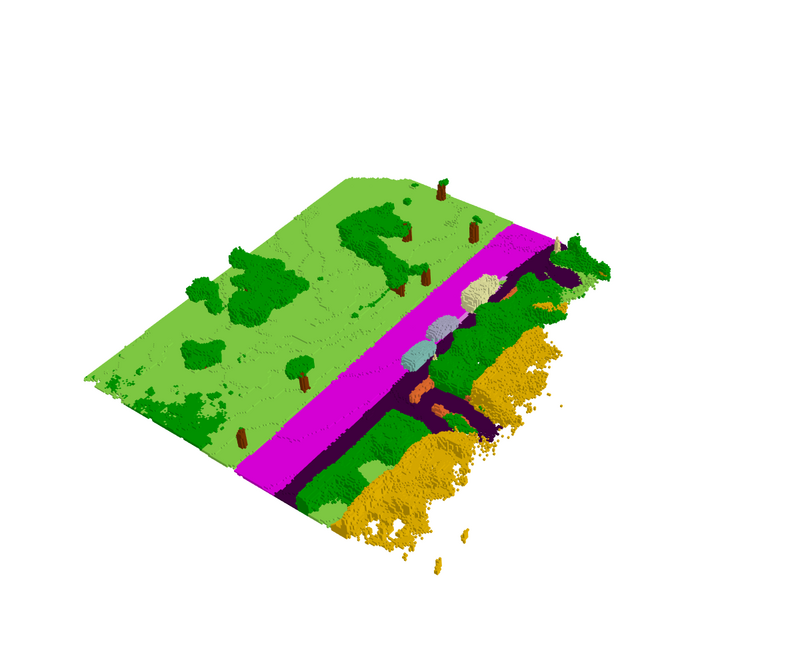} &
				\imsix{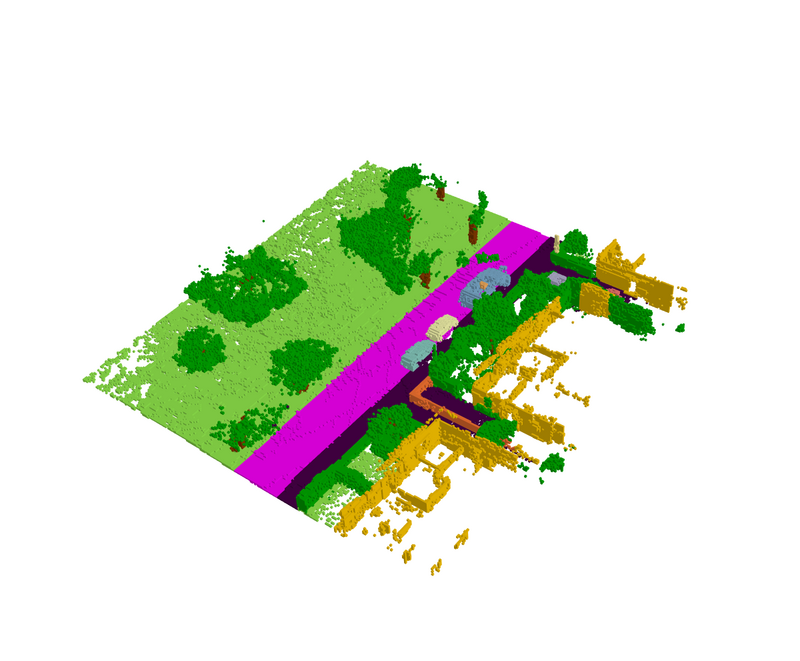}
				\\[-0.3em]
				&
				\imsecond{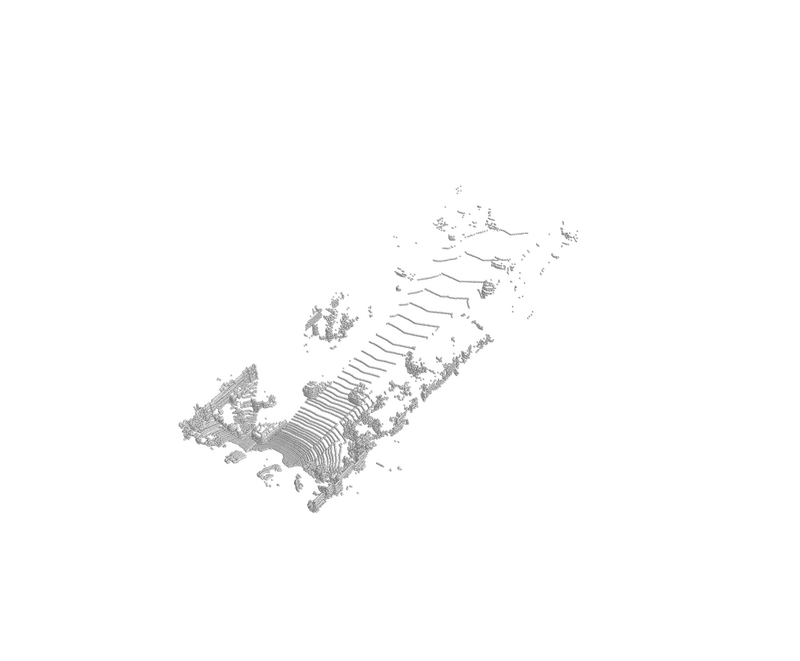} &
				\imsecond{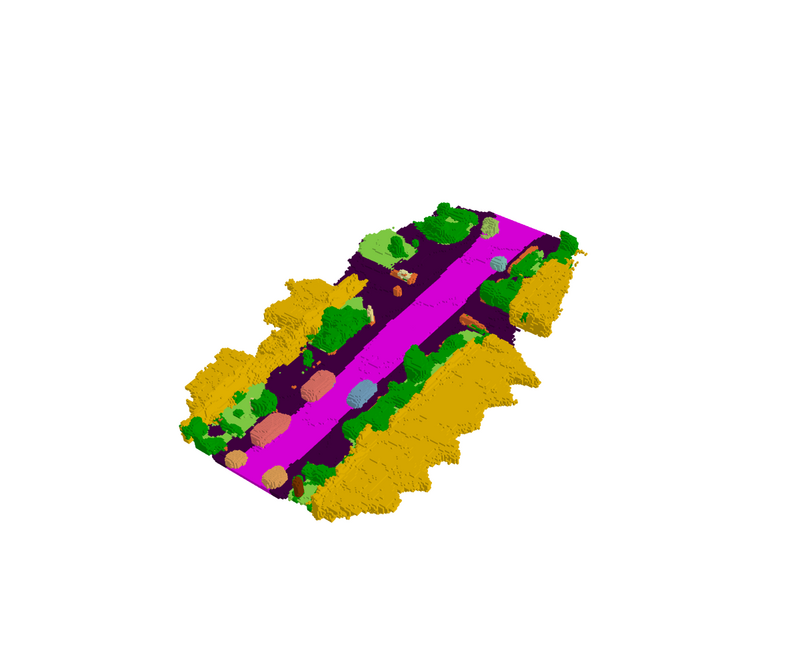} &
				\imsecond{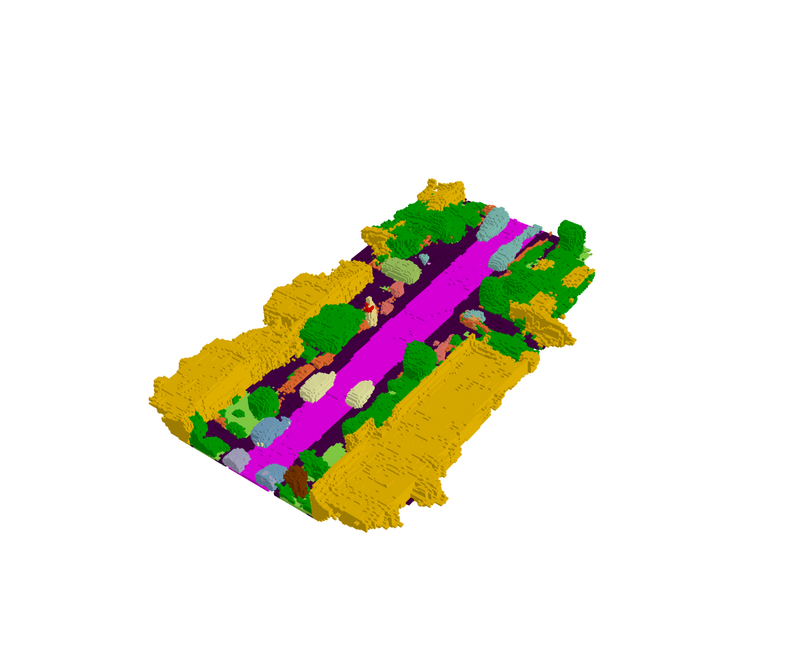} &
				\imsecond{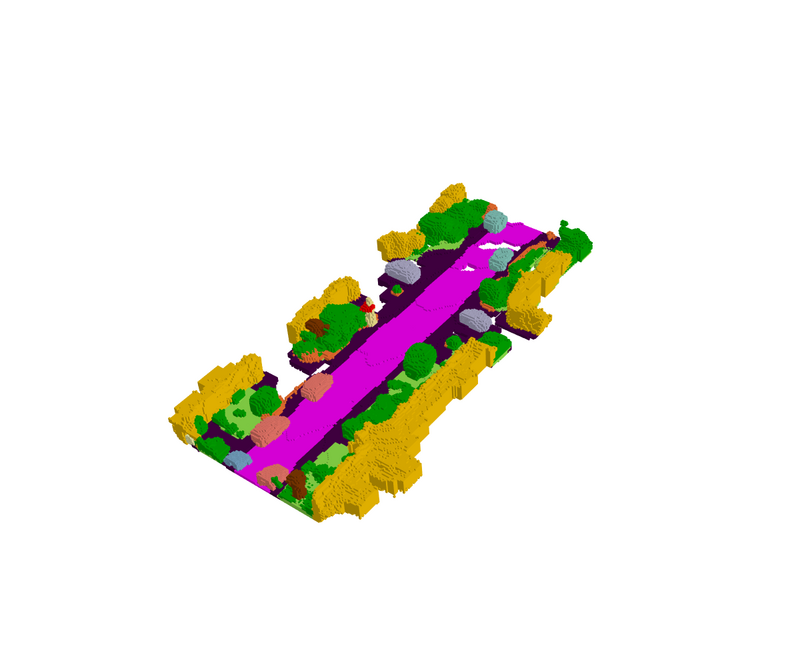} &
				\imsecond{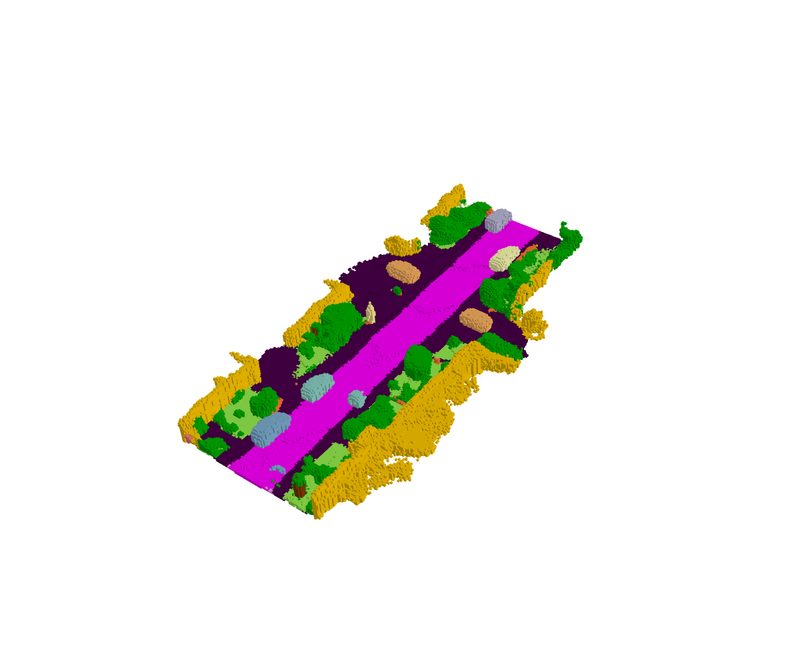} &
				\imsecond{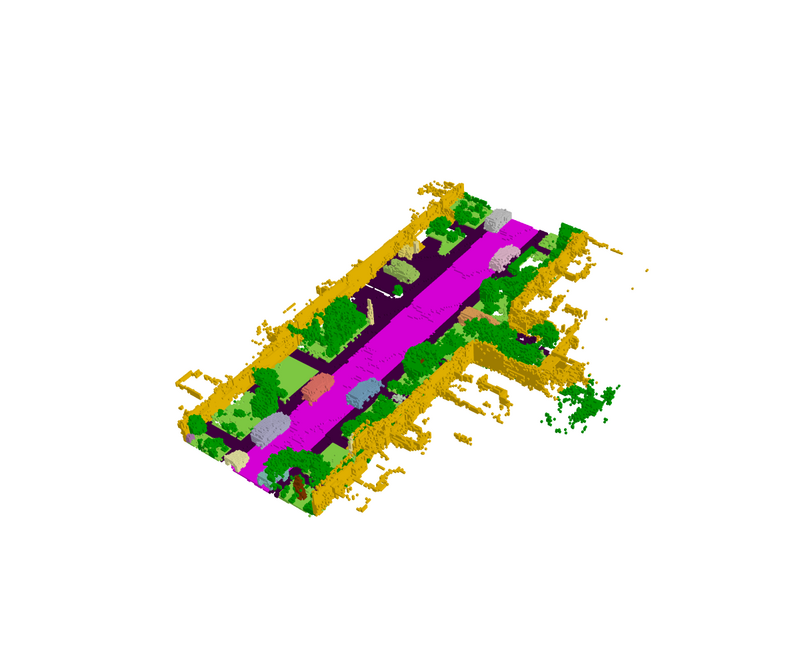}
				\\[-0.3em]
				\midrule
				
				\multirow{1}{*}{\rotatebox{90}{\textbf{SSCBench-KITTI360} (val set)\hspace{-2.3em}}}
				& \imthird{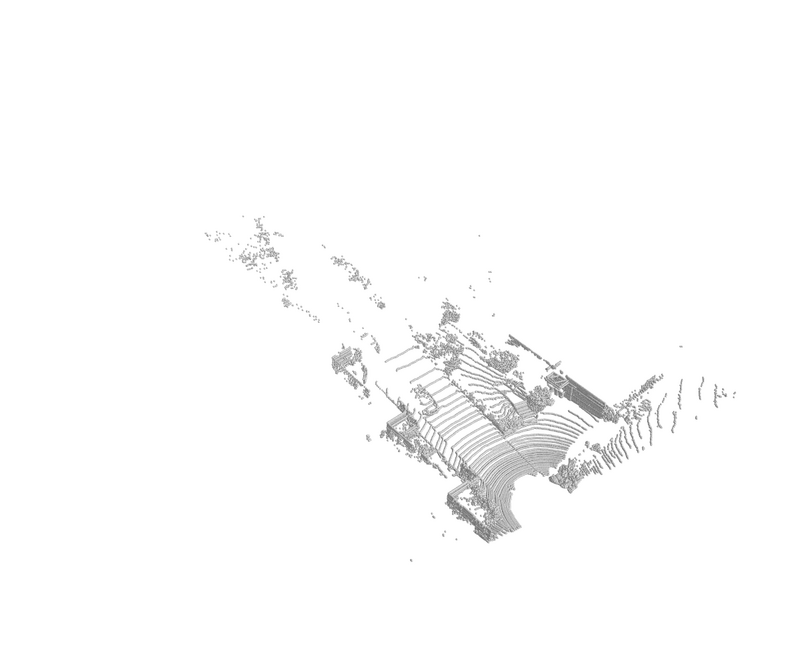} 
				& \imthird{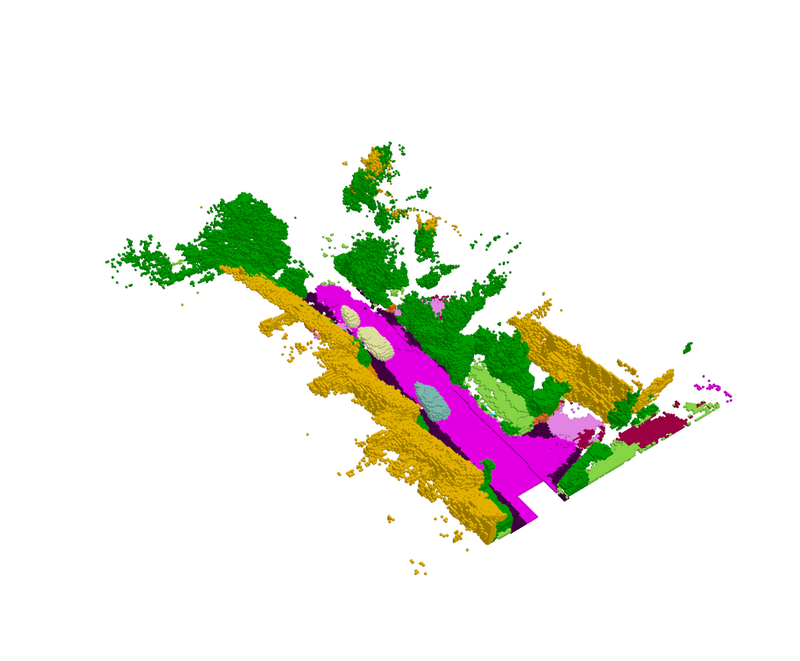} 
				& \imthird{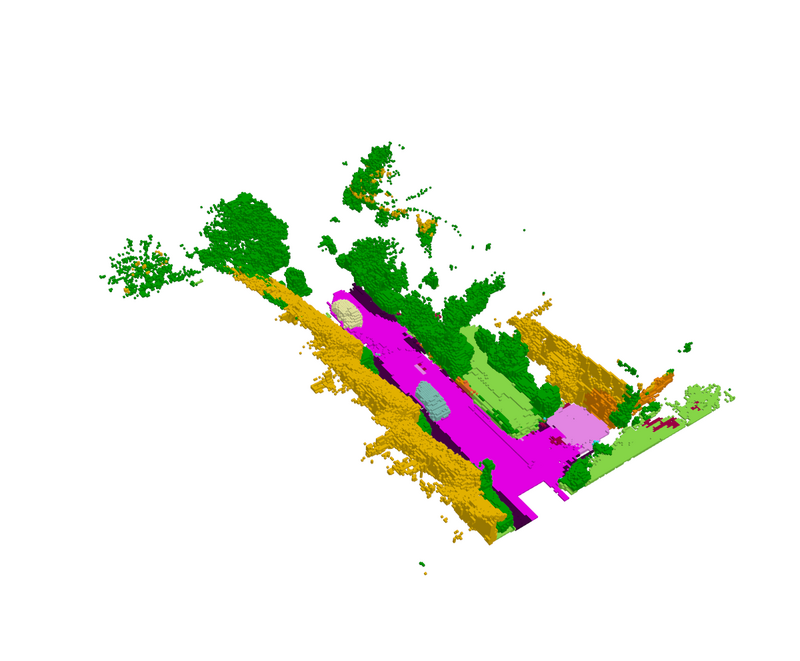} 
				& \imthird{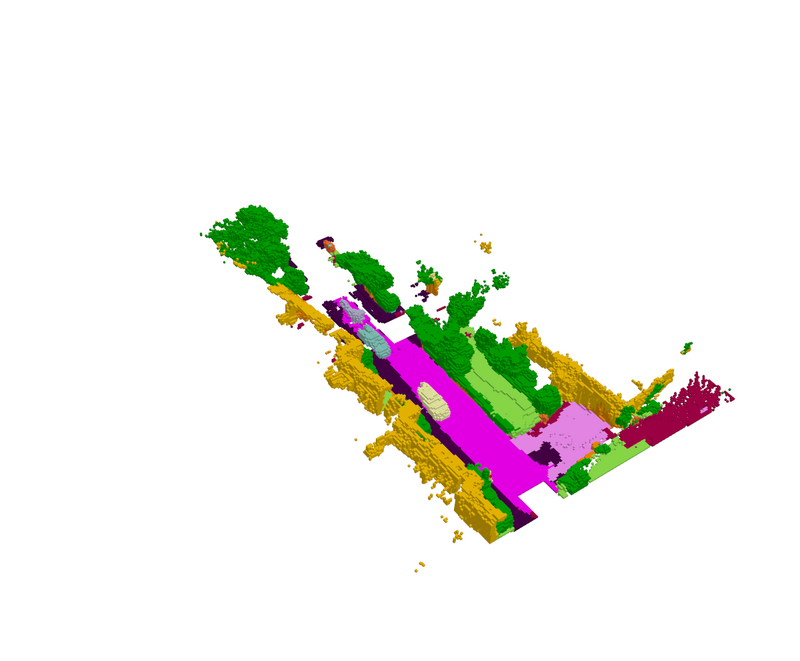} 
				& \imthird{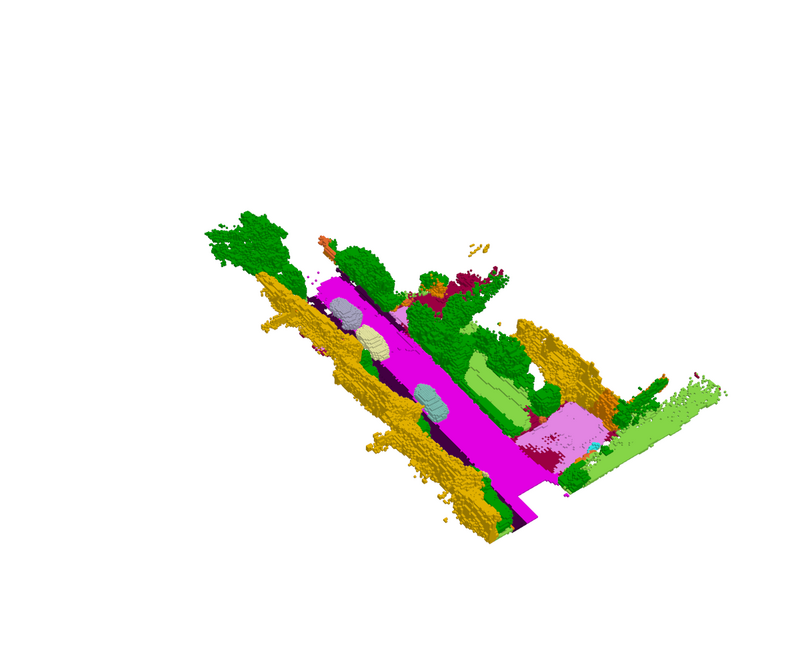} 
				& \imthird{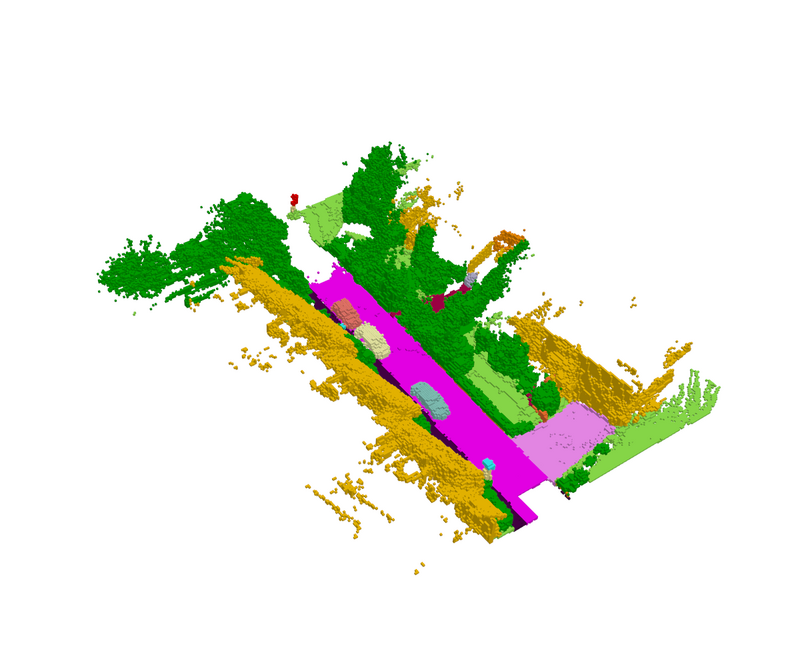}
				\\[-0.3em]

				& \imforth{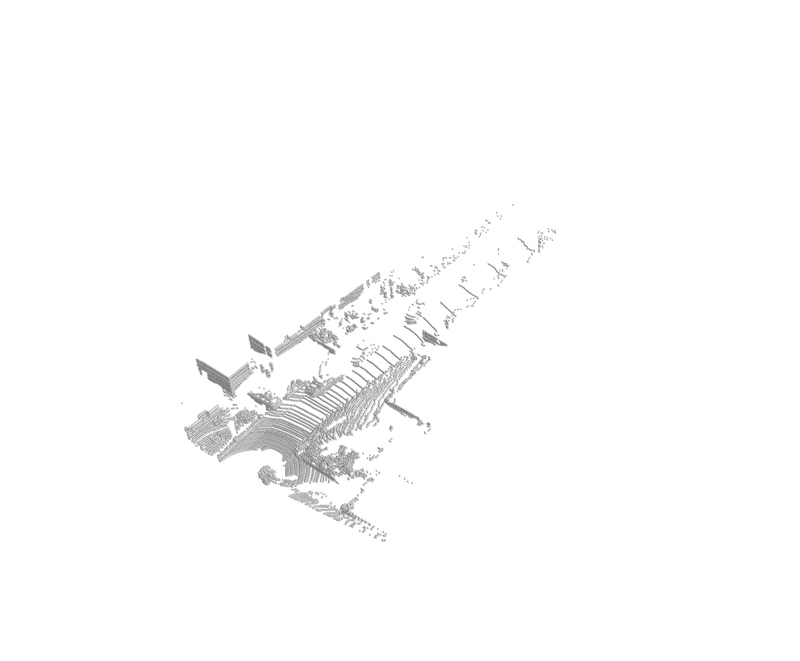} 
				& \imforth{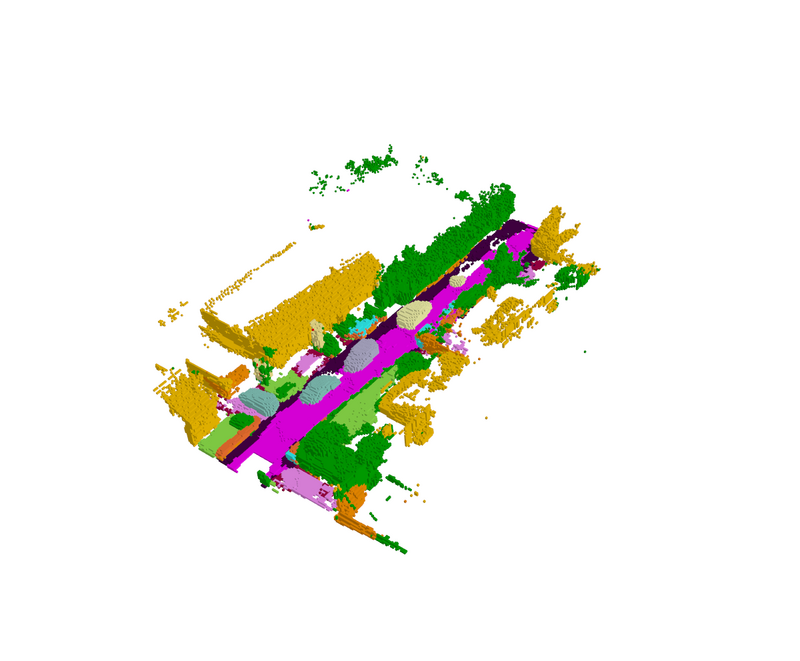} 
				& \imforth{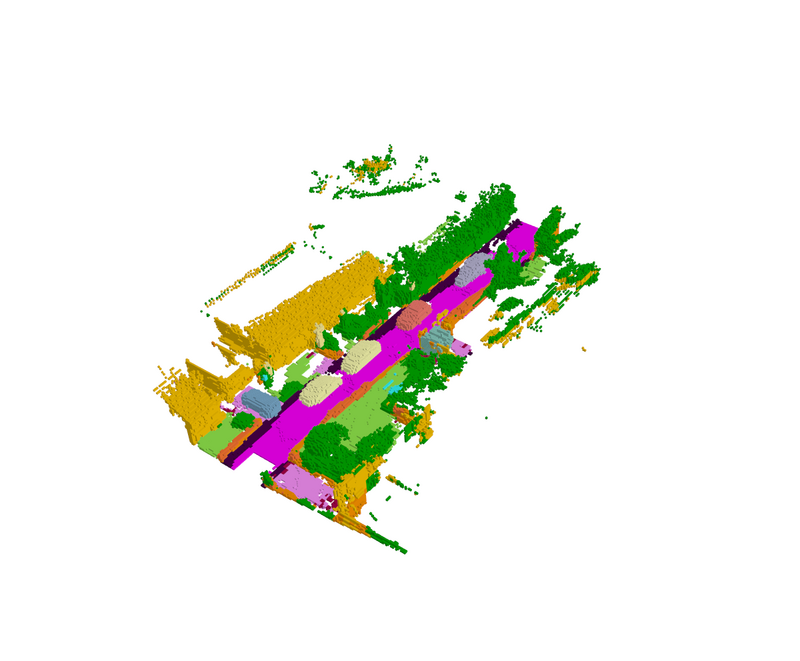} 
				& \imforth{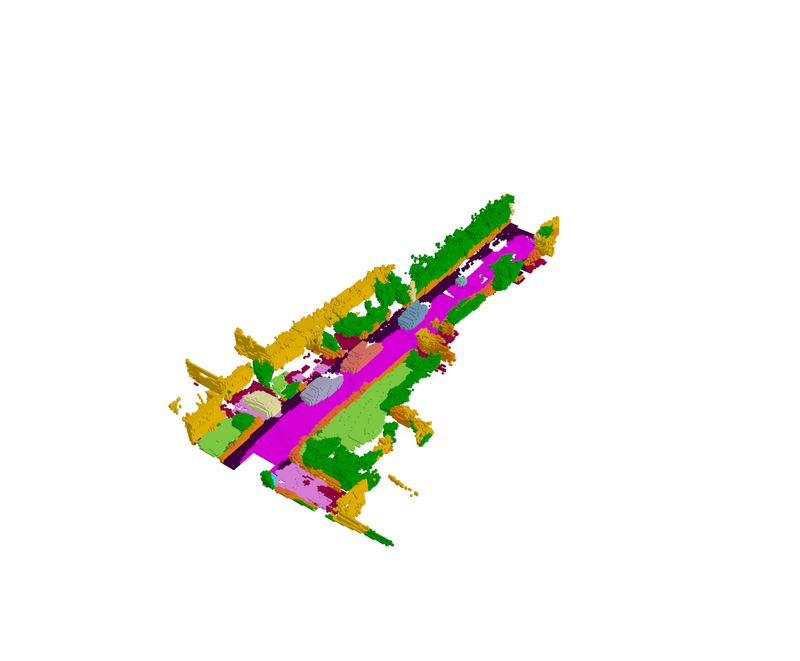} 
				& \imforth{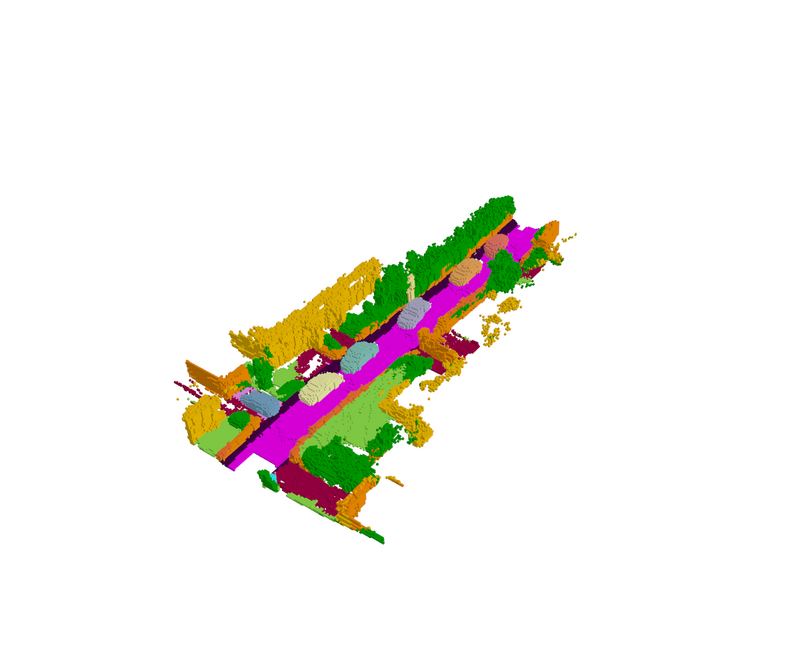} 
				& \imforth{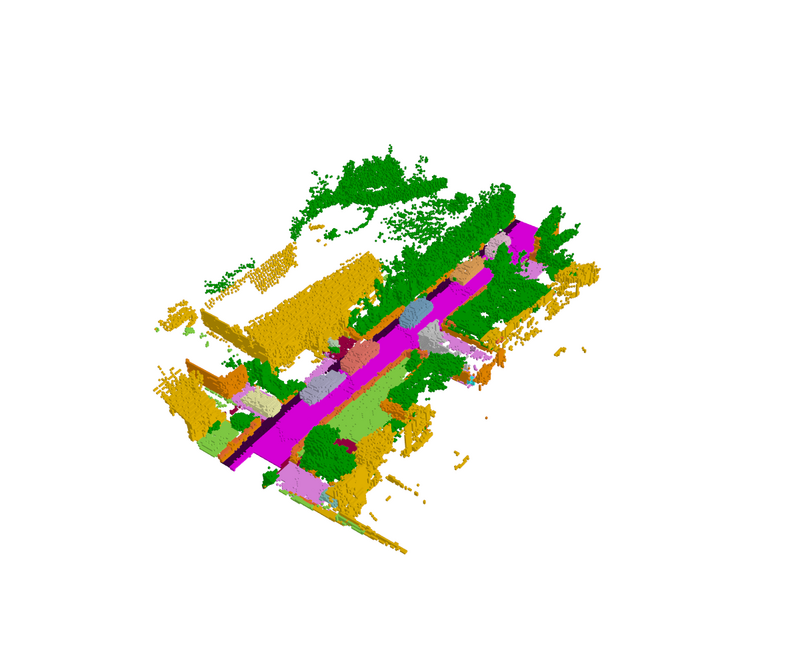}
				\\
				& \imfive{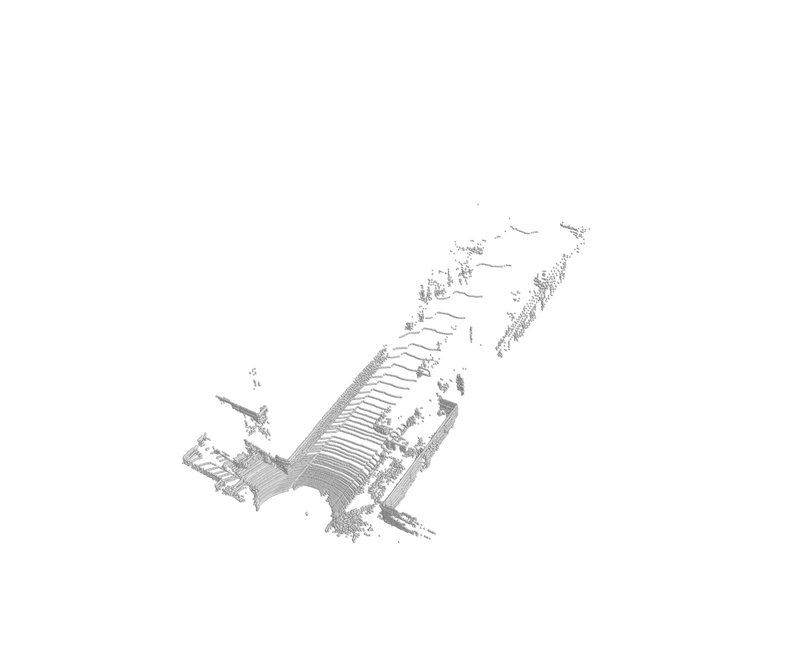} 
				& \imfive{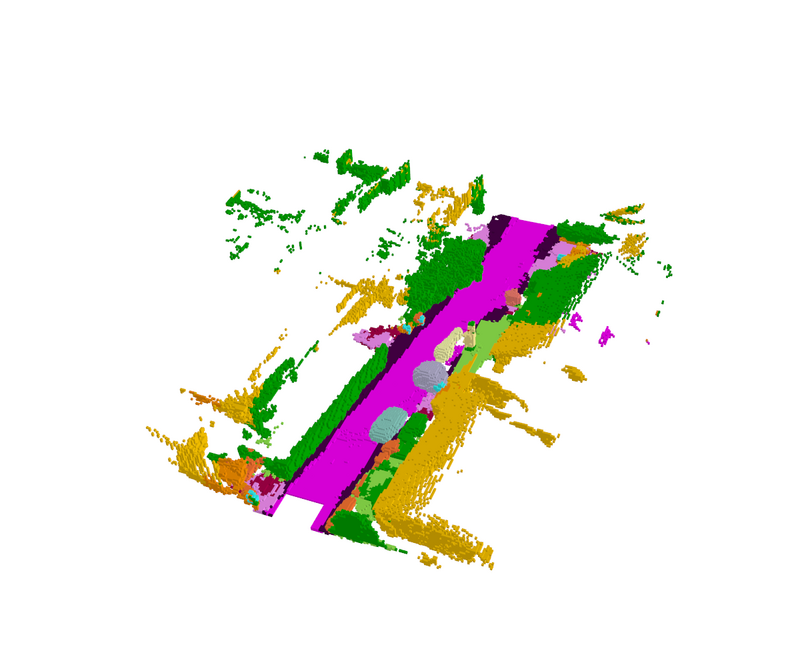} 
				& \imfive{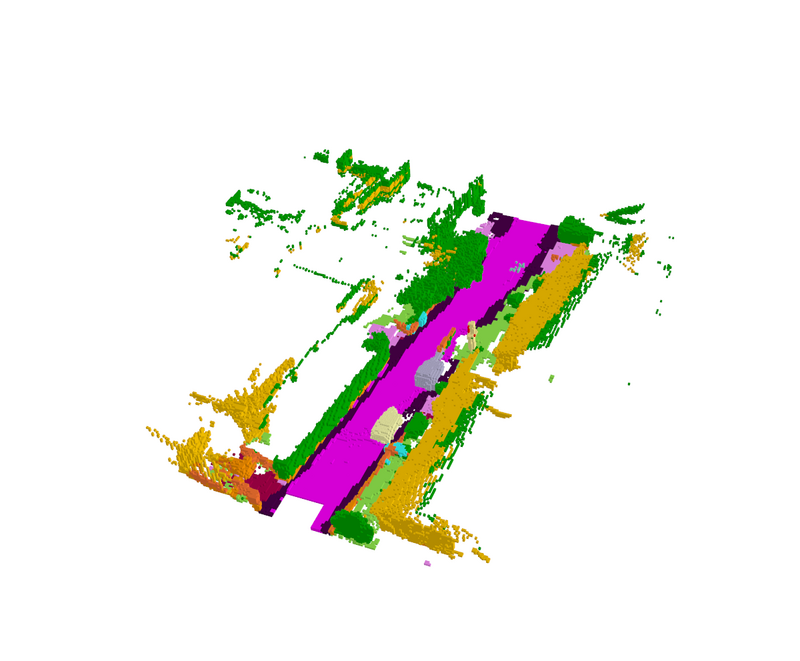} 
				& \imfive{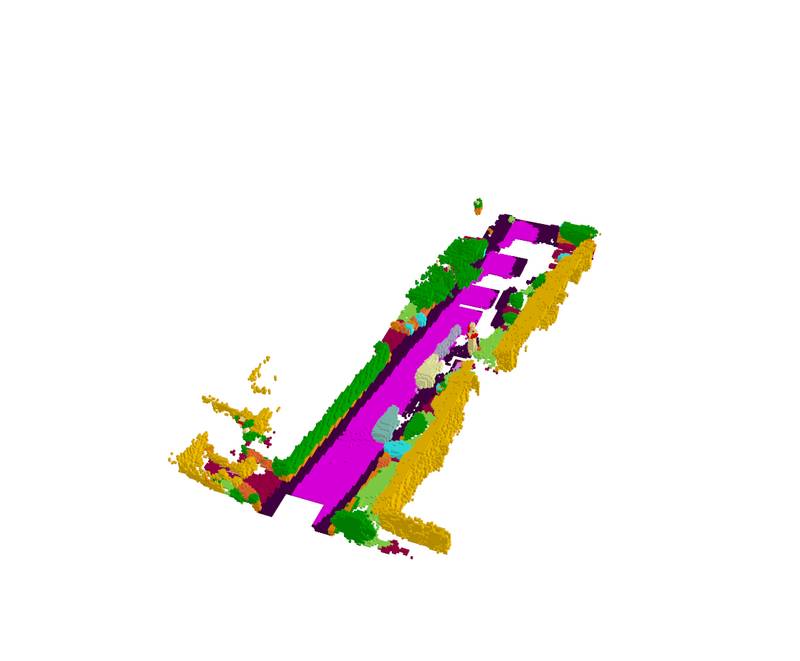} 
				& \imfive{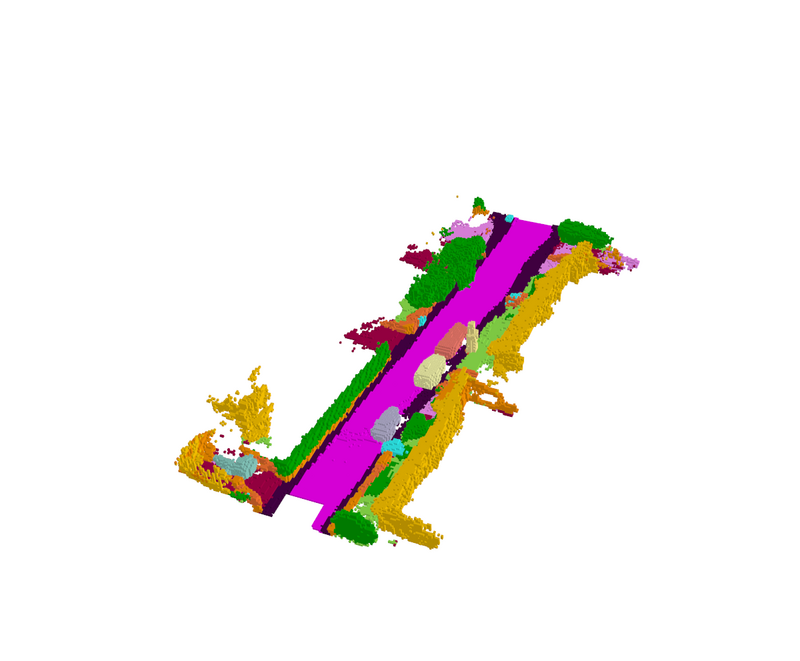} 
				& \imfive{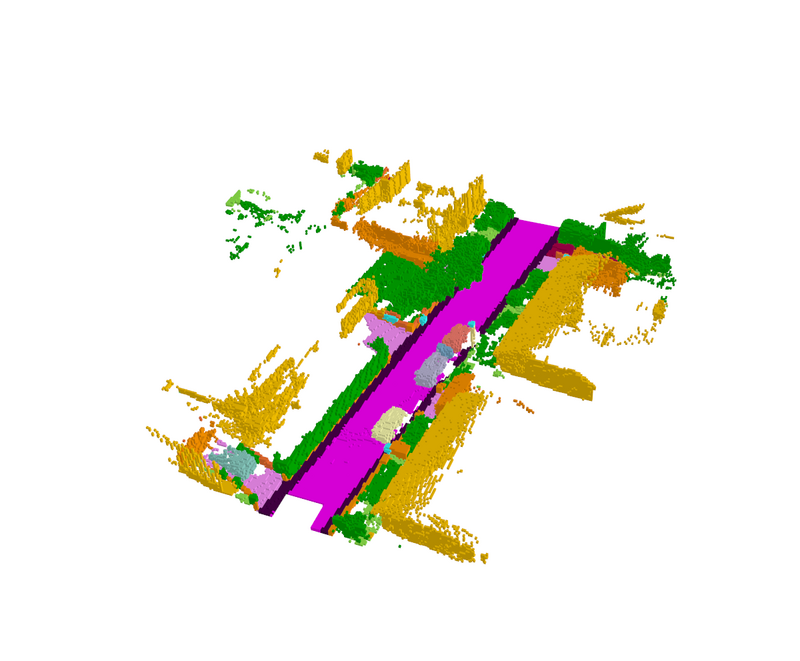}

			\end{tabular}
		}\\
		{				
			\tiny
			\textcolor{road}{$\blacksquare$}road~%
			\textcolor{parking}{$\blacksquare$}parking~%
			\textcolor{sidewalk}{$\blacksquare$}sidewalk~%
			\textcolor{other-ground}{$\blacksquare$}other ground~%
			\textcolor{building}{$\blacksquare$}building~%
			\textcolor{fence}{$\blacksquare$}fence~%
			\textcolor{vegetation}{$\blacksquare$}vegetation~%
			\textcolor{trunk}{$\blacksquare$}trunk~%
			\textcolor{terrain}{$\blacksquare$}terrain~%
			\textcolor{pole}{$\blacksquare$}pole~%
			\textcolor{traffic-sign}{$\blacksquare$}traffic sign~%
			\textcolor{other-struct}{$\blacksquare$}other structure~
			\textcolor{other-object}{$\blacksquare$}other object			
		}%
\caption{\textbf{Additional qualitative results on Panoptic Scene Completion}. \ours{} demonstrates improved instance quality, evident from superior geometry (rows 1, 3 to 6), and better separation (rows 2 to 4). Additionally, it predicts more accurate scene structure, with less missing geometry (rows 1, 3, 5, and 6).}
\label{fig:supp:qualitative_psc}
\end{figure*}

\paragraph{Qualitative results.}
\cref{fig:supp:qualitative_psc} presents further qualitative results on Semantic KITTI and SSCBench-KITTI360.
\ours predicts a more complete scene geometry, as illustrated in rows 1, 3, 5 and 6. It also infers better instance quality, illustrated by the increased accuracy in instance shape in rows 2, 4, 5 and 6, and by the clearer separation observable in rows 2, 3, 4 and 5.

\begin{figure*}
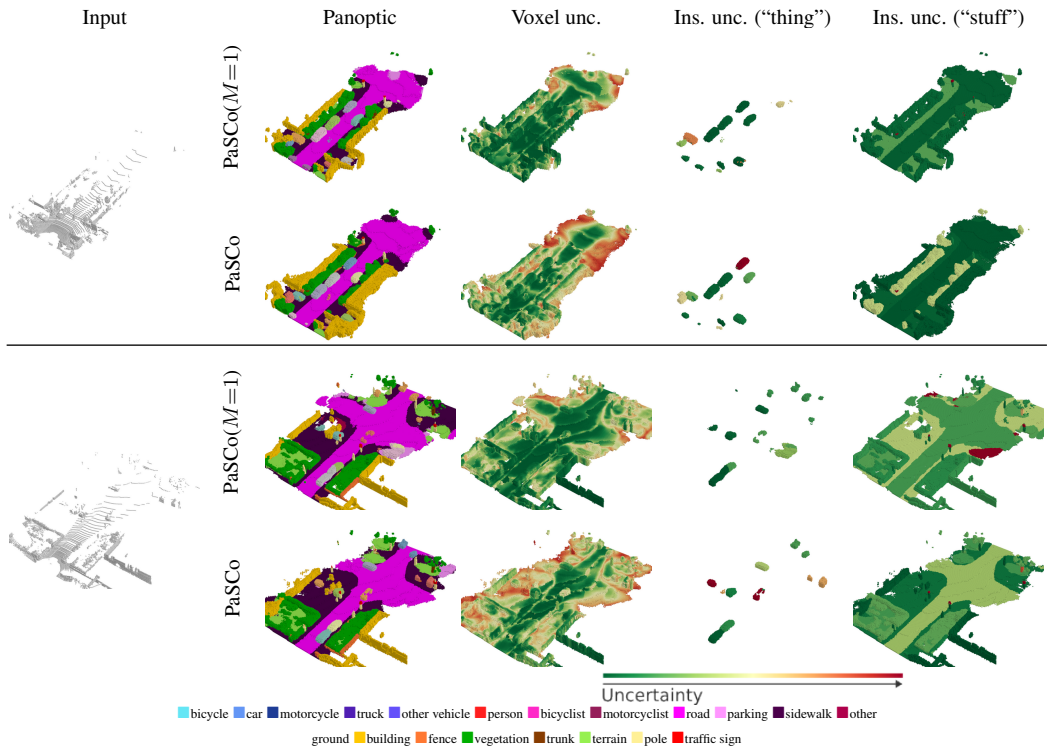
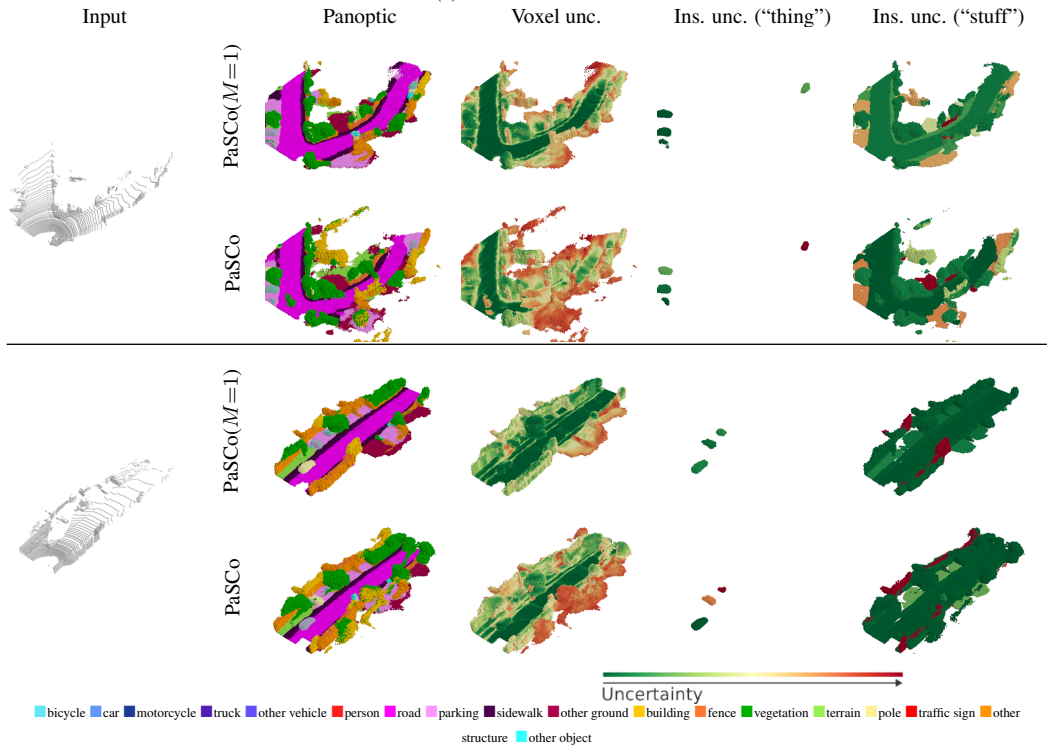

	\centering
	\begin{subfigure}{0.8\linewidth}
		\centering
		\footnotesize
		\newcommand{\SemanticKITTIEntry}[1]{
			
			\multirow{1}{*}{\adjincludegraphics[width=\linewidth, trim={.2\width} {.2\height} {.2\width} {.2\height}, clip]{figs_supp/ours1_input_#1_1.png}} & 
			\rotatebox{90}{\oursMIMO{1}} &
			\adjincludegraphics[width=\linewidth, trim={.2\width} {.2\height} {.2\width} {.2\height}, clip]{figs_supp/ours1_panop_#1_1.png} &
			\adjincludegraphics[width=\linewidth, trim={.2\width} {.2\height} {.2\width} {.2\height}, clip]{figs_supp/ours1_median_#1.png} &
			\adjincludegraphics[width=\linewidth, trim={.2\width} {.2\height} {.2\width} {.2\height}, clip]{figs_supp/ours1_thing_conf_#1_1.png} &
			\adjincludegraphics[width=\linewidth, trim={.2\width} {.2\height} {.2\width} {.2\height}, clip]{figs_supp/ours1_stuff_conf_#1_1.png} \\   
			& \rotatebox{90}{\ours} &
			\adjincludegraphics[width=\linewidth, trim={.2\width} {.2\height} {.2\width} {.2\height}, clip]{figs_supp/ours3_panop_#1_3.png} &
			\adjincludegraphics[width=\linewidth, trim={.2\width} {.2\height} {.2\width} {.2\height}, clip]{figs_supp/ours3_median_#1.png} &
			\adjincludegraphics[width=\linewidth, trim={.2\width} {.2\height} {.2\width} {.2\height}, clip]{figs_supp/ours3_thing_conf_#1_3.png} &
			\adjincludegraphics[width=\linewidth, trim={.2\width} {.2\height} {.2\width} {.2\height}, clip]{figs_supp/ours3_stuff_conf_#1_3.png} \\
			\\[-0.1em]
		}

		\renewcommand{\arraystretch}{0.0}
		\setlength{\tabcolsep}{0.003\textwidth}
		\newcolumntype{P}[1]{>{\centering\arraybackslash}m{#1}}
		\resizebox{\linewidth}{!}{
			\begin{tabular}{  P{0.18\textwidth} P{0.05\textwidth} P{0.18\textwidth} P{0.18\textwidth} P{0.18\textwidth} P{0.18\textwidth}}
				\\
				Input & & Panoptic & Voxel unc. &  Ins. unc. (``thing") & Ins. unc. (``stuff")
				\\
				\SemanticKITTIEntry{1230}
				\midrule
				\SemanticKITTIEntry{3950}
				& & & \multicolumn{3}{c}{\adjincludegraphics[width=4cm]{svg/colorbar.png}}  \\
				
			\end{tabular}
			
		}\\
		{				
			\tiny
			\textcolor{bicycle}{$\blacksquare$}bicycle~%
			\textcolor{car}{$\blacksquare$}car~%
			\textcolor{motorcycle}{$\blacksquare$}motorcycle~%
			\textcolor{truck}{$\blacksquare$}truck~%
			\textcolor{other-vehicle}{$\blacksquare$}other vehicle~%
			\textcolor{person}{$\blacksquare$}person~%
			\textcolor{bicyclist}{$\blacksquare$}bicyclist~%
			\textcolor{motorcyclist}{$\blacksquare$}motorcyclist~%
			\textcolor{road}{$\blacksquare$}road~%
			\textcolor{parking}{$\blacksquare$}parking~%
			\textcolor{sidewalk}{$\blacksquare$}sidewalk~%
			\textcolor{other-ground}{$\blacksquare$}other ground~%
			\textcolor{building}{$\blacksquare$}building~%
			\textcolor{fence}{$\blacksquare$}fence~%
			\textcolor{vegetation}{$\blacksquare$}vegetation~%
			\textcolor{trunk}{$\blacksquare$}trunk~%
			\textcolor{terrain}{$\blacksquare$}terrain~%
			\textcolor{pole}{$\blacksquare$}pole~%
			\textcolor{traffic-sign}{$\blacksquare$}traffic sign~%
		}%
		
		\caption{
			\textbf{Semantic KITTI.}}
		\label{fig:supp:qualitative_uncertainty_semkitti}
	\end{subfigure} 
	
	\begin{subfigure}{.8\linewidth}
		\centering
		\footnotesize
		\newcommand{\SSCBenchKITTIEntry}[1]{
			
			\multirow{1}{*}{\adjincludegraphics[width=\linewidth, trim={.2\width} {.2\height} {.2\width} {.2\height}, clip]{figs_supp/kitti360/ours1_input_#1_1.png}} & 
			\rotatebox{90}{\oursMIMO{1}} &
			\adjincludegraphics[width=\linewidth, trim={.2\width} {.2\height} {.2\width} {.2\height}, clip]{figs_supp/kitti360/ours1_panop_#1_1.png} &
			\adjincludegraphics[width=\linewidth, trim={.2\width} {.2\height} {.2\width} {.2\height}, clip]{figs_supp/kitti360/ours1_median_#1.png} &
			\adjincludegraphics[width=\linewidth, trim={.2\width} {.2\height} {.2\width} {.2\height}, clip]{figs/kitti360/ours1_ins_conf_#1_1.png} &
			\adjincludegraphics[width=\linewidth, trim={.2\width} {.2\height} {.2\width} {.2\height}, clip]{figs_supp/kitti360/ours1_stuff_conf_#1_1.png} \\   
			& \rotatebox{90}{\ours} &
			\adjincludegraphics[width=\linewidth, trim={.2\width} {.2\height} {.2\width} {.2\height}, clip]{figs_supp/kitti360/ours2_panop_#1_1.png} &
			\adjincludegraphics[width=\linewidth, trim={.2\width} {.2\height} {.2\width} {.2\height}, clip]{figs_supp/kitti360/ours2_median_#1.png} &
			\adjincludegraphics[width=\linewidth, trim={.2\width} {.2\height} {.2\width} {.2\height}, clip]{figs/kitti360/ours2_ins_conf_#1_1.png} &
			\adjincludegraphics[width=\linewidth, trim={.2\width} {.2\height} {.2\width} {.2\height}, clip]{figs_supp/kitti360/ours2_stuff_conf_#1_1.png} \\[-0.1em]
		}
		
		\renewcommand{\arraystretch}{0.0}
		\setlength{\tabcolsep}{0.003\textwidth}
		\newcolumntype{P}[1]{>{\centering\arraybackslash}m{#1}}
		\resizebox{\linewidth}{!}{
			\begin{tabular}{  P{0.18\textwidth} P{0.05\textwidth} P{0.18\textwidth} P{0.18\textwidth} P{0.18\textwidth} P{0.18\textwidth}}
				\\
				Input & & Panoptic & Voxel unc. &  Ins. unc. (``thing") & Ins. unc. (``stuff")
				\\
				
				\SSCBenchKITTIEntry{2665}
				\midrule
				\SSCBenchKITTIEntry{2580}
				& & & \multicolumn{3}{c}{\adjincludegraphics[width=4cm]{svg/colorbar.png}}  \\
				
			\end{tabular}	
		}		\\
		{				
			\tiny
			\textcolor{bicycle}{$\blacksquare$}bicycle~%
			\textcolor{car}{$\blacksquare$}car~%
			\textcolor{motorcycle}{$\blacksquare$}motorcycle~%
			\textcolor{truck}{$\blacksquare$}truck~%
			\textcolor{other-vehicle}{$\blacksquare$}other vehicle~%
			\textcolor{person}{$\blacksquare$}person~%
			\textcolor{road}{$\blacksquare$}road~%
			\textcolor{parking}{$\blacksquare$}parking~%
			\textcolor{sidewalk}{$\blacksquare$}sidewalk~%
			\textcolor{other-ground}{$\blacksquare$}other ground~%
			\textcolor{building}{$\blacksquare$}building~%
			\textcolor{fence}{$\blacksquare$}fence~%
			\textcolor{vegetation}{$\blacksquare$}vegetation~%
			\textcolor{terrain}{$\blacksquare$}terrain~%
			\textcolor{pole}{$\blacksquare$}pole~%
			\textcolor{traffic-sign}{$\blacksquare$}traffic sign~%
			\textcolor{other-struct}{$\blacksquare$}other structure~
			\textcolor{other-object}{$\blacksquare$}other object			
		}%
		\caption{
			\textbf{Qualitative uncertainty comparison on SSCBench-KITTI360 (val. set.}}
		
		\label{fig:supp:qualitative_uncertainty_sscbenchkitti360}
	\end{subfigure} 
	\caption{
		\textbf{Additional qualitative uncertainty comparison on \ Semantic KITTI and SSCBench-KITTI360.} 
		\ours~offers more insightful estimates of uncertainty, particularly in smaller instances with incomplete geometry, along the boundaries of segments, in areas with sparse input points, and in extrapolated regions.}
	\label{fig:supp:qualitative_uncertainty}
\end{figure*}

\clearpage
\clearpage

{
    \small
    \bibliographystyle{ieeenat_fullname}
    \bibliography{main}

\begin{thebibliography}{80}
\providecommand{\natexlab}[1]{#1}
\providecommand{\url}[1]{\texttt{#1}}
\expandafter\ifx\csname urlstyle\endcsname\relax
  \providecommand{\doi}[1]{doi: #1}\else
  \providecommand{\doi}{doi: \begingroup \urlstyle{rm}\Url}\fi

\bibitem[Abdar et~al.(2021)Abdar, Pourpanah, Hussain, Rezazadegan, Liu,
  Ghavamzadeh, Fieguth, Cao, Khosravi, Acharya, Makarenkov, and
  Nahavandi]{uncertaintyreview}
Moloud Abdar, Farhad Pourpanah, Sadiq Hussain, Dana Rezazadegan, Li Liu,
  Mohammad Ghavamzadeh, Paul Fieguth, Xiaochun Cao, Abbas Khosravi, U.~Rajendra
  Acharya, Vladimir Makarenkov, and Saeid Nahavandi.
\newblock A review of uncertainty quantification in deep learning: Techniques,
  applications and challenges.
\newblock \emph{Information Fusion}, 2021.

\bibitem[Ayhan and Berens(2018)]{tta}
Murat~Seckin Ayhan and Philipp Berens.
\newblock Test-time data augmentation for estimation of heteroscedastic
  aleatoric uncertainty in deep neural networks.
\newblock In \emph{MIDL}, 2018.

\bibitem[Behley et~al.(2019)Behley, Garbade, Milioto, Quenzel, Behnke,
  Stachniss, and Gall]{semkitti}
J. Behley, M. Garbade, A. Milioto, J. Quenzel, S. Behnke, C. Stachniss, and J.
  Gall.
\newblock {SemanticKITTI: A Dataset for Semantic Scene Understanding of LiDAR
  Sequences}.
\newblock In \emph{ICCV}, 2019.

\bibitem[Berman et~al.(2018)Berman, Triki, and Blaschko]{lovaszsoftmax}
Maxim Berman, Amal~Rannen Triki, and Matthew~B. Blaschko.
\newblock The lov\'asz-softmax loss: A tractable surrogate for the optimization
  of the intersection-over-union measure in neural networks.
\newblock In \emph{CVPR}, 2018.

\bibitem[Blundell et~al.(2015)Blundell, Cornebise, Kavukcuoglu, and
  Wierstra]{weightuncertainty}
Charles Blundell, Julien Cornebise, Koray Kavukcuoglu, and Daan Wierstra.
\newblock Weight uncertainty in neural networks.
\newblock In \emph{ICML}, 2015.

\bibitem[Caesar et~al.(2020)Caesar, Bankiti, Lang, Vora, Liong, Xu, Krishnan,
  Pan, Baldan, and Beijbom]{caesar2020nuscenes}
Holger Caesar, Varun Bankiti, Alex~H Lang, Sourabh Vora, Venice~Erin Liong,
  Qiang Xu, Anush Krishnan, Yu Pan, Giancarlo Baldan, and Oscar Beijbom.
\newblock nuscenes: A multimodal dataset for autonomous driving.
\newblock In \emph{CVPR}, 2020.

\bibitem[Cai et~al.(2021)Cai, Chen, Zhang, Lin, Wang, and Li]{SISNet}
Yingjie Cai, Xuesong Chen, Chao Zhang, Kwan-Yee Lin, Xiaogang Wang, and
  Hongsheng Li.
\newblock Semantic scene completion via integrating instances and scene
  in-the-loop.
\newblock In \emph{CVPR}, 2021.

\bibitem[Cao and de~Charette(2022)]{monoscene}
Anh-Quan Cao and Raoul de Charette.
\newblock Monoscene: Monocular 3d semantic scene completion.
\newblock In \emph{CVPR}, 2022.

\bibitem[Chen et~al.(2022)Chen, Liang, Liu, Wang, and Liu]{chen2022approach}
Hui Chen, Man Liang, Wanquan Liu, Weina Wang, and Peter~Xiaoping Liu.
\newblock An approach to boundary detection for 3d point clouds based on dbscan
  clustering.
\newblock \emph{Pattern Recognition}, 2022.

\bibitem[Chen et~al.(2020)Chen, Lin, Qian, Zeng, and Li]{3dsketch}
Xiaokang Chen, Kwan-Yee Lin, Chen Qian, Gang Zeng, and Hongsheng Li.
\newblock 3d sketch-aware semantic scene completion via semi-supervised
  structure prior.
\newblock In \emph{CVPR}, 2020.

\bibitem[Chen et~al.(2019)Chen, Garbade, and Gall]{Chen20193DSS}
Yueh-Tung Chen, Martin Garbade, and Juergen Gall.
\newblock 3d semantic scene completion from a single depth image using
  adversarial training.
\newblock In \emph{ICIP}, 2019.

\bibitem[Cheng et~al.(2021)Cheng, Schwing, and Kirillov]{maskformer}
Bowen Cheng, Alexander~G. Schwing, and Alexander Kirillov.
\newblock Per-pixel classification is not all you need for semantic
  segmentation.
\newblock In \emph{NeurIPS}, 2021.

\bibitem[Cheng et~al.(2022)Cheng, Misra, Schwing, Kirillov, and
  Girdhar]{mask2former}
Bowen Cheng, Ishan Misra, Alexander~G. Schwing, Alexander Kirillov, and Rohit
  Girdhar.
\newblock Masked-attention mask transformer for universal image segmentation.
\newblock In \emph{CVPR}, 2022.

\bibitem[Cheng et~al.(2020)Cheng, Agia, Ren, Li, and Liu]{s3cnet}
Ran Cheng, Christopher Agia, Yuan Ren, Xinhai Li, and Bingbing Liu.
\newblock S3cnet: A sparse semantic scene completion network for lidar point
  clouds.
\newblock In \emph{CoRL}, 2020.

\bibitem[Choy et~al.(2019)Choy, Gwak, and Savarese]{mink}
Christopher Choy, JunYoung Gwak, and Silvio Savarese.
\newblock 4d spatio-temporal convnets: Minkowski convolutional neural networks.
\newblock In \emph{CVPR}, 2019.

\bibitem[Dahnert et~al.(2021)Dahnert, Hou, Nie{\ss}ner, and
  Dai]{dahnert2021panoptic}
Manuel Dahnert, Ji Hou, Matthias Nie{\ss}ner, and Angela Dai.
\newblock Panoptic 3d scene reconstruction from a single rgb image.
\newblock \emph{Advances in Neural Information Processing Systems},
  34:\penalty0 8282--8293, 2021.

\bibitem[Dai et~al.(2018)Dai, Ritchie, Bokeloh, Reed, Sturm, and
  Nie{\ss}ner]{scancomplete}
Angela Dai, Daniel Ritchie, Martin Bokeloh, Scott Reed, J{\"u}rgen Sturm, and
  Matthias Nie{\ss}ner.
\newblock Scancomplete: Large-scale scene completion and semantic segmentation
  for 3d scans.
\newblock In \emph{Proceedings of the IEEE Conference on Computer Vision and
  Pattern Recognition}, pages 4578--4587, 2018.

\bibitem[Dai et~al.(2020)Dai, Diller, and Nie{\ss}ner]{sgnn}
Angela Dai, Christian Diller, and Matthias Nie{\ss}ner.
\newblock Sg-nn: Sparse generative neural networks for self-supervised scene
  completion of rgb-d scans.
\newblock In \emph{CVPR}, 2020.

\bibitem[Dourado et~al.(2020)Dourado, de~Campos, Kim, and Hilton]{EdgeNet}
Aloisio Dourado, Teofilo~E. de Campos, Hansung Kim, and Adrian Hilton.
\newblock {EdgeNet: Semantic scene completion from a single RGB-D image}.
\newblock In \emph{ICPR}, 2020.

\bibitem[Durasov et~al.(2021)Durasov, Bagautdinov, Baque, and Fua]{masksembles}
Nikita Durasov, Timur Bagautdinov, Pierre Baque, and Pascal Fua.
\newblock Masksembles for uncertainty estimation.
\newblock In \emph{CVPR}, 2021.

\bibitem[Dusenberry et~al.(2020)Dusenberry, Jerfel, Wen, Ma, Snoek, Heller,
  Lakshminarayanan, and Tran]{bnnrank1}
Michael~W. Dusenberry, Ghassen Jerfel, Yeming Wen, Yi-An Ma, Jasper Snoek,
  Katherine Heller, Balaji Lakshminarayanan, and Dustin Tran.
\newblock Efficient and scalable bayesian neural nets with rank-1 factors.
\newblock In \emph{ICML}, 2020.

\bibitem[Ester et~al.(1996)Ester, Kriegel, Sander, Xu,
  et~al.]{ester1996density}
Martin Ester, Hans-Peter Kriegel, J{\"o}rg Sander, Xiaowei Xu, et~al.
\newblock A density-based algorithm for discovering clusters in large spatial
  databases with noise.
\newblock In \emph{KDD}, 1996.

\bibitem[Franchi et~al.(2020)Franchi, Bursuc, Aldea, Dubuisson, and
  Bloch]{franchi2021tradi}
Gianni Franchi, Andrei Bursuc, Emanuel Aldea, Severine Dubuisson, and Isabelle
  Bloch.
\newblock Tradi: Tracking deep neural network weight distributions for
  uncertainty estimation.
\newblock In \emph{ECCV}, 2020.

\bibitem[Gal and Ghahramani(2016)]{mcdropout}
Yarin Gal and Zoubin Ghahramani.
\newblock Dropout as a bayesian approximation: Representing model uncertainty
  in deep learning.
\newblock In \emph{ICML}, 2016.

\bibitem[Garbade et~al.(2019)Garbade, Sawatzky, Richard, and Gall]{TS3D}
Martin Garbade, Johann Sawatzky, Alexander Richard, and Juergen Gall.
\newblock Two stream 3d semantic scene completion.
\newblock In \emph{CVPRW}, 2019.

\bibitem[Garipov et~al.(2018)Garipov, Izmailov, Podoprikhin, Vetrov, and
  Wilson]{garipov2018loss}
Timur Garipov, Pavel Izmailov, Dmitrii Podoprikhin, Dmitry Vetrov, and
  Andrew~Gordon Wilson.
\newblock Loss surfaces, mode connectivity, and fast ensembling of dnns.
\newblock In \emph{NeurIPS}, 2018.

\bibitem[Gasperini et~al.(2021)Gasperini, Mahani, Marcos-Ramiro, Navab, and
  Tombari]{panoster}
Stefano Gasperini, Mohammad-Ali~Nikouei Mahani, Alvaro Marcos-Ramiro, Nassir
  Navab, and Federico Tombari.
\newblock Panoster: End-to-end panoptic segmentation of {LiDAR} point clouds.
\newblock \emph{RA-L}, 2021.

\bibitem[Geiger et~al.(2013)Geiger, Lenz, Stiller, and
  Urtasun]{geiger2013vision}
Andreas Geiger, Philip Lenz, Christoph Stiller, and Raquel Urtasun.
\newblock Vision meets robotics: The kitti dataset.
\newblock \emph{International Journal of Robotics Research}, 2013.

\bibitem[Guo et~al.(2017)Guo, Pleiss, Sun, and Weinberger]{guo2017calibration}
Chuan Guo, Geoff Pleiss, Yu Sun, and Kilian~Q Weinberger.
\newblock On calibration of modern neural networks.
\newblock In \emph{ICML}. PMLR, 2017.

\bibitem[Gustafsson et~al.(2020)Gustafsson, Danelljan, and
  Schön]{gustafsson2020evaluating}
Fredrik~K. Gustafsson, Martin Danelljan, and Thomas~B. Schön.
\newblock Evaluating scalable bayesian deep learning methods for robust
  computer vision.
\newblock In \emph{CVPRW}, 2020.

\bibitem[Havasi et~al.(2021)Havasi, Jenatton, Fort, Liu, Snoek,
  Lakshminarayanan, Dai, and Tran]{mimo}
Marton Havasi, Rodolphe Jenatton, Stanislav Fort, Jeremiah~Zhe Liu, Jasper
  Snoek, Balaji Lakshminarayanan, Andrew~M. Dai, and Dustin Tran.
\newblock Training independent subnetworks for robust prediction.
\newblock In \emph{ICLR}, 2021.

\bibitem[He et~al.(2017)He, Gkioxari, Dollár, and Girshick]{he2018mask}
Kaiming He, Georgia Gkioxari, Piotr Dollár, and Ross Girshick.
\newblock Mask r-cnn.
\newblock In \emph{ICCV}, 2017.

\bibitem[Hong et~al.(2021)Hong, Zhou, Zhu, Li, and Liu]{lidarbasedpanop}
Fangzhou Hong, Hui Zhou, Xinge Zhu, Hongsheng Li, and Ziwei Liu.
\newblock Lidar-based panoptic segmentation via dynamic shifting network.
\newblock In \emph{CVPR}, 2021.

\bibitem[Hong et~al.(2024)Hong, Kong, Zhou, Zhu, Li, and Liu]{hong2024unified}
Fangzhou Hong, Lingdong Kong, Hui Zhou, Xinge Zhu, Hongsheng Li, and Ziwei Liu.
\newblock Unified 3d and 4d panoptic segmentation via dynamic shifting
  networks.
\newblock \emph{TPAMI}, 2024.

\bibitem[Hou et~al.(2020)Hou, Dai, and Nie{\ss}ner]{revealnet}
Ji Hou, Angela Dai, and Matthias Nie{\ss}ner.
\newblock Revealnet: Seeing behind objects in rgb-d scans.
\newblock In \emph{Proceedings of the IEEE/CVF Conference on Computer Vision
  and Pattern Recognition}, pages 2098--2107, 2020.

\bibitem[Huang et~al.(2017)Huang, Li, Pleiss, Liu, Hopcroft, and
  Weinberger]{snapshotensembles}
Gao Huang, Yixuan Li, Geoff Pleiss, Zhuang Liu, John~E. Hopcroft, and Kilian~Q.
  Weinberger.
\newblock Snapshot ensembles: Train 1, get m for free.
\newblock In \emph{ICLR}, 2017.

\bibitem[Huang et~al.(2023)Huang, Zheng, Zhang, Zhou, and Lu]{tpvformer}
Yuanhui Huang, Wenzhao Zheng, Yunpeng Zhang, Jie Zhou, and Jiwen Lu.
\newblock Tri-perspective view for vision-based 3d semantic occupancy
  prediction.
\newblock In \emph{CVPR}, 2023.

\bibitem[Hurtado et~al.(2020)Hurtado, Mohan, Burgard, and Valada]{mopt}
Juana~Valeria Hurtado, Rohit Mohan, Wolfram Burgard, and Abhinav Valada.
\newblock Mopt: Multi-object panoptic tracking.
\newblock In \emph{CVPR}, 2020.

\bibitem[Jordan et~al.(1999)Jordan, Ghahramani, Jaakkola, and Saul]{varinf}
Michael~I. Jordan, Zoubin Ghahramani, Tommi~S. Jaakkola, and Lawrence~K. Saul.
\newblock \emph{An Introduction to Variational Methods for Graphical Models},
  page 105–161.
\newblock MIT Press, 1999.

\bibitem[Kirillov et~al.(2019)Kirillov, He, Girshick, Rother, and
  Dollár]{panopticsegmentation}
Alexander Kirillov, Kaiming He, Ross Girshick, Carsten Rother, and Piotr
  Dollár.
\newblock Panoptic segmentation.
\newblock In \emph{CVPR}, 2019.

\bibitem[Kong et~al.(2023)Kong, Liu, Li, Chen, Zhang, Ren, Pan, Chen, and
  Liu]{robo3d}
Lingdong Kong, Youquan Liu, Xin Li, Runnan Chen, Wenwei Zhang, Jiawei Ren,
  Liang Pan, Kai Chen, and Ziwei Liu.
\newblock Robo3d: Towards robust and reliable 3d perception against
  corruptions.
\newblock In \emph{ICCV}, 2023.

\bibitem[Kuhn(1955)]{hungarian}
H.~W. Kuhn.
\newblock The hungarian method for the assignment problem.
\newblock \emph{Naval Research Logistics Quarterly}, 1955.

\bibitem[Lakshminarayanan et~al.(2017)Lakshminarayanan, Pritzel, and
  Blundell]{deepensemble}
Balaji Lakshminarayanan, Alexander Pritzel, and Charles Blundell.
\newblock Simple and scalable predictive uncertainty estimation using deep
  ensembles.
\newblock In \emph{NeurIPS}, 2017.

\bibitem[Laurent et~al.(2023)Laurent, Lafage, Tartaglione, Daniel, Martinez,
  Bursuc, and Franchi]{packedensemble}
Olivier Laurent, Adrien Lafage, Enzo Tartaglione, Geoffrey Daniel, Jean-Marc
  Martinez, Andrei Bursuc, and Gianni Franchi.
\newblock Packed-ensembles for efficient uncertainty estimation.
\newblock In \emph{ICLR}, 2023.

\bibitem[Li et~al.(2022{\natexlab{a}})Li, Razani, Xu, and Liu]{smacseg}
Enxu Li, Ryan Razani, Yixuan Xu, and Bingbing Liu.
\newblock Smac-seg: Lidar panoptic segmentation via sparse multi-directional
  attention clustering.
\newblock In \emph{ICRA}, 2022{\natexlab{a}}.

\bibitem[Li et~al.(2023{\natexlab{a}})Li, Razani, Xu, and Liu]{cpseg}
Enxu Li, Ryan Razani, Yixuan Xu, and Bingbing Liu.
\newblock Cpseg: Cluster-free panoptic segmentation of 3d lidar point clouds.
\newblock In \emph{ICRA}, 2023{\natexlab{a}}.

\bibitem[Li et~al.(2020)Li, Han, Wang, Liu, and Yuan]{aicnet}
Jie Li, Kai Han, Peng Wang, Yu Liu, and Xia Yuan.
\newblock Anisotropic convolutional networks for 3d semantic scene completion.
\newblock In \emph{CVPR}, 2020.

\bibitem[Li et~al.(2021)Li, Ding, and Huang]{IMENet}
Jie Li, Laiyan Ding, and Rui Huang.
\newblock Imenet: Joint 3d semantic scene completion and 2d semantic
  segmentation through iterative mutual enhancement.
\newblock In \emph{IJCAI}, 2021.

\bibitem[Li et~al.(2022{\natexlab{b}})Li, He, Wen, Gao, Cheng, and
  Zhang]{panopticphnet}
Jinke Li, Xiao He, Yang Wen, Yuan Gao, Xiaoqiang Cheng, and Dan Zhang.
\newblock Panoptic-phnet: Towards real-time and high-precision lidar panoptic
  segmentation via clustering pseudo heatmap.
\newblock In \emph{CVPR}, 2022{\natexlab{b}}.

\bibitem[Li et~al.(2023{\natexlab{b}})Li, Li, Liu, Gong, Li, Chen, Wang, Li,
  Jiang, Yu, Wang, Zhao, Yu, and Feng]{sscbench}
Yiming Li, Sihang Li, Xinhao Liu, Moonjun Gong, Kenan Li, Nuo Chen, Zijun Wang,
  Zhiheng Li, Tao Jiang, Fisher Yu, Yue Wang, Hang Zhao, Zhiding Yu, and Chen
  Feng.
\newblock Sscbench: Monocular 3d semantic scene completion benchmark in street
  views.
\newblock \emph{arXiv}, 2023{\natexlab{b}}.

\bibitem[Li et~al.(2023{\natexlab{c}})Li, Yu, Choy, Xiao, Alvarez, Fidler,
  Feng, and Anandkumar]{voxformer}
Yiming Li, Zhiding Yu, Christopher Choy, Chaowei Xiao, Jose~M Alvarez, Sanja
  Fidler, Chen Feng, and Anima Anandkumar.
\newblock Voxformer: Sparse voxel transformer for camera-based 3d semantic
  scene completion.
\newblock In \emph{CVPR}, 2023{\natexlab{c}}.

\bibitem[Liao et~al.(2022)Liao, Xie, and Geiger]{kitti360}
Yiyi Liao, Jun Xie, and Andreas Geiger.
\newblock {KITTI}-360: A novel dataset and benchmarks for urban scene
  understanding in 2d and 3d.
\newblock \emph{TPAMI}, 2022.

\bibitem[Liu et~al.(2022{\natexlab{a}})Liu, Zhou, Zhao, Li, Du, Keutzer, Du,
  and Zhang]{panopconstrast}
Minzhe Liu, Qiang Zhou, Hengshuang Zhao, Jianing Li, Yuan Du, Kurt Keutzer, Li
  Du, and Shanghang Zhang.
\newblock Prototype-voxel contrastive learning for lidar point cloud panoptic
  segmentation.
\newblock In \emph{ICRA}, 2022{\natexlab{a}}.

\bibitem[Liu et~al.(2018)Liu, Hu, Zeng, Tang, Jin, Han, and Li]{SATNet}
Shice Liu, Yu Hu, Yiming Zeng, Qiankun Tang, Beibei Jin, Yinhe Han, and Xiaowei
  Li.
\newblock {See and Think: Disentangling Semantic Scene Completion}.
\newblock In \emph{NeurIPS}, 2018.

\bibitem[Liu et~al.(2022{\natexlab{b}})Liu, Chen, Atashgahi, Chen, Sokar,
  Mocanu, Pechenizkiy, Wang, and Mocanu]{desparse}
Shiwei Liu, Tianlong Chen, Zahra Atashgahi, Xiaohan Chen, Ghada Sokar, Elena
  Mocanu, Mykola Pechenizkiy, Zhangyang Wang, and Decebal~Constantin Mocanu.
\newblock Deep ensembling with no overhead for either training or testing: The
  all-round blessings of dynamic sparsity.
\newblock In \emph{2022}, 2022{\natexlab{b}}.

\bibitem[Loshchilov and Hutter(2019)]{adamW}
Ilya Loshchilov and Frank Hutter.
\newblock Decoupled weight decay regularization.
\newblock In \emph{ICLR}, 2019.

\bibitem[MacKay(1992)]{mackay}
David J.~C. MacKay.
\newblock A practical bayesian framework for backpropagation networks.
\newblock \emph{Neural Computation}, 1992.

\bibitem[Maddox et~al.(2019)Maddox, Garipov, Izmailov, Vetrov, and
  Wilson]{maddox2019simple}
Wesley Maddox, Timur Garipov, Pavel Izmailov, Dmitry Vetrov, and Andrew~Gordon
  Wilson.
\newblock A simple baseline for bayesian uncertainty in deep learning.
\newblock In \emph{NeurIPS}, 2019.

\bibitem[Marcuzzi et~al.(2023)Marcuzzi, Nunes, Wiesmann, Behley, and
  Stachniss]{maskpls}
R. Marcuzzi, L. Nunes, L. Wiesmann, J. Behley, and C. Stachniss.
\newblock {Mask-Based Panoptic LiDAR Segmentation for Autonomous Driving}.
\newblock \emph{RA-L}, 2023.

\bibitem[Mei et~al.(2023)Mei, Yang, Wang, Li, Hou, Ra, Li, and Liu]{CenterLPS}
Jianbiao Mei, Yu Yang, Mengmeng Wang, Zizhang Li, Xiaojun Hou, Jongwon Ra,
  Laijian Li, and Yong Liu.
\newblock Centerlps: Segment instances by centers for lidar panoptic
  segmentation.
\newblock In \emph{ACM MM}, 2023.

\bibitem[Milioto et~al.(2020)Milioto, Behley, McCool, and
  Stachniss]{rangepanop}
Andres Milioto, Jens Behley, Chris McCool, and Cyrill Stachniss.
\newblock Lidar panoptic segmentation for autonomous driving.
\newblock In \emph{IROS}, 2020.

\bibitem[Ovadia et~al.(2019)Ovadia, Fertig, Ren, Nado, Sculley, Nowozin,
  Dillon, Lakshminarayanan, and Snoek]{ovadia2019trust}
Yaniv Ovadia, Emily Fertig, Jie Ren, Zachary Nado, D Sculley, Sebastian
  Nowozin, Joshua~V. Dillon, Balaji Lakshminarayanan, and Jasper Snoek.
\newblock Can you trust your model's uncertainty? evaluating predictive
  uncertainty under dataset shift.
\newblock In \emph{NeurIPS}, 2019.

\bibitem[Porzi et~al.(2019)Porzi, Bulò, Colovic, and
  Kontschieder]{porzi2019seamless}
Lorenzo Porzi, Samuel~Rota Bulò, Aleksander Colovic, and Peter Kontschieder.
\newblock Seamless scene segmentation.
\newblock In \emph{CVPR}, 2019.

\bibitem[Puy et~al.(2023)Puy, Boulch, and Marlet]{waffleiron}
Gilles Puy, Alexandre Boulch, and Renaud Marlet.
\newblock Using a waffle iron for automotive point cloud semantic segmentation.
\newblock In \emph{ICCV}, 2023.

\bibitem[Razani et~al.(2021)Razani, Cheng, Li, Taghavi, Ren, and
  Bingbing]{gps3net}
Ryan Razani, Ran Cheng, Enxu Li, Ehsan Taghavi, Yuan Ren, and Liu Bingbing.
\newblock Gp-s3net: Graph-based panoptic sparse semantic segmentation network.
\newblock In \emph{ICCV}, 2021.

\bibitem[Rold{\~a}o et~al.(2020)Rold{\~a}o, de~Charette, and
  Verroust-Blondet]{lmscnet}
Luis Rold{\~a}o, Raoul de Charette, and Anne Verroust-Blondet.
\newblock Lmscnet: Lightweight multiscale 3d semantic completion.
\newblock In \emph{3DV}, 2020.

\bibitem[Rold{\~a}o et~al.(2021)Rold{\~a}o, De~Charette, and
  Verroust-Blondet]{sscsurvey}
Luis Rold{\~a}o, Raoul De~Charette, and Anne Verroust-Blondet.
\newblock {3D Semantic Scene Completion: a Survey}.
\newblock \emph{IJCV}, 2021.

\bibitem[Sirohi et~al.(2021)Sirohi, Mohan, Büscher, Burgard, and
  Valada]{efficientlps}
Kshitij Sirohi, Rohit Mohan, Daniel Büscher, Wolfram Burgard, and Abhinav
  Valada.
\newblock Efficientlps: Efficient lidar panoptic segmentation.
\newblock \emph{T-RO}, 2021.

\bibitem[Song et~al.(2017)Song, Yu, Zeng, Chang, Savva, and Funkhouser]{sscnet}
Shuran Song, Fisher Yu, Andy Zeng, Angel~X Chang, Manolis Savva, and Thomas
  Funkhouser.
\newblock Semantic scene completion from a single depth image.
\newblock In \emph{CVPR}, 2017.

\bibitem[Sun et~al.(2020)Sun, Kretzschmar, Dotiwalla, Chouard, Patnaik, Tsui,
  Guo, Zhou, Chai, Caine, et~al.]{sun2020scalability}
Pei Sun, Henrik Kretzschmar, Xerxes Dotiwalla, Aurelien Chouard, Vijaysai
  Patnaik, Paul Tsui, James Guo, Yin Zhou, Yuning Chai, Benjamin Caine, et~al.
\newblock Scalability in perception for autonomous driving: Waymo open dataset.
\newblock In \emph{CVPR}, 2020.

\bibitem[Wang et~al.(2019)Wang, Tan, Navab, and Tombari]{ForkNet}
Yida Wang, David~Joseph Tan, Nassir Navab, and Federico Tombari.
\newblock {ForkNet: Multi-branch volumetric semantic completion from a single
  depth image}.
\newblock In \emph{ICCV}, 2019.

\bibitem[Wang et~al.(2024)Wang, Chen, Liao, Fan, and Zhang]{panopocc}
Yuqi Wang, Yuntao Chen, Xingyu Liao, Lue Fan, and Zhaoxiang Zhang.
\newblock Panoocc: Unified occupancy representation for camera-based 3d
  panoptic segmentation.
\newblock In \emph{CVPR}, 2024.

\bibitem[Wen et~al.(2020)Wen, Tran, and Ba]{batchensemble}
Yeming Wen, Dustin Tran, and Jimmy Ba.
\newblock Batchensemble: An alternative approach to efficient ensemble and
  lifelong learning.
\newblock In \emph{ICLR}, 2020.

\bibitem[Wilson and Izmailov(2020)]{wilson2022bayesian}
Andrew~Gordon Wilson and Pavel Izmailov.
\newblock Bayesian deep learning and a probabilistic perspective of
  generalization.
\newblock In \emph{NeurIPS}, 2020.

\bibitem[Xia et~al.(2023)Xia, Liu, Li, Zhu, Ma, Li, Hou, and Qiao]{scpnet}
Zhaoyang Xia, Youquan Liu, Xin Li, Xinge Zhu, Yuexin Ma, Yikang Li, Yuenan Hou,
  and Yu Qiao.
\newblock Scpnet: Semantic scene completion on point cloud.
\newblock In \emph{CVPR}, 2023.

\bibitem[Yan et~al.(2021)Yan, Gao, Li, Zhang, Li, Huang, and Cui]{js3cnet}
Xu Yan, Jiantao Gao, Jie Li, Ruimao Zhang, Zhen Li, Rui Huang, and Shuguang
  Cui.
\newblock Sparse single sweep lidar point cloud segmentation via learning
  contextual shape priors from scene completion.
\newblock In \emph{AAAI}, 2021.

\bibitem[Ye et~al.(2021)Ye, Chen, Han, and Liao]{pnal}
Shuquan Ye, Dongdong Chen, Songfang Han, and Jing Liao.
\newblock Learning with noisy labels for robust point cloud segmentation.
\newblock In \emph{ICCV}, 2021.

\bibitem[Zhang et~al.(2018)Zhang, Zhao, Yao, Chen, Zhang, and
  Liao]{zhang2018efficient}
Jiahui Zhang, Hao Zhao, Anbang Yao, Yurong Chen, Li Zhang, and Hongen Liao.
\newblock Efficient semantic scene completion network with spatial group
  convolution.
\newblock In \emph{ECCV}, 2018.

\bibitem[Zhang et~al.(2019)Zhang, Liu, Lei, Lu, and Yang]{CCPNet}
Pingping Zhang, Wei Liu, Yinjie Lei, Huchuan Lu, and Xiaoyun Yang.
\newblock Cascaded context pyramid for full-resolution 3d semantic scene
  completion.
\newblock In \emph{ICCV}, 2019.

\bibitem[Zhu et~al.(2021)Zhu, Zhou, Wang, Hong, Ma, Li, Li, and
  Lin]{zhu2020cylindrical}
Xinge Zhu, Hui Zhou, Tai Wang, Fangzhou Hong, Yuexin Ma, Wei Li, Hongsheng Li,
  and Dahua Lin.
\newblock Cylindrical and asymmetrical 3d convolution networks for lidar
  segmentation.
\newblock In \emph{CVPR}, 2021.

\end{thebibliography}
}

\end{document}